\documentclass{ut-thesis}
\usepackage[colorlinks]{hyperref} %
\usepackage{natbib}
\usepackage{graphicx} %
\usepackage{booktabs} %
\usepackage{lipsum} %
\usepackage{amsmath}
\usepackage{mathtools}
\usepackage{overpic}
\usepackage{color}
\usepackage{listings} %
\usepackage[skip=.2em]{caption} %
\usepackage{xcolor}
\usepackage{currfile}
\usepackage{cancel}
\usepackage{float}
\usepackage{rotating}
\usepackage[rightcaption]{sidecap}  %
\sidecaptionvpos{figure}{t} %
\usepackage{microtype}
\usepackage{graphicx}
\usepackage{lipsum}
\usepackage{tikz}
\usepackage{booktabs} %
\usepackage{caption}
\usepackage{subcaption}

\usepackage{hyperref}

\usepackage{xspace}

\usepackage{overpic}
\usepackage{caption}        %
\usepackage{amsthm}
\usepackage{bbm}
\usepackage{amsmath}
\usepackage{mathtools}
\usepackage{amssymb}
\usepackage{float}

\usepackage{pgfplots, pgfplotstable}
\usepackage{algorithm}

\usepackage{enumitem}
\usepackage{multirow}
\usepackage[title]{appendix}
\usepackage{listings}
\usepackage{xcolor}
\usepackage{wrapfig}

\definecolor{dkgreen}{rgb}{0,0.6,0}
\definecolor{gray}{rgb}{0.5,0.5,0.5}
\definecolor{mauve}{rgb}{0.58,0,0.82}

\definecolor{dark_green}{rgb}{0, 0.5, 0}

\newcommand{\argmin}[1]{\underset{#1}{\operatorname{argmin}}}

\definecolor{Fcolor}{HTML}{af2418}

\definecolor{Ccolor}{HTML}{ffd359}

\def\Real{{\mathbb R}}

\def\int{\mathrm{int}}

\def\eg{{\em e.g.}}

\newcommand{\Figure}[1]{Figure~\ref{fig:#1}}

\newcommand{\Table}[1]{Table~\ref{tab:#1}}
\newcommand{\eq}[1]{(\ref{eq:#1})}

\renewcommand{\paragraph}[1]{\vspace{.1em}\noindent\textbf{#1.~}}

\usepackage{lipsum}

\def\image{\mathbf{I}}
\renewcommand{\next}[1]{{#1}'}
\def\params{\omega}
\def\encoder{\mathcal{E}_\params}

\def\masknet{\mathcal{D}_\params}

\def\capsule{\mathbf{c}}
\def\pose{\boldsymbol{\theta}}
\def\flow{\Phi}

\newcommand{\depth}{d}

\newcommand{\shape}{{\bf s}}

\newcommand{\pixel}{{\bf u}} %

\newcommand{\expect}[2]{\mathbb{E}_{#1}~#2}
\newcommand{\mask}{\boldsymbol{\Lambda}}

\newcommand{\T}{\mathbf{P}}
\newcommand{\F}{\mathbf{T}}

\usepackage{bbm}
\usepackage{amstext}
\usepackage{amsmath}
\usepackage{xcolor}
\usepackage[format=plain,labelformat=simple,labelsep=period,font=small,compatibility=false]{caption}
\usepackage{cleveref}
\usepackage{xspace}
\usepackage{amssymb}

\newcommand{\clean}{clean}

\newcommand{\methodname}{SpotLessSplats\xspace}
\newcommand{\methodshortname}{SLS\xspace}
\newcommand{\methodsdmlp}{\methodshortname-mlp\xspace}
\newcommand{\methodsdcluster}{{\methodshortname}-agg\xspace}
\newcommand{\methodprune}{UBP\xspace}
\newcommand{\methodfilter}{RobustFilter\xspace}

\newcommand{\tdgsshortname}{3DGS\xspace}

\newcommand{\mean}{\mu}
\newcommand{\covariance}{\boldsymbol{\Sigma}}
\newcommand{\opacity}{\alpha}
\newcommand{\sphcoef}{\mathbf{c}}
\newcommand{\viewxform}{\mathbf{W}}
\newcommand{\projjacobian}{\mathbf{J}}
\newcommand{\rot}{\mathbf{R}}
\newcommand{\scale}{\mathbf{S}}
\newcommand{\nImages}{N}
\newcommand{\model}{\mathcal{G}}
\newcommand{\element}{g}
\newcommand{\latent}{\mathbf{z}}
\newcommand{\mlp}{\mathcal{H}}
\newcommand{\mlpparam}{\theta}

\newcommand{\features}{\mathbf{F}}

\newcommand{\upperm}{\mathbf{U}}
\newcommand{\lowerm}{\mathbf{L}}

\newcommand{\step}{\scalebox{0.8}{(t)}}
\newcommand{\steps}{T}
\newcommand{\residual}{\mathbf{R}}
\newcommand{\real}{\mathbb{R}}
\newcommand{\indicator}{\mathbbm{1}}
\newcommand{\filter}{\mathbf{B}}

\newcommand{\loss}[1]{\mathcal{L}_\text{#1}}
\newcommand{\sched}{\alpha}
\newcommand{\maxx}{\text{max}}
\newcommand{\x}{x}
\newcommand{\util}{u}
\newcommand{\pthresh}{\kappa}
\newcommand{\aglolatent}{z}
\newcommand{\aglonet}{\mathcal{Q}}
\newcommand{\cluster}{\mathbf{C}}
\newcommand{\numclusters}{C}

\definecolor{cvprblue}{rgb}{0.21,0.49,0.74}
\definecolor{gold}{rgb}{0.7, 0.5, 0}

\newcommand{\lpips}{\scalebox{0.8}{LPIPS$\downarrow$}}
\newcommand{\ssim}{\scalebox{0.8}{SSIM$\uparrow$}}
\newcommand{\psnr}{\scalebox{0.8}{PSNR$\uparrow$}}

\renewcommand{\paragraph}[1]{\vspace{.2em}\noindent\textbf{#1}.}

\usepackage{soul}
\setuldepth{foobar}

\usepackage{multirow}

\usepackage{amsmath}
\usepackage{mathtools}
\usepackage{overpic}
\usepackage{color}
\usepackage{listings} %
\usepackage[skip=.2em]{caption} %
\usepackage{xcolor}
\usepackage{currfile}
\usepackage{cancel}
\usepackage{float}
\usepackage{rotating}

\usepackage[rightcaption]{sidecap}  %
\sidecaptionvpos{figure}{t} %
\definecolor{turquoise}{cmyk}{0.65,0,0.1,0.3}
\definecolor{purple}{rgb}{0.65,0,0.65}
\definecolor{dark_green}{rgb}{0, 0.5, 0}
\definecolor{orange}{rgb}{0.8, 0.6, 0.2}
\definecolor{red}{rgb}{0.8, 0.2, 0.2}
\definecolor{darkgray}{rgb}{0.5, 0.5, 0.5}
\definecolor{darkred}{rgb}{0.6, 0.1, 0.05}
\definecolor{blueish}{rgb}{0.0, 0.3, .6}
\definecolor{light_gray}{rgb}{0.7, 0.7, .7}
\definecolor{pink}{rgb}{1, 0, 1}
\definecolor{greyblue}{rgb}{0.25, 0.25, 1}

\newcommand{\At}[1]{\marginpar{\tiny{\textcolor{blueish}{ANDREA}}}}
\newcommand{\Df}[1]{\marginpar{\tiny{\textcolor{greyblue}{DAVID}}}}

\usepackage{bbm}
\usepackage{lipsum}
\usepackage{blindtext}

\renewcommand{\paragraph}[1]{\vspace{.5em}\noindent\textbf{#1}.}

\newcommand{\NeRF}{NeRF\xspace}
\newcommand{\RobustNeRF}{RobustNeRF\xspace}
\newcommand{\mipNeRFthreesixty}{mip-NeRF 360\xspace}
\newcommand{\ddnerf}{D$^2$NeRF\xspace}
\newcommand{\DDNeRF}{\ddnerf}
\newcommand{\NeRFW}{NeRF-W\xspace}
\newcommand{\SupplementaryMaterial}[1]{{supplementary material}}

\usepackage{enumitem}
\setlist[itemize]{noitemsep,leftmargin=*,topsep=0em}
\setlist[enumerate]{noitemsep,leftmargin=*,topsep=0em}

\newcommand{\C}{\mathbf{C}} %
\newcommand{\oracle}{\mathbf{S}}
\newcommand{\segmenter}{\mathcal{S}}

\newcommand{\radiance}{\mathbf{c}}

\newcommand{\ray}{\mathbf{r}}
\newcommand{\origin}{\mathbf{o}}
\newcommand{\dir}{\mathbf{d}}
\newcommand{\position}{\mathbf{x}}

\newcommand{\density}{\sigma}

\newcommand{\kernel}{\kappa}
\newcommand{\weight}{\omega}

\newcommand{\residuals}{\boldsymbol{\epsilon}}
\newcommand{\residuall}{\epsilon}
\newcommand{\threshold}{\mathcal{T}}

\newcommand{\paramss}{\boldsymbol{\theta}}

\author{Sara Sabour}
\title{Object Learning and Robust 3D Reconstruction}
\degree{Doctor of Philosophy}
\department[]{Department of Computer Science}
\gradyear{2025}
\begin{document}
  \frontmatter
    \maketitle
    \begin{abstract}
      As humans we have an inherent notion of what an object is in a scene or a picture. Even without color or texture, we can identify the presence of objects. Our perception of objects are usually not atomic and sometimes it is task dependant. We can even identify and count instances of  unseen categories of objects without hesitation. Psychological studies have talked about Gestalt laws as our cues for finding objects without supervision. Having the right foundation that would enable a machine to learn about identifying objects in a scene without supervision leads to less dependence on a set of manually labelled training data. Therefore, the model would have better out-of-domain generalization skills. Also, it can bypass the inherent biases of the training dataset by limiting the interaction of humans with the training dataset. Potentially, the model would be more reliable, trustworthy, and explainable, depending on the design choices. 
In this thesis we discuss architectural designs and training methods for a neural network to have the ability of dissecting an image into objects of interest without supervision.  The main challenge in 2D unsupervised object segmentation is distinguishing between foreground objects of interest and background. FlowCapsules uses motion as a cue for the objects of interest in 2D scenarios.
The last part of this thesis focuses on 3D applications where the goal is detecting and removal of the object of interest from the input images. In these tasks, we leverage the geometric consistency of scenes in 3D to detect the inconsistent dynamic objects. Our transient object masks are then used for designing robust optimization kernels to improve 3D modeling in a casual capture setup. One of our goals in this thesis is to show the merits of unsupervised object based approaches in computer vision. Furthermore, we suggest possible directions for defining objects of interest or foreground objects without requiring supervision. Our hope is to motivate and excite the community into further exploring explicit object representations in image understanding tasks.

    \end{abstract}
    \begin{dedication}
        Dedicated to my mother, Dr. Robabeh Moosavi.
    \end{dedication}
    \begin{acknowledgements}
      I am eternally thankful to every amazing and inspiring person in my life who have supported and mentored me during my graduate studies. I cannot describe the immense gratitude I feel toward David Fleet, my mentor and advisor. His openness to new problems, the freedom and encouragement of choosing my own directions, brilliant discussions and unwavering support made our meetings the highlight of my weeks. I have also been most fortunate to be supervised by Geoffrey Hinton. Geoff's dedication to his students, his tenacious curiosity, his integrity and moral compass and lastly his quick wit and delightful humor has been and is going to be my beacon of inspiration. Thank you Geoff for encouraging and enabling me to pursue my graduate studies. I am also thankful to Andrea Tagliasacchi for mentoring and encouraging me forward every step of the way. Andrea's unending ocean of ideas, his quick solutions, and support enabled me to navigate through the pandemic and other ups and downs of the past years. I am also thankful to Sanja Fidler whose guidance, feedback and suggestions during our supervisory committee meetings helped shape the direction of my research.

Most of all, I want to express the deepest gratitude to my family. My husband, Fartash Faghri, for being my source of comfort and joy, for always listening to my rambling ideas and bringing clarity. My mother, Robabeh Moosavi, for being the strongest woman that I know, for her love of learning and for dedicating her life to us. My father, Alireza Sabour, for his inspirational perseverance and never losing hope through hardships. My sister, Shiva Sabour, whose been always present for me with a warm hug and a holding hand, for bringing love and joy to every moment of life and for your unending pursuit of excellence. Thank you for being the best family I could ever imagine.

I am fortunate to have had the opportunity to collaborate with brilliant and inspiring members of Google Research and ML group at the Vector Institute including Nick Frosst, Yanshuai Cao, Adam Kosiorek, Klaus Greff, Thomas Kipf, Daniel Duckworth, Suhani Vora, Soroush Yazdani, William Chan, Mohammad Norouzi, Yee Whye Teh, Lily Goli, Laura Culp, Marcus Brubaker, Alec Jacobson, Saurabh Saxena, and George Kopanas. I would also like to express my gratitude to Kevin Swerski, Lala li, Maria Attarian, and Paul Vicol for the fruitful discussions. 

    \end{acknowledgements}
    \tableofcontents
    \listoftables
    \listoffigures
  \mainmatter
    \chapter{Introduction}

Infants from early on are able to learn about objects and tend to group perceptual arrays based on different criteria, one of which is their consistency over motion \citep{spelke1990principles}. In another study, 
\cite{spelke2007core} shows that as early as five months infants are able to conceptualize objects and understand the presence of an object in their mind even when perceptual arrays are not present. Human's idea of the definition of objects, grants them the ability to compose the objects and generalize their understanding to new scenes \citep{johnson2010mental}. Therefore, human and many animals are able to find the common groupings of visual arrays without supervision and form perceptual objects. Furthermore, studies into human psychology \citep{mather2006foundations}, suggest several properties that result in the human's generalization and robustness capabilities.

\begin{figure}[t]
    \centering
    \begin{subfigure}[t]{0.48\textwidth}
    \centering
    \includegraphics[width=.75\linewidth]{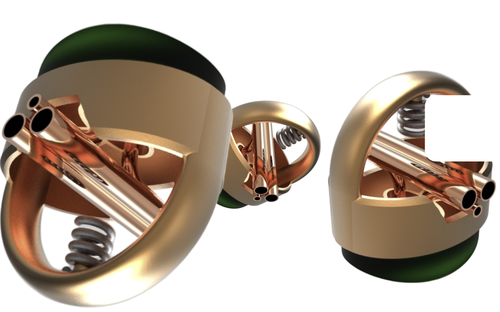}
    \caption{Viewpoint invariance. Independent of rotation, translation, or scale we are able to compare and count the unidentifiable objects.}
    \label{fig:invariance}
    \end{subfigure}
    \begin{subfigure}[t]{0.48\textwidth}
\includegraphics[width=\linewidth]{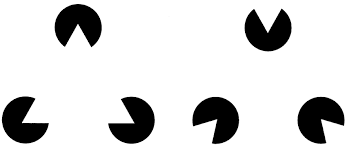}
    \caption{Kanizsa Triangle. Our perception completes a triangle, but only in the left image. Hence a whole is more than the sum of its parts.}
    \label{fig:kanisza}
    \end{subfigure}
    \caption{Examples of human visual perception properties based on gestalt psychology.}
\end{figure}

One such property is invariance to viewpoint transformations such as rotations, transformation and scale. Fig.~\ref{fig:invariance} is a collection of shapes that many may have not seen before. Yet, when faced with the question of the number of the objects in the figure, the unanimous answer would be three. Furthermore, if asked we would consider these shapes instances of the same object. Fig.~\ref{fig:invariance} shows an example for objects that we can not name or label, but we can count, compare and recognize in future. Presence of such properties in our object perception hints at a task independent understanding of objects in humans.

\begin{figure}[t]
    \centering
    \begin{subfigure}[t]{0.48\textwidth}
        \includegraphics[width=\linewidth]{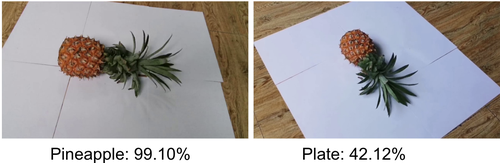}
    \caption{ResNet50 \citep{he2016deep} misclassifies a pineapple from a new viewpoint \cite{dong2022viewfool}.}
    \label{fig:pivsplate}
    \end{subfigure}
    \begin{subfigure}[t]{0.48\textwidth}
        \includegraphics[width=\linewidth]{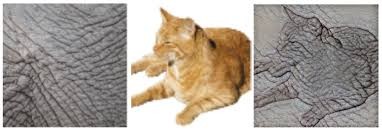}
    \caption{Deep neural networks are more sensitive to texture. A cat with the texture of elephant skin is classified as an elephant \cite{greff2020binding}.}
    \label{fig:catvselephant}
    \end{subfigure}
    \caption{Deep Learning has different principals for image understanding than humans.}
\end{figure}

Advancing the field of artificial intelligence to human level understanding requires a deep investigation into explicit and measurable differences between AI and human image understanding. Currently, deep learning is the leading technology in many artificial intelligence approaches. There are various factors involved in a deep neural network model that is successfully trained for a task. One of which is the architecture used and how the neurons are connected. Previously it has been shown \citep{sabour2017dynamic} that convolutional neural networks lack the viewpoint invariance property. With the advancements of new architectures, such as vision transformers \citep{dosovitskiy2020image}, neural networks were hoped to improve in their viewpoint invariance. \cite{dong2022viewfool} shows that both convolutional based state-of-the-art approaches and transformer based ones are still susceptible to suffering from slight viewpoint variations from their training data (Fig.~\ref{fig:pivsplate}).

Another such discrepancy is in human's way of perceiving Gestalt objects as opposed to deep learning methods. Fig.~\ref{fig:kanisza} shows an example of reification in human perception. By looking at the Kanizsa image we immediately perceive a triangle only on the left composition. The emergent triangle suggests our understanding of even simple geometric shapes is a gestault of their parts. The relative positioning of parts is instrumental in our recognition of whole objects. Furthermore, it shows that our image understanding can fill in missing sections pointing to the law of completeness in Gestault psychology. On the other hand several works \citep{baker2018deep, brendel2019approximating} has shown that a bag of features approach works well for deep neural networks. This was initiated by an interesting visualization in \cite{olah2017feature} which hints at correlation between strong and repeated presence of parts and whole detection. Another peculiar aspect of deep learning object recognition is their fixation on texture rather than shape as shown in Fig.~\ref{fig:catvselephant}. Short-cut learning \citep{geirhos2020shortcut} provides interesting insight into the significant role of dataset and task biases in deep learning models.

Humans are capable of generalizing and comprehending beyond their immediate experiences (datasets and tasks). We can understand and reason about novel scenes, being atoms under a microscope or stars and galaxies seen through telescopes. The generalization and robustness gap between our capabilities and deep learning methods is wide. Systematically researching to address the discrepancies between neural network understanding and human perception is a viable approach for improving the general capabilities of our AI models. A short-sighted method to improve the viewpoint invariance of neural networks can be augmenting the datasets with more viewpoint variation. Moreover, we can even improve the shape bias of the neural networks by data augmentation \citep{lee2022improving}. We argue that this would be addressing the symptom rather than the underlying issues in current deep learning techniques. The difference of properties in deep learning and human perception, such as viewpoint invariance or texture bias, is just an indicator of underlying missing components in our neural networks. 
An excerpt from \cite{geirhos2020shortcut} says: There was a time when the existence of tides, the planet’s orbit
around the sun, and the observation that “things fall down” were all largely considered to be
independent phenomena—until 1687, when Isaac Newton formulated his law of gravitation
that provided an elegantly simple explanation to all of these (and many more). Physics has
made tremendous progress over the last few centuries, but the thriving field of deep learning
is still very much at the beginning of its journey—often lacking a detailed understanding
of the underlying principles.

Our research on capsule networks and on unsupervised 3D object removal \citep{sabour2017dynamic,hinton2018matrix,kosiorek2019stacked,sun2021canonical,sabour2021unsupervised, kipf2021conditional, Sabour2023robustnerf, sabour2024spotlesssplats} during the past few years have been targeting these principal gaps between deep learning and human vision properties. Our contributions aim at giving proof of concept alterations and techniques that exceeds expectations by principal approach in object perception. The two main aspects that we want to emphasize is first the importance of unsupervised training of models and second the significance of shape-pose disentanglement in object-based learning.
 In general, our works in this thesis show the robustness merits of having objects in representation learning.
\begin{itemize}
    \item \textbf{Unsupervised training} offers significant advantages for deep learning models in computer vision, particularly in the context of fairness and flexibility. By utilizing vast amounts of unlabelled data, these models can learn patterns and features without being constrained by dataset and labeling biases that often affect supervised approaches. This lack of predefined categories allows the models to generalize more effectively to novel objects and scenarios, reducing the risk of shortcut learning \citep{geirhos2020shortcut}—a phenomenon where models latch onto superficial cues instead of meaningful patterns. Additionally, unsupervised learning fosters a more equitable representation of diverse data, as it does not impose artificial boundaries that might overlook minority classes or underrepresented features. Ultimately, this approach enhances the model's ability to adapt and perform across a wider range of applications, making it a powerful tool in the applicability of our methods in-the-wild.
\item \textbf{Shape-Pose Disentanglement.} \cite{biederman1987recognition} argues that humans recognize a modest set of components invariant over viewing position. Then by learning the gestalt patterns between those parts we are able to recognize the whole of objects. Disentangling the positional encodings of the visual components from their shape and other visual arrays results in inherent viewpoint invariation. \cite{greff2020binding} correctly argues that the lack of explicit object encodings in current neural networks for image processing limits their compositional ability.  They propose that the addition of dynamically binding the neurons would enable segregation, representation, and composition of objects. While we have explored the dynamic binding approach in Dynamic Capsules \citep{sabour2017dynamic}, EM Capsules \citep{hinton2018matrix} and Stacked Capsule Autoencoders \citep{kosiorek2019stacked}, learning the basic components with disentangled position encodings was left to be explored in FlowCapsules \citep{sabour2021unsupervised} in Chapter~\ref{chap:flowcaps}.

\end{itemize}

\section{Chronology and Outline of The Thesis}
This thesis covers three main chapters that represent my research projects spanning from 2020 to 2024.  Each chapter provides the introduction and discussion based on the literature available at the time that the
work was carried out and published. Each of the published work provided novel and even ground breaking work at the time. The provided novel techniques has been build upon by the community and further developed into recent techniques. As such, the chapters are not significantly re-written to reflect the state of the art and relevancy of the research at their time of publication.
Therefore, each chapter has its own introduction to the specific problem being addressed, alongside the background and related work followed by the proposed novel method. Each chapter also provides the experimental results and discussion on the task at hand.

\paragraph{Chapter~\ref{chap:flowcaps}: Unsupervised Part Representation by Flow Capsules} was published at \textit{ICML 2021}. FlowCapsules provides a novel technique for part based embedding learning based on motion explanation. This work was among the pioneers for unsupervised object centric learning incorporating video frames while remaining applicable to images. The results show that not only the model is able to segment the scene into rigid parts without any supervision, but also it can reason about the occluded sections of those parts as well.

\paragraph{Chapter~\ref{chap:robustnerf}: RobustNeRF: Ignoring Distractors with Robust Losses} published at \textit{CVPR 2023} tackles the problem of transient objects during a casual capture for 3D modelling. NeRF as technique designed for synthesizing novel views of complex scenes by optimizing an underlying continuous volumetric scene function using a sparse set of input views relies on photometric consistency of the said input views. During a casual capture this assumption breaks easily by outlier moving objects as well as shadows. In this work we propose an unsupervised method for calculating segmentation masks of the transients parts of the scene. Based on the segmentation masks a robust loss is then proposed which successfully ignores the outlier objects and restores the photometric consistency of the underlying NeRF model. As such the resultant 3D rendering would be sharp and artifact free.

\paragraph{Chapter~\ref{chap:spotlesssplats}: \methodname: Ignoring Distractors in 3D Gaussian Splatting} is accepted at \textit{ACM TOG Journal 2025} and presented at \textit{SigGraph 2025}. \methodname focuses on the same problem as RobustNeRF but with a 3D Gaussian Splatting as the underlying method. 3D Gaussian splatting explains a 3D scene as an aggregation of 3D Gaussians. As it learns the Gaussian centers and colors it also needs to densify and prune such Gaussians. These densification and pruning steps result in radical and abrupt changes in the loss manifold. Therefore, we propose a new pruning regularizer to accommodate a robust optimization. Furthermore, \methodname improves the unsupervised masking technique by incorporating semantic feature of pixels into training a mask generation MLP. \methodname also shows that even a simple clustering of the semantics can improve the object mask heuristics of RobustNeRF significantly.

    \chapter{Related Works}

Each of the research projects discussed in this thesis were providing significant novel contributions to the literature at their time of publication. As such, each chapter provides a brief related work section which is a snapshot of the literature before the publication. With the rapid development of the field in the past four years, we find it prudent to provide an expanded literature review which includes up to date advances. In what follows we will discuss an expanded background as well as the current (at the time of thesis) state of the art in object centric and 3D reconstruction areas.

\section{Object-Part Representation Learning}
One approach to introduce objectness into neural networks is parsing objects into a hierarchy of parts. Incorporating {\em part-whole hierarchies} as in Transforming auto encoders \citep{hinton2011transforming}, improves out of domain generalization while promising scalability. If the basic parts of an object are detected in an image along with their positions, it suffices to learn part-whole relationships to compose them into higher level parts and ultimately a whole object. The advantage of this approach is mainly its inherent out of domain viewpoint generalization. If the input image is transformed, i.e., captured from a new viewpoint, the same part-whole relationship still holds to transform a part pose into a higher level part pose. In such a system, the part poses (hidden layer neurons) are viewpoint equivariant and the part-whole relationships (weight matrices) are viewpoint invariant. Each part pose is named a {\em capsule}. 
A layer consists of several capsules. Each capsule contains the instantiation parameters of a part, including its position. \cite{hinton2011transforming} proposes an auto-encoder to learn the basic part capsules from input images. If the capsule parameters hold the part position, then affinely transforming an image should result in applying the same affine transformation to the capsule parameters. The transforming auto encoder is trained to reconstruct a transformed variant of the input image. The same transformation is applied to the capsule parameters in the last layer. \cite{hinton2011transforming} show that training an auto-encoder in such a manner will result in capturing the spatial coordinates of the object position in the last layer capsule.

Dynamic capsules \citep{sabour2017dynamic} extends work on transforming auto-encoders and introduces dynamic routing layers for learning the part-whole relationships connecting part capsules. The main challenge is deciding how part capsules in a layer should be grouped together into a higher level part capsules. In order to find the agreement between the capsule parts, each lower capsule sends their prediction for the higher level capsule poses. The prediction of lower parts that belong to the the same higher part would form a cluster, while the  predictions from the other lower parts would be treated as outliers. Dynamic routing is a simple method for finding the agreement clusters and refine the predictions. The agreement is measured as the cosine similarity. After all the capsule predictions are summed, the contribution of lower capsules are updated based on their dot product (cosine similarity) with the summed vector (capsule instantiation parameter). This procedure can be repeated several times to remove outliers and settle on tight clusters. Our experiments show that even one iteration (no refining) can achieve state-of-the-art image classification results. With further refining iterations, we show that the proposed capsule network is able to generalize to unseen viewpoints. 

EM Capsules \cite{hinton2018matrix}, is a variation on dynamic routing that proposes a proper expectation maximization algorithm for finding the agreement clusters rather than dyanmic routing. Also, we switch from a 2D pose vector to 3D pose matrices for better affine viewpoint generalization and 3D scenes. EM capsules improve the classification results on more realistic datasets and achieve better viewpoint generalization compared to dynamic capsules.

The iterative dynamic routing in previously mentioned capsule networks is computationally expensive and limits their applicability to larger and more realistic datasets. Furthermore, the mentioned capsule architectures are designed for supervised image classification tasks. In Stacked Capsule auto encoders \citep{kosiorek2019stacked}, we address these issues by turning the iterative refinement into optimization of autoencoders. Each pair of consecutive capsule layers in SCAE is an autoencoder, where the higher level part capsules should reconstruct the lower level part capsules. Furthermore, SCAE is an unsupervised method for learning object representations and it achieved the state-of-the-art performance in unsupervised object classification while maintaining good viewpoint generalization.

One issue with previous capsule works is they assume there is one instance of each type of capsule (part or whole) in each receptive field. This assumption is limiting and does not hold in many application domains of interest. Also assigning specific types to capsules limits the scalability of capsule networks. Modeling a dataset with 1000s of object types requires at least that many capsules in each layer. This further limits their scalability and fairness. GLOM \citep{hinton2021represent} proposes a model to mitigate these issues. GLOM proposes to use a transformer like architecture \citep{vaswani2017attention} rather than a CNN based one. This reduces the coupling of part types and capsules. Rather than a specific capsule representing a specific part type, parts in a layer form an island and are not aggregated into a single capsule. GLOM proposes to use a combination of contrastive loss \citep{sohn2016improved} and attention to form the islands. 

The Visual Parser (ViP) prosposed in \citep{bai2021visual} learns two-level, part-whole hierarchies with part prototypes and whole feature maps. They iteratively refine the part-whole embeddings using top-down and bottom-up attention similar to GLOM. As the closest inceptualization of GLOM, they achieve state-of-the-art in object detection.

Canonical capsules  \citep{sun2021canonical} propose a capsule network for 3D point clouds which enables decomposition of objects into semantically meaningful parts and simultaneously learns a canonical coordinate frame for the objects, which enables unsupervised registration and unsupervised classification of ShapeNet \citep{chang2015shapenet} point clouds. Canonical Capsules, like transforming autoencoders, use two transformed variants of an object for training. Architecture wise, it is a siamese PointNet \citep{qi2017pointnet} and computes capsule decompositions through permutation-equivariant attention. Canonical Capsules aggregate the attention masks into semantic keypoint capsules such that the decomposition satisfies the equivariance properties.

In order to learn object representations without supervision, motion is a strong cue for humans \citep{spelke1990principles}. FlowCapsules \citep{sabour2021unsupervised} is an unsupervised method for learning capsule shapes and positions using unsupervised optical flow estimation methods. Each capsule has a canonical shape embedding and a pose transformation. The encoder, given a single image, outputs all the movable capsule representations. Given the proposed capsule representations for two separate frames of a video, the optical flow is rendered using the affine pose transformations between respective capsules. \cite{sabour2021unsupervised} showed that FlowCapsules are able to segment the images and objects into rigid parts. They also show that, FlowCapsules learn an inherent notion of movable objects and can detect their full silhouette in the presence of occlusion.
Details of this work is further discussed in Chapter~\ref{chap:flowcaps}.

\subsection{Object Centric Architectures}

A closely related line of research is slot attention \citep{locatello2020object}, which learns object representations directly rather than decomposing the scene into parts. In slot attention networks  objects embeddings are represented by ``slots''. Slots are initialized to random Gaussian noise and compete with each other as queries. The keys are feature embedding extracted from an image by a CNN. The slot attention module is then refined based on the aggregated attention with a GRU. Since slot attention uses a recurrent network and generates the full object rather than solving for part-whole relations, there is no fixed binding of object type to slots and the model is applicable to set prediction tasks similar to SCAE.

Similar to FlowCapsules, \cite{bao2022discovering} builds upon slot attention to discover objects that can move. They use either ground truth or off the shelf optical flow techniques to supervise slot attention on video frames. Also, since slots don't have explicit pose parameters, they rely on ground truth segmentation of motion maps to reconstruct the next frame. By doing so, they are able to generalize to real world datasets such as KITTI \citep{geiger2013vision}.

Like  \cite{bao2022discovering}, \cite{kipf2021conditional} also proposed a slot attention model for video (SaVi). But rather than just predicting and reconstructing the next frame, a window of frames is processed recurrently by updating the object slots. Since there is no {\em a priori} notion of what the objects are in the SaVi setup, SaVi relies on conditioning cues for initializing the first slots for each task. Therefore, rather than just recognizing the objects that can move, SaVi can learn about task-dependent objects. The training objective can either be optical flow or image reconstruction. \cite{elsayed2022savi++} built upon SaVi by adding per frame random crop augmentation and depth prediction. By using sparse depth signals from LiDAR in the Waymo Open dataset, SAVi++ is able to scale object segmentation of SaVi to a real world dataset. 

\cite{chen2022unsupervised} had a similar architecture to GLOM but uses the off-the-shelf RAFT optical flow network \citep{teed2020raft} to supervise the affinity graph (islands of GLOM). They also iteratively refined the estimated optical flow based on predicted spelke objects to explain away objects that move together. EISEN is able to improve the unsupervised (flow supervised) segmentation of objects in both synthetic and real image sequences.

Object centric learning is closely related to a line of  work on compositional generative scene models~\citep{greff2016tagger,van2018relational,greff2019multi,burgess2019monet,engelcke2019genesis} that represent a scene in terms of a collection of latent variables with the same representational format. An interesting approach is the IODINE~\citep{greff2019multi} model, which uses iterative variational inference to infer a set of latent variables, each describing an object in an image. In each inference iteration, IODINE performs a decoding step followed by a comparison in pixel space and a subsequent encoding step. Related models such as MONet~\citep{burgess2019monet} and GENESIS~\citep{engelcke2019genesis} similarly use multiple encode-decode steps.

Object centric models can be seen as a variant of interacting memory models~\citep{van2018relational,veerapaneni2020entity}, which utilize a set of embeddings and their pairwise interactions to reason about components in the input images (or objects in a video). Common components of these models are a recurrent update function that acts independently on individual embeddings and an interaction function that introduces communication between those embeddings. Typically, the embeddings in these models are fully interchangeable since the recurrent update functions and interaction functions are shared, with the exception of the RIM model~\citep{goyal2019recurrent}, which uses a separate set of parameters for each embedding. Notably, RMC~\citep{santoro2018relational} and RIM~\citep{goyal2019recurrent} utilize an attention module to aggregate information from inputs to the embeddings. In Slot Attention and Capsule Networks, the attention-based assignment is normalized between the embeddings (as opposed to  over the inputs), which introduces competition between the slots/capsules to perform a clustering of the input. 
Expert models~\citep{goyal2019recurrent} are also related to object-based approaches, but they do not fully share parameters between individual experts. This results in the specialization of individual experts to, e.g., different tasks or object types. In Slot Attention for example, slots use a common representational format and each slot can bind to any part of the input.

\section{3D Reconstruction}
The evolution of neural implicit functions for 3D scene representation starts with early work like DeepSDF \citep{Park19cvpr_deepsdf} and OccNet \citep{OccNet2019}, which used implicit functions to overcome limitations of traditional auto-encoders. These methods offered a compact and efficient way to represent 3D scenes compared to meshes or voxels.

However, these early methods struggled with capturing fine details and scaling to complex scenes.  Later work like DeepLS \citep{Chabra20eccv_DLS}, NGLOD\citep{Takikawa21cvpr_nglod}, and ConvOccNets\citep{peng2020convolutional} addressed these issues by dividing the scene into local grids, which allowed for better representation of high-frequency details.

The inverse rendering techniques then answered whether it is possible to supervise using 2D images only. In these techniques the goal is to learn 3D representations from 2D images.  Methods like Deferred Neural Rendering \citep{thies2019deferred}, SRN\citep{Sitzmann19neurips_srn}, and Neural Volumes \citep{neuralvolumes} introduced differentiable rendering techniques to achieve this.

\subsection{Neural Radiance Fields}
Neural Radiance Fields (NeRF) \citep{nerf} combined the above mentioned ideas, using volume rendering and positional encoding to achieve high-quality novel view synthesis from a set of images.  NeRF addressed limitations of previous methods but still had challenges, such as requiring many input views and being limited to static scenes. They employ volume rendering techniques and use an MLP as the transfer function to map 3D points in space to view-dependent color and density. Nerf also has hierarchical aspects since it trains on both coarse and fine scale. Nerf does not face boundary issues like DeepLS since they sample points randomly rather than having a set of fixed positions. A simple MLP however would still suffer from the oversmoothness because of its spectral bias. Nerf alleviates this shortcoming by augmenting the 3D point coordinates with positional encoding at different sinusoidal frequencies. The positional encoding again shows the importance of explicit multiplicative conversions done before passing input to MLP, which is the main idea and reason of success in transformers, too. Also, similar to Neural Volumes, Nerf conditions the rgb generation on the view direction to better handle viewpoint dependent artifacts.

Subsequent work like pixelNeRF \citep{yu2021pixelnerf}, Stereo Radiance Fields \citep{chibane2021stereo}, and SRT \citep{sajjadi2022scene} focused on reducing the number of input images needed for novel view synthesis.  Other research explored multi-resolution approaches (mip-NeRF \citep{mipnerf}, NSVF \citep{liu2020neural}, mip-NeRF-360 \citep{mipnerf360}) to handle scenes with varying levels of detail and unbounded size. Finally, methods for faster training (InstantNGP \citep{ingp}, Plenoxels \citep{fridovich2022plenoxels}, ReLUFields \citep{relufields}) and joint optimization of camera parameters (BARF \citep{barf}, NeRF-~- \citep{wang2021nerf}, GARF \citep{garf}) have also recently become a focus for further research.

\subsection{3D Gaussian Splatting}
3D Gaussian Splatting (3DGS) \citep{Kerbl2023tdgs} has emerged as a significant advancement in the domain of computer graphics, offering substantial improvements over previous methods like Neural Radiance Fields (NeRFs) in terms of rendering speed and quality for novel view synthesis and 3D reconstruction. By utilizing Gaussian ellipsoids for scene representation, 3DGS facilitates efficient and high-fidelity rendering. 

3DGS represents a paradigm shift from neural implicit representations to explicit modeling with Gaussian ellipsoids. This transition offers several advantages:
\begin{itemize}
    
\item Real-time Rendering: 3DGS enables real-time rendering speeds, making it highly suitable for applications such as virtual reality and real-time simulations.
\item High-Quality Reconstruction: The technique enhances the visual quality of novel view synthesis by mitigating artifacts and effectively managing occlusions, surpassing the performance of previous methods.
\end{itemize}

Several innovative techniques have been developed to augment the capabilities of 3DGS, including 
 Mip-Splatting \citep{Yu2024MipSplatting}, which addresses aliasing artifacts by employing a 3D smoothing filter to preserve the maximum frequency of 3D Gaussian primitives, thereby enhancing rendering quality, even in out-of-distribution scenarios.
Techniques like 4D Gaussian Splatting \citep{wu20244d} capture dynamic details and motions within scenes, enabling realistic animations and simulations. For instance, Dynamic 3D Gaussians \citep{luiten2023dynamic} leverages local-rigidity constraints to represent consistent spatial rotations, facilitating dense 6-DOF tracking and reconstruction without correspondence or flow input.
Another interesting augmentation is Gaussian-Dreamer \citep{yi2024gaussiandreamer} which utilize diffusion models to generate detailed 3D models from textual descriptions, demonstrating potential for creative content generation.

The versatility of 3DGS is evident in its wide range of applications. Techniques like Drivable 3D Gaussian Splatting (D3GA) \citep{qian20243dgs} enable real-time rendering of human avatars, crucial for virtual reality, gaming, and digital communication.
On the other hand GS-SLAM \citep{yan2024gs} utilizes 3D Gaussian representations for real-time SLAM applications, significantly enhancing the efficiency and accuracy of autonomous navigation systems.
Finally models like MVSplat \citep{chen2024mvsplat} employ multi-view depth estimation to accurately reconstruct 3D scenes from sparse image sets, enhancing the flexibility and efficiency of 3D modeling.

\subsection{3D Reconstruction in a Casual Capture Setup}

While significant progress has been made in the field of multi-view 3D modeling \citep{nerf, Kerbl2023tdgs}, many existing methods are limited by assumptions of controlled acquisition, such as static scenes and densely captured viewpoints. On the other hand, advances in 3D learning and the widespread availability of 2D and 3D visual data offer new opportunities to overcome these challenges and make 3D modeling more robust, widely applicable, easily accessible, and cost effective.

NeRF excels in controlled and static scenes. In practice such meticulously gathered datasets are not always available. For example given a set of photos of a tourist attraction site such as Eiffel tower (i.e. photo tourism datasets), the ambient light and transient objects such as pedestrians invalidate the NeRF assumption of a static scene. The papers in this section utilize robust losses, bundle adjustment techniques, and trainable embedding vectors to model such variations in the dataset.

There are two major types of photo inconsistencies that present challenges for multi-view 3D modeling:
\begin{itemize}
    \item Appearance of the scenes change over time, usually captured overal several days with different cameras. In this case the model needs to adapt and render consistent novel views of the scene as its appearence changes~\citep{gao2021dynamic,li2021neural,park2021nerfies,li2022neural,wu2022d,liu2023robust,park2023temporal}, \eg, scenes with different camera setups (black and white vs RGB) or environmental effects like weather changes and clouds passing.
    \item A casual capture, usually captured in a single attempt but obstructed with transient objects. In this case the model should exclude the outlier dynamic objects such as pedestrians or shadows of the camera and camera man. 
\end{itemize}

 The second scenario of removing transient objects is closely related to unsupervised segmentation of foreground objects in 3D. As such, we can group the approaches for handling transient objects into two categories:

\begin{itemize}
    \item {\it Generally Pretrained Supervised Segmentation based methods }~\citep{tancik2022block,turki2022mega,rematas2022urban,sun2022neural,liu2023robust,karaoglu2023dynamon} use pre-trained semantic or video segmentation models to identify the distractors. Given the segmentations masks they either pre-process the data or incorporate the oracle segmentation masks.
    These models can produce reliable results in only certain scenarios and have several limitations such as: They require additional priors like the semantic class of transient objects or the temporal relationships of the images as video frames, which are hard to satisfy in practice as it is intractable to enumerate over all possible object classes, and images may be unordered.
    Another limitation is that task independent semantic segmentation cannot distinguish between static and transient objects of the same class. Furthermore, shadows are a common distractor which are not easy to handle as a semantic segmentation class. 
    \item {\it Task based unsupervised methods}~\citep{martin2021nerf,chen2022hallucinated,lee2023semantic,Sabour2023robustnerf,li2023nerf,kim2023upnerf} use hand-crafted heuristics to segment transient distractors from static scenes during training, making themselves more generalizable as they require no prior.
    However, many challenges of unsupervised object segmentation mentioned in the previous sections are present and therefore these masks are difficult to design.
    For example, NeRF-W~\citep{martin2021nerf} observes that the density of transient objects is usually small and uses this to regularize NeRF training. As such it can easily produce foggy residuals with small densities that are not transient objects.
    RobustNeRF~\citep{Sabour2023robustnerf} distinguishes transient objects from static ones through the spike in color residuals. They use a spatially consistent filter to insure the masks cohere to a single object.
    In practice high-frequency details of static objects also lead to high residual areas in the scene, causing RobustNeRF to easily ignore them when dealing with transient distractors.
    \methodname~\citep{sabour2024spotlesssplats} incorporates semantic features to ensure the objectness of the high value residual region while adapting the method to be applicable to the recent 3D modeling technique of Gaussian Splatting.
\end{itemize}

\chapter{Unsupervised Part Representation by Flow Capsules}
\label{chap:flowcaps}

Capsule networks aim to parse images into a hierarchy of objects, parts and relations.
While promising, they remain limited by an inability to learn effective low level part descriptions.
To address this issue we propose a way to learn primary capsule encoders that 
detect atomic parts from a single image.
During training we exploit motion as a powerful perceptual cue for part definition, 
with an expressive decoder for part generation within a layered image model with occlusion.
Experiments demonstrate robust part discovery in the presence of multiple objects, cluttered 
backgrounds, and occlusion. The  part decoder infers the underlying shape 
masks, effectively filling in occluded regions of the detected shapes.
We evaluate FlowCapsules on unsupervised part segmentation and unsupervised image classification.

\section{Introduction}
\label{sec:intro}

Humans learn to perceive shapes in terms of  parts and their spatial relationships~\citep{Hoffman2001}.
Studies show that infants form early object perception by dividing visual inputs into units that move rigidly 
and separately~\citep{spelke1990principles}, and they do so in a largely unsupervised way.
Inspired by this and recent work on part discovery, we propose a self-supervised way to learn visual part descriptors for Capsule networks~\citep{Hinton2011tae}.

Capsule networks represent objects in terms of primary part descriptors, in a local canonical frame, 
and coordinate transformations between parts and the whole.  As a result of their architecture, 
they are robust to various challenges, including viewpoint changes and adversarial attacks.
Stacked capsule network architectures (SCAE)~\citep{kosiorek2019stacked} have shown promising results 
on a number of simple image datasets. Nevertheless, because they are trained with an image reconstruction 
loss, foreground-background separation and part discovery in cluttered images remain challenging.

\begin{figure}[t]
\vspace*{-0.15cm}
\begin{center}
        \begin{overpic}
            [width=0.9\linewidth]            
            {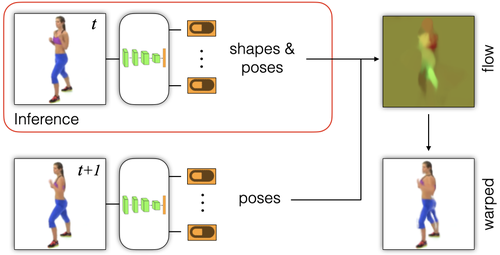}
        \end{overpic}
\end{center}
\vspace*{-0.1cm}
\caption{
\textbf{Self-supervised training for learning primary capsules:} 
An image encoder is trained to decompose the scene into a collection of \textit{primary capsules}.
Learning is accomplished in an unsupervised manner, using flow estimation from capsule shapes and poses as a \textit{proxy} task.
}
\label{fig:teaser}
\vspace*{-0.3cm}
\end{figure}

This chapter introduces a way to learn encoders for object parts (aka., primary capsules) 
to address these challenges.
The encoder takes as input a single image (see Fig.\ \ref{fig:teaser}), but 
for training part discovery, it uses motion-based self-supervision~\citep{Bear-NeurIPS20,Mahendran2018}.
Like the classical literature on perceptual organization and common fate in Gestalt psychology~\citep{spelke1990principles,WagemansPsychBull2012}, we exploit the fact that regions 
of the image that move together often belong together.  
This is a strong perceptual cue that facilitates foreground-background segmentation
and part discovery, and allows one to disentangle texture and other aspects of appearance from shape.

The proposed part encoder  captures the underlying part shapes, their relative poses, 
and their relative depth ordering
(see Fig.\ \ref{fig:archTest}).
The introduction of depth ordering is particularly useful in order to account for occlusion, 
as it is in layered motion models \citep{WangAdelson1994}.
In this way, learning aggregates information about shape over many images,
even though a given part may rarely be visible in its entirety in any single frame.  
In essence, the model prefers simple part-based descriptions, where many variations in appearance
can be explained by a coordinate transform or by occlusion, rather than by changes in shape per se.

We demonstrate the FlowCapsules approach on several datasets showcasing challenges due to 
texture, occlusions, scale, and instance variation. We compare FlowCapsules to recent 
related work including PSD \citep{xu2019unsupervised} and R-NEM \citep{van2018relational}, 
where part masks and dynamics are learnt using motion.
FlowCapsules provide unsupervised shape segmentation, even in the face of texture 
and backgrounds, outperforming PSD \citep{xu2019unsupervised}.
FlowCapsules also provide a depth ordering to account for occlusion, with the added benefit that
part inference yields shape completion when parts are partially occluded. 

We also report unsupervised classification of images using FlowCapsules part embeddings. 
We compare our results on several datasets with different challenges against SCAE~\citep{kosiorek2019stacked}. 
Experiments show that FlowCapsules consistently outperform SCAE in unsupervised object classification, 
especially on images with textured backgrounds.

\section{Related Work}

Given the vast literature of part-based visual representations,
we focus here only on the most closely related recent work.

Transforming autoencoders \citep{Hinton2011tae} introduced capsule networks. 
\citet{Sabour2017capsule} revisited the capsule concept and introduced capsule hierarchies 
for object classification, and subsequent work has produced improved routing algorithms \citep{Hinton2018capsule,hahn2019self,ahmed2019star}.
Nevertheless, learning primary capsules from images has remained largely untouched. 
An analogy to text understanding would be a language with a well defined 
grammar and parser, but no good definition or representation of words. 
We introduce a technique for learning primary capsules to address this shortcoming.

Unsupervised capsule learning with an \textit{image reconstruction} loss for part discovery
has been explored by \cite{kosiorek2019stacked} and \cite{Rawlinson2018sparsecaps}.
Several works learn capsule autoencoders for 3D objects from point clouds~\citep{srivastava2019geometric,zhao20193d,sun2020caca}.
But with the exception of capsule models trained with class labels \citep{Hinton2018capsule} 
or segmentation masks \citep{Lalonde2018capsule, Duarte}, previous methods struggle with natural images. 
Object-background discrimination with cluttered, textured scenes is challenging for an image
reconstruction loss. With self-supervised training and visual motion, FlowCapsules achieve
part discovery without ground truth labels or segmentation masks.

Recent approaches to object-centric learning, e.g., MONet \citep{burgess2019monet}, IODINE~\citep{greff2019multi}, 
and Slot-attention~\citep{locatello2020object}, focus on learning object representations via image reconstruction.
Beyond the need to reconstruct image backgrounds, they require iterative refinement for
symmetry breaking and forcing scenes into slots. 
In contrast, FlowCapsule learning relies on reconstruction of the flow rather than the image,
and with motion as the primary cue, scenes are decomposed into parts without needing iterative refinement. 
Most recently, \citep{Bear-NeurIPS20, veerapaneni2020entity} extend such networks to incorporate motion,
but still rely on iterative refinement. 
FlowCapsule encodings further disentangle shape and pose, enabling shape completion during partial occlusion.

FlowCapsules currently represent 2D objects, reminiscent of layered models but with a feedforward encoder.
Classical layered models \citep{WangAdelson1994,SpritesCVPR2001} used mixture models and assigned pixels to layers independently, often failing to capture the coherence or compactness of object occupancy.
Some methods use  MRFs to encourage spatial coherence \citep{WeissCVPR1997}.  Others  
enforce coherence via local parametric masks~\citep{Jepson2002}.

Visual motion is well-known to be a strong cue for self-supervised learning.
For example, \cite{vijayanarasimhan2017sfm} learn to infer depth, segmentation, and relative 3D motion from 
consecutive frames using self-supervised learning with photometric constraints.

\begin{figure*}[th]
\begin{center}
\begin{overpic}
    [width=0.995\linewidth]
    {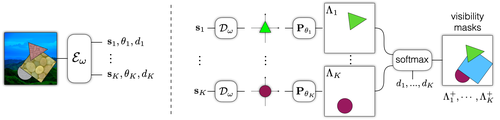}
\end{overpic}
\end{center}
\vspace*{-0.1cm}
\caption{
\textbf{Inference architecture.}
(left) The encoder $\encoder$ parses an image into part capsules, each 
comprising a shape vector $s_k$, a pose $\theta_k$, and a scalar depth value $d_k$. 
(right) The shape decoder $\mathcal{D}_\omega$ is an implicit function.  
It takes as input a shape vector, $s_k$, and a location in canonical 
coordinates and returns the probability that the location is inside the shape.
Shapes are mapped to image coordinates, using $\theta_k$, and layered 
according to the relative depths $d_k$, yielding visibility masks.
}
\label{fig:archTest}
\vspace*{-0.2cm}
\end{figure*}

These and related methods use \textit{optical flow} or multiple frames as an input.
FlowCapsules use video frame pairs during training, but the part encoder (see Fig.\ \ref{fig:archTest}),
takes as input a \textit{single} frame. In essence, it learns to decompose images into {\em movable} objects.

S3CNNs \citep{Mahendran2018} take a similar approach, but do not learn per-part shape encoders or coordinate frames. Rather, they learn to group pixels using patch-wise affine flow, rather than expressing flow in terms of coherent parts and their coordinate frames.
A closely related method is PSD~\citep{xu2019unsupervised}, which 
uses optical flow to learn hierarchical part-based models of shape and dynamics in a layered image model.  
It trains a VAE flow encoder and an image encoder to predict the next frame. 
Both PSD and S3CNNs require ground truth flow during training and lack an explicit canonical 
part descriptor like FlowCapsules.
There also exist related methods applied to point could data; e.g.,
SE3Net \citep{byravan2017se3nets} uses a part-based representation, taking a point cloud and an action vector as input and predicts the point cloud in the next frame.

KeypointNet \citep{keypointNet2019} addresses the problem of keypoint discovery. One  might  view  Flow-Capsules  as  a  generalization  from  a  sparse to a dense  setting, and from a single object to multiple objects, 

Our work is also related to generative shape models.
\citet{HuangMurphyICLR16} learn parts in a layered model with depth order and occlusion.  
Given an image, \textit{variational} inference is used to infer shape and foreground/background separation. 
FlowCapsule encoders, by comparison, are trained as auto-encoders and are therefore easier to learn.
Several recent papers learn generative models that \textit{disentangle} shape and deformation ~\citep{SkafteHauberg2019,deng2021generative}.
FlowCapsules disentangle shape and transformations from canonical to image coordinates. In doing so they
decompose shapes into multiple near-rigid parts with occlusions. 
FlowCapsules thereby disentangle shape at a finer granularity. 
Also, \citet{SkafteHauberg2019} and \citet{deng2021generative} use an  image reconstruction loss, much like SCAE, while FlowCapsules 
only encode shape silhouettes, which simplifies training and the disentangled representation.

\section{Model}

Our goal is to learn an encoder that parses images of familiar shapes into parts.
To facilitate training, and downstream tasks, we also learn a decoder capable of 
generating segment masks for the parts in the image.
Below we describe the form of the proposed capsule encoder and the mask decoder.
We then describe the objective and training procedure.

\begin{figure}[t]
\begin{center}
        \begin{overpic}
            [width=0.83\linewidth]
            {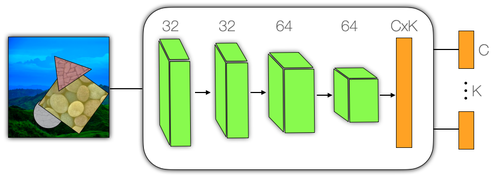}
        \end{overpic}
\end{center}
\vspace*{-0.1cm}
\caption{
\textbf{Encoder architecture.} The encoder comprises convolution layers with ReLU activation,
followed by down-sampling via $2\! \times \!2$ AveragePooling. 
Following the last convolution layer is a \texttt{tanh} fully connected layer,
and a fully connected layer grouped into $K$, $C$-dimensional capsules.
}
\vspace*{-0.05cm}
\label{fig:archEncode}
\end{figure}

\paragraph{Image encoder}
The capsule encoder $\encoder$, with parameters $\params$, encodes a given image as a collection of $K$ primary capsules.
The architecture we propose is depicted in Figure~\ref{fig:archEncode}.
Each capsule, $\capsule_k$, comprises a vector $\shape_k$ that encodes the \textit{shape} of the part,
a \textit{pose} vector $\pose_k$, and a depth scalar $\depth_k$:
\begin{equation}
    \encoder(I) = \{ \capsule_0, \dots, \capsule_k \}, \quad \capsule_k = (\shape_k, \pose_k, \depth_k) ~.
\end{equation}
Capsule shapes are encoded in a canonical coordinate frame.  
The scalar $\depth_k$ specifies relative inverse depth (larger for foreground objects).
The pose vector specifies a mapping from part-centric coordinates ${\bf v}$ to image coordinates $\pixel$
(or scene coordinates more generally), i.e., $\pixel = \T_{\pose_k} {\bf v}$.

As we focus on planar layered models with depth $\depth$, we define $\T_{\pose_k}$ to be a conformal map.  
Accordingly, let $\pose_k {\in} \Real^4$, where $[\pose_k]_{0,1}$ represents the translation,  $[\pose_k]_{2}$ is the rotation angle, and $[\pose_k]_{3}$ is the change in scale. 
More concretely (subscript $k$ is dropped for readability):
\begin{align}
    \T_{\pose} \,  = \,
    \begin{bmatrix}
   \pose_3\cos(\pose_2) & -\pose_3 \sin(\pose_2) & \pose_0\\ 
   \pose_3\sin(\pose_2) & \pose_3 \cos(\pose_2) & \pose_1 \\
   0 & 0 & 1
   \end{bmatrix}
\end{align}
Taken together, 
$\capsule_k {\in} {\Real}^C $, where $\pose_k {\in} \Real^4$, $\depth_k {\in} \Real$, and therefore $\shape_k {\in} \Real^{C-5}$.

\begin{figure*}[t]
\begin{center}
        \begin{overpic}
            [width=0.78 \linewidth]
            {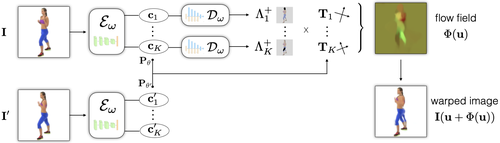}
        \end{overpic}
\end{center}
\vspace*{-0.1cm}
\caption{
\textbf{Self-supervised training} -- 
Training uses a \textit{proxy} motion task in which the capsule encoder is applied to a pair of successive 
video frames, providing $K$ primary capsule encodings from each frame.
Visible part masks, $\Lambda^+_k$,  and their corresponding poses, $ {\T}_{\pose} $, determine a flow field 
$\flow$ that is used to warp image $\image$ to predict $\next\image$ in the loss
$\mathcal{L}_\text{render}$ in (\ref{eq:loss_render}).
}
\vspace*{-0.2cm}
\label{fig:archTrain}
\end{figure*}

\paragraph{Mask decoder}
A mask decoder facilitates self-supervised learning of the encoder, as well as downstream segmentation tasks.
It allows one to visualize parts and connect them to image observations.
As depicted in \Figure{archTest}, the mask decoder $\masknet$ generates an object silhouette~(or mask) 
in canonical coordinates, which is then mapped to image coordinates, incorporating occlusion and visibility.

\begin{figure}[t]
\vspace*{-0.2cm}
\begin{center}
        \begin{overpic}
            [width=0.71\linewidth]
            {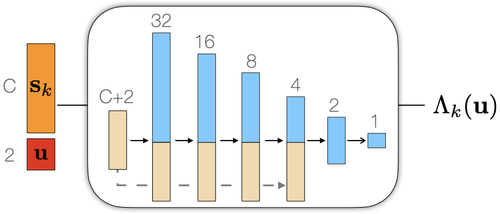}
        \end{overpic}
\end{center}
\vspace*{-0.35cm}
\caption{
\textbf{Decoder architecture.}  A neural implicit function \citep{imnet} is used to represent part masks.
An MLP with SELU activations \citep{klambauer2017self} takes as input a shape vector $s$ and a pixel position ${\bf u}$. 
Applied to a pixel grid, it produces a logit grid for the mask.
}
\vspace*{-0.25cm}
\label{fig:archDecode}
\end{figure}

Our current decoder architecture is depicted in \Figure{archDecode}.
The mask decoder, given the latent code $\shape_k$, represents the part shape in a canonical coordinate frame, $ \masknet({\bf v}; \shape_k) $.  
This is then mapped into image coordinates according to the pose vector~$\pose_k$, 
yielding the shape mask~$\Lambda_k$ in the image frame:
\begin{align}
    \Lambda_k (\pixel) &~=~ 
    \masknet( \T^{-1}_{\pose_k} \pixel \, ;\, \shape_k)  ~,
\end{align}
where the map $\T_{\pose_k}$ has parameters $\pose_k$.
We also note that $\Lambda_{k}$ is a \textit{function} of spatial position and a latent code~\citep{imnet,OccNet2019}, but unlike previous work, our encoder disentangles individual part shapes and their poses with respect to canonical coordinates.

{\bf Occlusion:} 
With opaque objects, parts will not always be visible in their entirety.
To account for occlusion, part masks are layered according to their depth order, 
thereby determining the visible portion of each part in a given image.
To ensure differentiable image formation, enabling gradient-based learning, we treat the scalar $\depth_k$ as a logit, and apply a softmax across the logits (depths) of all parts at every pixel to generate the visibility masks~\citep{gadelha2019shape}; see~Fig.\ \ref{fig:archTest}.
The visible portion of the $k$-th part is given by 
\begin{equation}
    \Lambda^+_k(\pixel) = \frac{e^{d_k\Lambda_k(\pixel)}}{\sum_{k'} e^{d_{k'} \Lambda_{k'}(\pixel)}}
    \label{eqn:lambdaplus}
\end{equation}
As the gap between the largest $\depth_k$ and other values grows, the softmax approaches the argmax,
ideal for opaque layers.

A typical auto-encoder might reconstruct the image in terms of these masks, to formulate an image reconstruction loss.
The problem with such an approach is that the encoder would also need to encode other properties of the images, 
such as texture, lighting and the background, with pixel level accuracy.
To avoid this problem, here we aim only to learn an encoder for the part shapes, positions and depth layering.
To this end we consider a form of self-supervised learning that  relies on primarily on 
motion (optical flow) between consecutive frames in video.
The use of flow provides a strong image cue for the segmentation of parts, {\it without} the need to 
model texture, lighting and other fine-grained properties tied to appearance.

\section{Self-Supervised Learning}
Training the capsule encoder exploits motion as a visual cue for separating objects 
and their parts from the immediate background.
To that end, we assume that the training data comprises pairs of adjacent video frames.
Given an image pair, the encoder provides an \textit{ordered} set of capsules for each of the two images.
The poses from corresponding capsules and their visibility masks then determine a deformation that is used to warp one frame to predict the other. 
This allows use of brightness constancy and other common objectives in optical flow estimation as a self-supervised training loss.

In more detail, let the two images of a training pair be denoted $\image$ and $\next\image$.
As shown in \Figure{archTrain}, the capsule encoder extracts an ordered set of capsules from each image.
The part capsules are denoted $\capsule_k {=} (\shape_k, \pose_k, \depth_k)$ and $\capsule_k^\prime {=} (\shape_k^\prime, \pose_k^\prime, \depth_k^\prime)$, for $k \in \{ 1 , ... , K \}$.
From corresponding part capsules we then compute the predicted  flow $\flow$ from the capsule poses,
yielding a mapping $\F_k$ from one image to the next,
\begin{eqnarray}
\F_k ~=~ ~\T_{\next\pose_k} \circ (\T_{\pose_k})^{-1}   ~.
\label{eq:partTransform} 
\end{eqnarray}
This transform maps image locations in $\image$ to the canonical coordinate 
frame of part $k$, and then into the next frame~$\next\image$.
When combined with the layered visibility masks, this provides the flow field:
\begin{eqnarray}
\flow(\pixel  \, |\, \encoder(\image),\encoder(\next\image) ) ~=~ \sum_{k=1}^{K}  ~
\underbrace{\Lambda^+_k(\pixel)}_{\text{visibility}} \, %
\underbrace{\left [\F_k(\pixel)-\pixel\right]}_\text{~flow of $k$-th capsule} 
\label{eq:flow_basic}
\end{eqnarray}
where $ \pixel \in [-1,1]^2 $ denotes 2D normalized image coordinates.
Note that the use of $[\F_k(\pixel)-\pixel]$ in \eq{flow_basic} ensures that the 
generation of an \textit{identity} flow is the easiest prediction for the network 
$\F_k(\pixel)$ to make (like a residual connection).

Given the estimated flow between a given training pair, we warp the pixels of $\image$ according 
to $\flow$, providing a prediction for $\next\image$. 
Then we optimize an L2 brightness constancy loss on the residual errors between
our warped version of the first frame and the second frame,
\begin{equation}
\mathcal{L}_\text{render} \,=\, \expect{\pixel \sim [0,1]^2}
{\| \,  \image(\pixel + \flow(\pixel)) - \next\image(\pixel) \, \|^2_2} \, ~,
\label{eq:loss_render}
\end{equation}
where we abbreviated $\flow(\pixel\, |\, \encoder(\image), \encoder(\next\image))$ by $\flow(\pixel)$
for notational simplicity.

We also exploit two simple but effective regularizers on flow and the canonical shape representation. 
They are useful as we do not make use of ground truth segmentation masks or flow fields during training.
The first regularizer, $\mathcal{L}_\text{smooth}$, is a smoothness term often used in 
optical flow estimation \citep{DerpanisECCV2016} to enhance gradient 
propagation through larger movements and regions with negligible image gradients:
\begin{align}
\mathcal{L}_\text{smooth} \,=\, \left\| \frac{\partial \flow}{\partial u_x}, \frac{\partial \flow}{\partial u_y} \right\|^2_2 ~.
\label{eq:loss_smooth}
\end{align}

The second regularizer encourages part shapes to be centered at the origin in the canonical coordinate frame; i.e.,
\begin{equation}
\mathcal{L}_\text{center} = 
\tfrac{1}{K} \sum_{k=1}^{K} \frac{\sum_{\bf v} \| {\bf v} \, \Lambda_k({\bf v})  \|^2_2 }{\sum_{\bf v'} \Lambda_k({\bf v'})} 
\label{eq:loss_center}
\end{equation}
Keeping parts centered at $(0,0)$ improves the inference  of rotations. 
For example, a part located far from the origin can easily be projected outside the image
during training.
Keeping it near the center tends to produce a smoother loss function.
The final loss is a weighted sum of the render loss and the two regularizers.

\section{Experiments}

We evaluate FlowCapsules on images with different dynamics, shapes, backgrounds and textures. 

\paragraph{Geo}
For this synthetic dataset, we use the same code and setup as \cite{xu2019unsupervised},
generating 100k images for training, 1k for validation, and 10k for testing.
Images have different background colors, with geometrical shapes (circle, triangle, square) 
of various colors, scales and positions. Objects in Geo undergo translation from frame to frame.

\paragraph{Geo$^+$}
This variant of Geo incorporates  natural image backgrounds (random images from ImageNet~\citep{deng2009imagenet}),
and textured foreground shapes. 
Textures are random samples from the Brodatz dataset~\citep{picard1993real}.

\paragraph{Exercise}
This dataset contains natural images of trainers demonstrating exercises, with articulated and out of plane motion (used by \citet{xu2019unsupervised}).
It has 49356 pairs of images for training, extracted from 20 exercise demo videos. 
The test set has 30 images, for which \citet{xu2019unsupervised} provided ground truth segmentation masks.

\vspace*{-0.1cm}
\paragraph{Experimental setup} 
Models are trained using the Adam optimizer \citep{Kingma2014adam} with a fixed learning rate of $1e{-}4$ for 150 epochs. 
We use $C{=}32$ and $K{=}8$ for Geo models and $C{=}16$ and $K{=}16$ for Exercise model.
Regularization constants for $\mathcal{L}_\text{center}$ and $\mathcal{L}_\text{smooth}$
are $1e{-}2$ and $1e{-}4$.
To calculate the intersection-over-union (IoU) performance measure on visibility masks, 
we normalize and then threshold the masks at $0.5$ to get a binary mask.

\begin{figure}[t]
    \centering
    \includegraphics[width=0.7\linewidth]{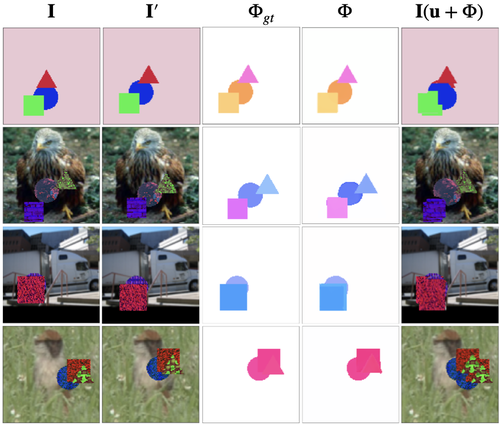}
    \caption{Estimated flows and predicted next frames on training data from \texttt{Geo}~(first row) and \texttt{Geo}$^+$~(rows 2-- 4).
    }
    \label{fig:geo_flow}
\end{figure}

\subsection{Estimated Part Motion}

To verify that the model estimates flow effectively in an unsupervised manner we first inspect 
the quality of the flow inferred by FlowCapsules after training on each dataset.

\Figure{geo_flow} shows estimated flow $\flow$ alongside the ground truth $\flow_{gt}$  for {\em training} 
image pairs from Geo and Geo$^+$.  The flow is accurate for both datasets.
Comparing the warped version of the first frame $I$ (last column) with the other frame $I'$ 
(second column), one can appreciate some of the challenges in unsupervised flow estimation. 
Because our prediction of~$\next\image$ using~$\flow$ does not account for unoccluded pixels,  $\mathcal{L}_\text{render}$ is not expected to reach $0$. 
We note that while the model uses conformal transformations from frame to frame, these datasets only have
translation; for these data our model correctly estimates zero rotation and unit scale.

\begin{figure}[t]
    \centering
\includegraphics[width=.635\linewidth]{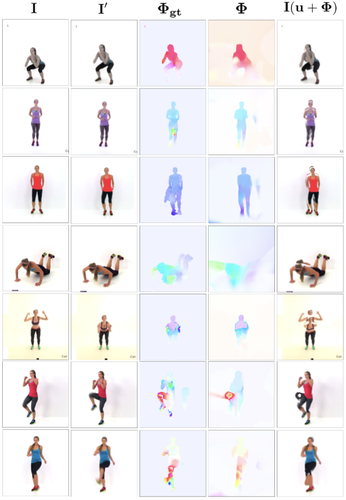}
    \caption{
    Estimated flows and predicted frames on randomly selected images from the \texttt{Exercise} \textit{validation} set. 
    Approximating articulated motion with conformal maps yields reasonable flow fields. 
    The goal is not the best possible flow estimation, but rather, 
    as long as different parts have different flow estimates, our encoder is be able 
    to learn the correct part decomposition.
    }
    \label{fig:ex_flow}
\end{figure}

\Figure{ex_flow} shows examples of  model flow estimates for the Exercise dataset.
The true flow here reflects the articulated motion of the people, and it is notable
that the parts here are much smaller than those in Geo/Geo$^+$.
Although the estimated flows are somewhat blurred, they still capture the movements  
of the different parts reasonably well, even though the model is limited to conformal deformations 
from one frame to the next.

\subsection{Unsupervised Part Segmentation}

One effective way to evaluate FlowCapsules is to see how well it learns to 
decompose a single image into its \textit{movable} parts.  
We view this as an unsupervised part segmentation task
and we note that, while trained on image pairs, inference is performed on a \textit{single} 
test image, yielding part shapes and a coordinate transform for each part.
Conversely, methods relying on optical flow only generate masks for parts \textit{in motion}, 
as these models effectively build masks by \textit{segmenting} the flow \citep{xu2019unsupervised}.

\begin{figure}[t]
    \centering
    \includegraphics[width=0.7\linewidth]{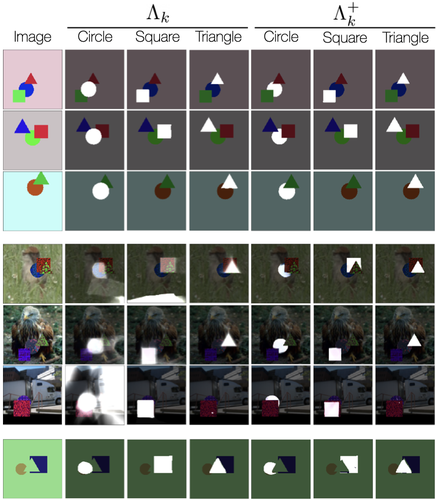}
    \vspace*{0.2cm}
    \caption{Inferred FlowCapsule shapes and corresponding visibility masks on Geo (rows 1--3), and Geo$^+$ (rows 4--6).
    The third row for each dataset shows an instance with only two objects present, so one mask is empty.
    The last row shows an interesting case in which the 
    triangle is detected by the encoder even though it shares the color of the background, reminiscent of subjective contours \citep{Kanizsa1976}.
    }
    \label{fig:geo_masks}
\end{figure}

\begin{figure*}[t]
    \centering
    \includegraphics[width=0.975\linewidth]{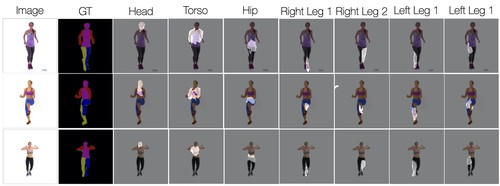}
    \caption{The ground truth segment masks along with sample FlowCapsule masks $\Lambda_k^+$ on Exercise test data.}
    \label{fig:ex_masks}
\end{figure*}

\paragraph{Qualitative analysis on \texttt{Geo} (\Figure{geo_masks})}
Masks shown in Fig.\ \ref{fig:geo_masks} demonstrate that FlowCapsules learn to detect
meaningful part shapes (e.g., a triangle or circle). Indeed, the model tends to explain 
a given image in terms of a small number of generic shapes and occlusion of overlapping parts, effectively performing {\em part completion} \citep{Hoffman2001}.
This is particularly interesting because the model does not include an explicit regularizer 
that encourages the model to learn a specific number of shapes, or sparsity in the space of shapes.
One might not expect the model to learn to represent the entire shapes (e.g.~an entire circle). For example, one might have expected the model to have learned a large number 
of different shapes from which the observed shapes are constructed, especially with occlusion
where the entire shape is often not observed in a single image.
Nevertheless, the model opts to explain the images with relatively few parts, and hence the 
capsule masks tend to cover all the pixels of a shape in the Geo dataset. 
This can be attributed to the architecture we use for mask decoders, and the inductive 
bias of MLPs in generating low-frequency functions~\citep{tancik2020fourier,atzmon2020sal,basri2020frequency,rahaman2019spectral}.

\begin{table}[t]
\begin{center}
    
\begin{tabular}{@{}llccc@{}}
\toprule
                          &           & R-NEM & PSD  & Flow Capsules \\ \midrule
\multirow{4}{*}{Geo}      & Circle    & 0.54 & 0.93 & \textbf{0.94} \\
                          & Square    & 0.56 & 0.82 & \textbf{0.98} \\
                          & Triangle  & 0.58 & 0.90 & \textbf{0.98} \\
                          & All       & 0.56 & 0.88 & \textbf{0.95} \\ \midrule
\multirow{4}{*}{Exercise} & Torso     & 0.32 & 0.57 & \textbf{0.62} \\
                          & Left Leg  & 0.29 & 0.37 & \textbf{0.59} \\
                          & Right Leg & 0.23 & 0.34 & \textbf{0.54} \\
                          & All       & 0.28 & 0.43 & \textbf{0.58} \\ \bottomrule
\end{tabular}
\caption{
\textbf{Quantitative / Segmentation -- }
IoU of inferred segment masks w.r.t ground truth on Geo and Exercise data.}
\label{tab:iou}

\end{center}
\end{table}

Geo is synthetic, so correct masks for the full shapes are known. 
Since FlowCapsules provide both the part shapes, via $\Lambda_k$, and the
associated visibility masks $\Lambda_k^+$ taking occlusion into account, 
we can compare $\Lambda_k$ to the full ground truth shapes.  
One can then quantify performance using the usual intersection over union (IoU) measure.
FlowCapsules achieves segments with an IoU of \textbf{0.96} on all the shapes, circle, square, 
and triangle  (see Table \ref{tab:iou}). 
This result indicates how well the model encodes the full shape, effectively filling in 
occluded portions of shapes in test images.

\paragraph{Qualitative analysis on \texttt{Exercise} (\Figure{ex_masks})}
On the Exercise dataset, FlowCapsules learn to segment the body into \textit{roughly} rigid parts.
Fig.\ \ref{fig:ex_masks} illustrates the segmentation masks of some of the part capsules.
The masks for individual capsules consistently capture the pixels associated with meaningful 
body parts, such as the head or right leg, regardless of the input image. 
As such, capsule identities are tied to semantic parts rather than spatial position. 
We also note that the capsules tend to delimit parts at joints, and separate the hips (lower torso) 
from the legs and from the upper torso, even though we do not use a kinematic prior.

\begin{figure}[t]
\centering
\includegraphics[width=.65\linewidth]{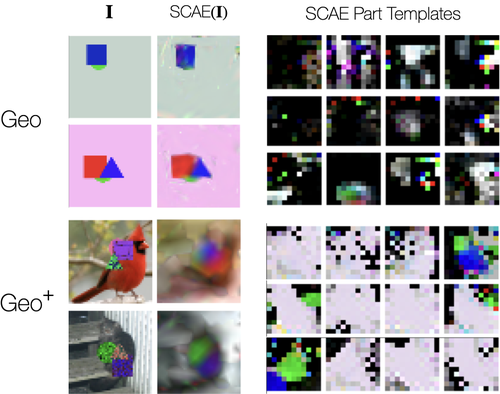}
\caption{(left) SCAE reconstructions after training on Geo and Geo$^+$. 
(right) The learned part templates. 
SCAE approximately reconstructs the image but the part templates are not coherent parts. 
Comparing Geo$^+$ and Geo, the learned parts loose all shape information to enable reconstructing the color, 
texture and background in the images. 
}
\label{fig:scae}
\end{figure}

\vspace*{-0.1cm}
\paragraph{SCAE (\Figure{scae})}
The most relevant prior work to FlowCapsules vis-a-vis part discovery is SCAE~\citep{kosiorek2019stacked}.
\Figure{scae} shows part templates and image reconstructions generated by SCAE.
Even in simple cases without backgrounds or texture, SCAE fails to segment images into meaningful parts, unlike FlowCapsules, Fig.~\ref{fig:geo_masks}.
This failure becomes markedly worse for Geo$^+$ when object textures and background are added. FlowCapsules are able to detect and focus on foreground objects with coherent part masks. But SCAE has to reconstruct the background, so the part shapes become blobs.

\paragraph{SCAE Training Details}
While comparing FlowCapsules against SCAE, we updated SCAE training at various spots to make it more
suitable for Geo and Geo$^+$ datasets. Here we detail these changes. First, We resized input images to
$48 \times 48$ for memory reasons. Second,
we added the option of inferring the background color as well as background image using a two level MLP.
Similarly, we added the option of adding color or texture to each template. To enable colorization and texturization based on input image, the primary capsule features are passed to the template decoder. The color/texture is generated by a 2 layer MLP (32 dimensional hidden representation). The original fixed templates are used as masks and multiplied to the output of the color/texture MLP. 

For generating a background template, we use the second to last hidden representation of the primary encoder as the image embedding. We pass the image embedding through a 2 layer MLP (32 dimensional hidden representation). We mix this background template with a presence probability of $0.5$.

All the other parameters, including training schedule is kept the same as the original SCAE.

\vspace*{-0.1cm}
\paragraph{PSD and R-NEM (\Table{iou})}
We compare the IoU of our masks against PSD and R-NEM~\citep{van2018relational}.
Although PSD additionally receives the \textit{ground truth flow} during training, FlowCapsules consistently outperforms
with equal or better IoUs during testing, on both the Geo and Exercise datasets
(see Tab.\ \ref{tab:iou}).
One difference between PSD and FlowCapsules stems from the way they generate shape masks. 
PSD generates segmentation masks directly using convolutional layers with no encoding of the shape per se. In contrast, FlowCapsules uses a low-dimensional shape code to explicitly model the shape, from which the decoder generates the mask. As such the FlowCapsules encoder disentangles meaningful shape and pose information.

On Geo$^+$, FlowCapsule IoU performance degrades approximately 10\% to \textbf{0.85} (circle),  
\textbf{0.93} (square),  \textbf{0.90} (triangle) and overall to \textbf{0.89}. 
But compared to results in \Table{iou}, they remains as good or better than 
PSD on the simpler Geo data; we were not able to train PSD effectively on Geo$^+$. 

\begin{table}
\begin{center}
\begin{tabular}{@{}lcccc@{}}
            & \multicolumn{2}{c}{Geo} & \multicolumn{2}{c}{Geo+} \\ \toprule
            & N=4       & N=100         & N=4        & N=100       \\ \midrule
SCAE        & 0.48      & 0.59          & 0.49       & 0.51        \\
FlowCapsule & \textbf{0.79}      & \textbf{0.99}          & \textbf{0.52}       & \textbf{0.74} \\
\bottomrule
\end{tabular}
\end{center}
\vspace{-0.6em}
\caption{
\textbf{Quantitative / Classification: } K-means clustering accuracy with 4 or 100 clusters for Geo and Geo$^+$. 
FlowCapsule part representations yields higher classification accuracy than those learned from SCAE.
}
\vspace*{-0.35cm}
\label{tab:classification}
\end{table}

\vspace*{-0.1cm}
\subsection{Unsupervised Classification}
\vspace*{-0.1cm}

To evaluate FlowCapsules in the broader context of capsule classification, we replace the \textit{primary} 
capsule autoencoder (bottom of the stack) in SCAE \citep{kosiorek2019stacked} with FlowCapsules.
We call the new model \textit{FlowSCAE}.
We then train the top SCAE \textit{object} capsules to reconstruct the pose of FlowCapsules, 
following the original SCAE paper.
We compare the results against SCAE trained on reconstructing images from Geo and Geo$^+$.
SCAE training was modified slightly to produce coloured templates for the GEO dataset, and to 
produce textured templates in the primary capsules for Geo$^+$ (see supplementary material for details).

\begin{figure}[h]
    \centering
    \includegraphics[width=0.975\linewidth]{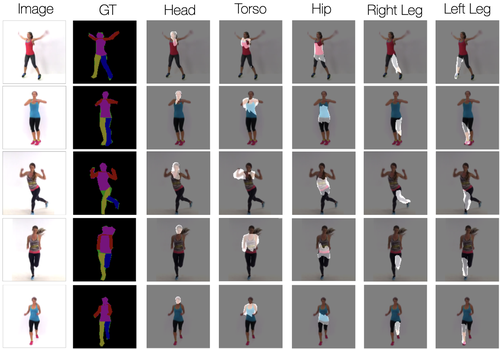}
    \vspace*{-0.1cm}
    \caption{The ground truth segment masks along with sample FlowCapsule masks $\Lambda_k^+$ on Exercise test data.}
    \label{fig:ex_masks}
\end{figure}

\Table{classification} reports unsupervised classification results using k-means clustering 
with $N$ clusters, for which the predicted label is set to the most common label in a given cluster.
We report the accuracy with $N{=}4$ and $N{=}100$ clusters.
Note that even though we trained the K-means of FlowSCAE with $N{=}100$ on the Geo data,
the learnt representations contained only $28$ clusters.

\vspace*{-0.1cm}
\subsection{Ablation Studies}
\vspace*{-0.1cm}

To better understand and analyze the significance of our design elements we perform ablations on various parameters. 

\paragraph{Number of capsules~($K$)}
Results in Tab.~\ref{tab:ablation} show that increasing the number of capsules tends to improve IoU performance. 
Given that our model \textit{does not} have an explicit sparsity regularizer on the capsules, this result is intriguing. 
Even with large numbers of capsules available, FlowCapsules does not break shapes into smaller pieces.
Rather, it learns one capsule per shape, relying more heavily on the layer occlusion to explain observed shape variation.

\paragraph{Encoding length $|s_k|$} 
The models are quite robust with Geo and Geo$^+$ data.  
As the encoding dimension decreases from $27$ to $11$, IofU performance changes by only $2\%$. 
Degradation occurs mainly with the circle class, where the circle boundary appears locally linear in places. 
The degradation becomes worse with  $|s_k| =3$, although even then, FlowCapsules still outperforms PSD. 

\paragraph{Number of hidden layers in $\masknet$}
One can hypothesize that deeper decoders can offset issues due to shorter shape encodings. 
Table \ref{tab:ablation} shows that increasing decoder depth from $2$ to $6$ improves IoU scores.
With Geo, the deeper decoder produces smoother circles.

\paragraph{Occlusion inductive bias}
Finally, we consider the effect of depth ordering in Eq.~(\ref{eqn:lambdaplus}) for occlusion  handling. 
Without depth ordering, Tab.~\ref{tab:ablation} shows a significant drop in performance.
In this case the masks become smoother and less certain in local regions, and the flow fields appear to be
the result of mixing a larger number of capsules, which tend to fit the observations less well in most cases.

\begin{table}[t]
\begin{center}

\begin{tabular}{llll}
\toprule
K  & $|s_k|$  & Geo & Geo$^+$ \\
\midrule
4  & 11 & 0.94 &   0.77\\
8  & 11 & 0.93 &   0.83\\
16 & 11 & 0.94 &   0.88\\
\midrule
8  & 3  & 0.91 &   0.86\\
8  & 27 & 0.96 &   0.89\\ 
\bottomrule
\end{tabular}
\begin{tabular}{lll}
\toprule
Depth  & Decoder   & Geo  \\
\midrule
No & 6-Layer    & 0.54\\
\hline
Yes & 2-Layer   & 0.87\\
Yes & 6-Layer   & 0.96\\
\bottomrule
\end{tabular}
\caption{{\bf IoU on Geo and Geo$^+$} for different number of capsules, 
encoding lengths, decoder depths, and depth ordering.
}
\label{tab:ablation}
    
\end{center}
\end{table}

\section{Conclusion}

We introduce FlowCapsules, an unsupervised method for learning capsule part representations 
(i.e.,~primary capsules). 
The capsule encoder takes as input a single frame and estimates a set of primary 
capsules, each comprising a shape mask in canonical coordinates, a pose transformation 
from canonical to image coordinates, and a scalar representing relative depth.
Training is done in a self-supervised manner from consecutive video frames.
We use a~Siamese architecture to estimate a parametric optical flow field between 
two frames, for which the flow is parameterized in terms of the poses of corresponding part capsules in the two frames.
Given a single frame, our capsule encoder learns to detect and encode the movable parts in an image. 
This approach differs significantly from other approaches that essentially 
segment the flow field itself into \textit{moving} parts (vs.~\textit{movable} parts in FlowCapsules).

Empirical results show that motion self-supervision in FlowCapsules is effective on real (see Fig.~\ref{fig:ex_masks}) and synthetic data, 
learning meaningful representations, completing shapes when partially occluded.
While formulated and tested within a specific capsule framework, our approach to self-supervised parts discovery 
is applicable to myriad encoder architectures, and to other approaches that currently use an image-reconstruction 
loss or rely on optical flow as input. Combining motion-based self-supervision with attention-based encoders  
\citep{locatello2020object} would enhance compositionality, allowing scenes with different numbers of objects.
Future work will also include scaling to larger video datasets and 3D parts.
To that end it will be important to extend the approach to include camera motion, and to handle
large motions of small objects for which more sophisticated losses for self-supervised learning will be necessary.
Alternatively, the FlowCapsules framework should be directly applicable to 3D observations, like point cloud data
\citep{zhao20193d}.

\section*{Broader Impact}
FlowCapsules impacts several tasks in image understanding, including Image segmentation and object detection. One of the major issues with Neural Networks is privacy and lack of {\bf interpretability}. Typical Neural Networks are black boxes that learn hidden representations. Analyzing these general representation and adding structural regularizers \citep{zhang2018interpretable} is one of the paths researchers are taking recently to improve interpretability of CNNs. A strong positive impact of Capsule Networks is their {\bf structured representation}. The hidden neurons in a Capsule Network encode specific information relevant to the image understanding task. Hence they have been widely adapted in the {\bf Medical Imaging} domain \citep{afshar2018brain,mobiny2018fast}. FlowCapsules have enhanced the level of structure enforcement on their representations to a higher level than previous Capsules. The part poses are explicitly used to estimate Optical Flow (without any hidden layer in between) and the shape vectors are trained only to generate masks. Therefore, a FlowCapsule network exactly knows and can visualize all the information its neurons encode. 

Another major advantage of Capsule Networks is their inherent robustness to Adversarial attacks \citep{marchisio2023robcaps, zhao2021investigation, shah2023adversarial}. Replacing the current free-form, black box CNNs with Capsule Networks can give a boost to their {\bf advarsarial robustness}. However, previous capsule networks were struggling in face of complex images with background and texture. FlowCapsules is a step toward making Structured Object Oriented Networks the state of the art for image understanding tasks such as object recognition, detection, and segmentation.

SaVi \citep{kipf2021conditional} followed up FlowCapsules by adapting the technique to Slot Attention modules \citep{locatello2020object}. Replacing Capsules with Slots adds more flexibility in terms of representation capacity of objects and their motion. Furthermore, SaVi leverages a recurrent network architecture to aggregate object representation over the full length of the video as opposed to just two consecutive frames. While losing the ability of the FlowCapsules to detect objects even on a single image, SaVi expands the range of application to semi real videos. SaVi++ \citep{elsayed2022savi++} improves upon SaVi by estimating depth in addition to optical flow. The advances in SaVi++ enables experiments on Waymo Open dataset \citep{sun2020scalability} with promissing results. Most recently, object oriented representation learning combined with recent advances such as ViViT \citep{arnab2021vivit} and diffusion models \citep{kingma2021variational} is able to achieve state of the art in several downstream tasks including generative tasks \citep{wu2025neural} and object segmentation \citep{van2025moving}.

\chapter{RobustNeRF: Ignoring Distractors with Robust Losses
}
\label{chap:robustnerf}

Neural radiance fields (\NeRF) excel at synthesizing new views given multi-view, calibrated images of a static scene.
When scenes include distractors, which are not persistent during image capture (moving objects, lighting variations, shadows), artifacts appear as  view-dependent effects or 'floaters'.
To cope with distractors, we advocate a form of robust estimation for NeRF training, modeling distractors in training data as outliers of an optimization problem.
Our method successfully removes outliers from a scene and improves upon our baselines, on synthetic and real-world scenes.
Our technique is simple to incorporate in modern \NeRF frameworks, with 
few hyper-parameters.
It does not assume a priori knowledge of the types of distractors, and is instead focused on the optimization problem rather than pre-processing or modeling transient objects.

\section{Introduction}
\label{sec:intro}

The ability to understand the structure of a \textit{static} 3D scene from 2D images alone is a fundamental problem is computer vision~\citep{tewari2022advances}.  It finds applications in AR/VR for mapping virtual environments~\citep{mobilenerf,zhu2022nice,rosinol2022nerf}, in autonomous robotics for action planning~\citep{adamkiewicz2022vision}, and in photogrammetry to create digital copies of real-world objects~\citep{reizenstein2021common}.

Neural fields~\citep{neuralfields} have recently revolutionized this classical task, by storing 3D representations within the weights of a neural network~\citep{sitzmann2019srns}.
These representations are optimized by back-propagating image differences.
When the fields store view-dependent \textit{radiance} and volumetric rendering is employed~\citep{neuralvolumes}, we can capture 3D scenes with photo-realistic accuracy, and we refer to the generated representation as Neural Radiance Fields, or NeRF~\citep{nerf}.

Training of NeRF models generally requires a large collection of images equipped with accurate camera calibration, which can often be recovered via structure-from-motion~\cite{schonberger2016structure}.
Behind its simplicity, NeRF hides several assumptions.
As models are typically trained to minimize error in RGB color space, it is of paramount importance that images are photometrically consistent -- two photos taken from the same vantage point should be \emph{identical} up to noise.
Unless one employs a method explicitly accounting for it~\citep{urf}, one should manually hold a camera's focus, exposure, white-balance, and ISO fixed.

\begin{figure}
\centering
\includegraphics[width=\linewidth]{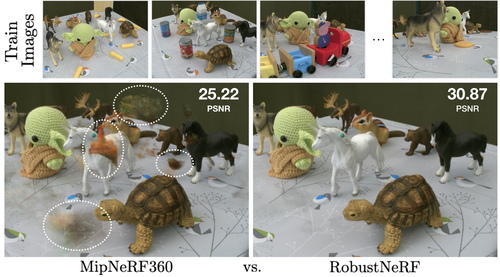}
\caption{
NeRF assumes photometric consistency in the observed images of  a scene.
Violations of this assumption, 
as with the images in the top row, yield reconstructed scenes with inconsistent content in the form of ``floaters'' (highlighted with ellipses).
We introduce a \textit{simple} technique that produces \textit{clean} reconstruction by automatically \textit{ignoring distractors} \textit{without explicit supervision}.
}
\vspace*{-0.30cm}
\label{fig:\currfilebase}
\end{figure}

However, properly configuring one's camera is not all that is required to capture high-quality NeRFs -- it is also important to avoid \textit{distractors}: anything that isn't persistent throughout the entire capture session.
Distractors come in many shapes and forms, from the hard-shadows cast by the operators as they explore the scene to a pet or child casually walking within the camera's field of view.
Distractors are tedious to \textit{remove} manually, as this would require pixel-by-pixel labeling.
They are also tedious to \textit{detect}, as typical NeRF scenes are trained from hundreds of input images, and the types of distractors are not known a priori.
If distractors are \textit{ignored}, the quality of the reconstruction scene suffers significantly; see \autoref{fig:teaser}.

In a typical capture session, it is difficult to 
to capture multiple images of the same scene from the same viewpoint, rendering distractors challenging to model mathematically.
As such, while view-dependent effects are what give NeRF their realistic look, \textit{how can the model tell the difference} between a distractor and a view-dependent effect?

Despite the challenges, the research community has devised several approaches to overcome this issue:
\begin{itemize}
\item If distractors are known to belong  to a specific class (e.g., people), one can remove them with a pre-trained semantic segmentation model~\citep{urf,tancik2022block} -- this process does \textit{not generalize} to ``unexpected'' distractors such as shadows.
\item One can model distractors as per-image \textit{transient} phenomena, and control the balance of transient/persistent modeling~\citep{nerfw} -- however, it is \textit{difficult to tune} the losses that control this Pareto-optimal objective.
\item One can model data in time (i.e., high-framerate video) and decompose
the scene into static and dynamic (i.e., distractor) components~\citep{ddnerf} -- but this clearly only applies to \textit{video} rather than photo collection captures.
\end{itemize}
\noindent
Conversely, we approach the problem of distractors by modeling them as \textit{outliers} in NeRF optimization. 

We analyze the aforementioned techniques through the lens of  robust estimation, 
allowing us to understand their behavior, and to  design a method that is not only simpler to implement but also more effective (see~\autoref{fig:teaser}).
As a result, we obtain a method that is straightforward to implement, requires minimal-to-no hyper-parameter tuning, and achieves state-of-the-art performance.
We evaluate our method: %
\begin{itemize}
\item quantitatively, in terms of reconstruction with synthetically, yet photo-realistically, rendered data;
\item qualitatively on publicly available datasets 
(often fine-tuned to work effectively with previous methods);
\item on a new collection of natural and synthetic scenes, including those autonomously acquired by a robot, allowing us to demonstrate the sensitivity of previous methods to hyper-parameter tuning.
\end{itemize}

\section{Related Work}
\label{sec:related}
We briefly review the basics and notation of Neural Radiance Fields.
We then  
describe recent progress in NeRF research, paying particular attention to techniques for modeling of static/dynamic scenes.

\paragraph{Neural Radiance Fields}
\label{sec:nerfreview}
A neural radiance field (NeRF) is a continuous volumetric representation of a 3D scene, stored within the parameters of a neural network $\params$.
The representation maps a position $\position$ and view direction $\dir$ to a \textit{view-dependent} RGB color and \textit{view-independent} density:
\begin{equation}
\begin{rcases}
\radiance(\position, \dir) \\
\density(\position)
\end{rcases} 
f(\position, \dir ; \params)
\label{eq:nerfmodel}
\end{equation}
This representation is trained from a collection,~$\{(\image_i, \pose_i)\}$, of images~$\image_i$ with corresponding calibration parameters~$\pose_i$~(camera extrinsics and intrinsics).

During training the calibration information is employed to convert each pixel of the image into a ray~$\ray {=} (\origin, \dir)$, and rays are drawn randomly from input images to form a training mini-batch~($\ray {\sim} \image_i$).

The parameters $\params$ are optimized to correctly predict the colors of the pixels in the batch via the L2 photometric-reconstruction loss:
\begin{align}
\loss{rgb}(\params) &= \sum_{i} \expect{\ray \sim \image_i}
{\left[
\loss{rgb}^{\ray, i}(\params) 
\right]}
 \\
\loss{rgb}^{\ray, i}(\params) &= ||\C(\ray; \params) - \image_i(\ray) ||_2^2
\label{eq:nerfloss}
\end{align}
Parameterizing the ray as $\ray(t)=\origin + t\mathbf{d}$, the NeRF model image $\C(\ray; \params)$ is generated pixel-by-pixel volumetric rendering based on $\density(\cdot)$ and $\radiance(\cdot)$ (e.g., see \cite{volrendigest,nerf}).

\paragraph{Recent progress on NeRF models}
\label{sec:recentnerf}
NeRF models have recently been extended
in several ways.
A major thread has been the speedup of training~\citep{ingp, relufields} and inference~~\citep{snerg, mobilenerf}, enabling today's models to be trained in minutes~\citep{ingp}, and rendered on mobile in real-time~\citep{mobilenerf}.
While initially restricted to forward-facing scenes, researchers quickly found ways to model real-world $360^\circ$ scenes~\citep{mipnerf360, nerf++}, and to reduce the required number of images, via sensor fusion~\citep{urf} or hand-designed priors~\citep{regnerf}.
We can now deal with image artifacts such as motion blur~\citep{deblurnerf}, exposure~\citep{rawnerf}, and lens distortion~\citep{scnerf}.
And the requirement of (precise) camera calibrations is quickly being relaxed with the introduction of techniques for local camera refinement~\citep{barf, garf}, or  direct inference~\citep{relpose}.
While a NeRF typically represents geometry via volumetric density, there exist models custom-tailored to predict surfaces~\citep{neus, unisurf}, which can be extended to use predicted normals to significantly improve reconstruction quality~\citep{monosdf, neuris}.
Given high-quality normals~\citep{refnerf}, inferring the (rendering) structure of a scene becomes a possibility~\citep{samurai}.
We also note recent papers about additional applications to generalization~\citep{yu2021pixelnerf}, semantic understanding~\citep{nesf}, generative modeling~\citep{lolnerf}, robotics~\citep{adamkiewicz2022vision}, and text-to-3D~\citep{dreamfusion}.

\paragraph{Modeling non-static scenes}
\label{sec:staticvsdynamic}
\begin{figure*}[t]
\begin{center}
    \includegraphics[width=0.99\linewidth]{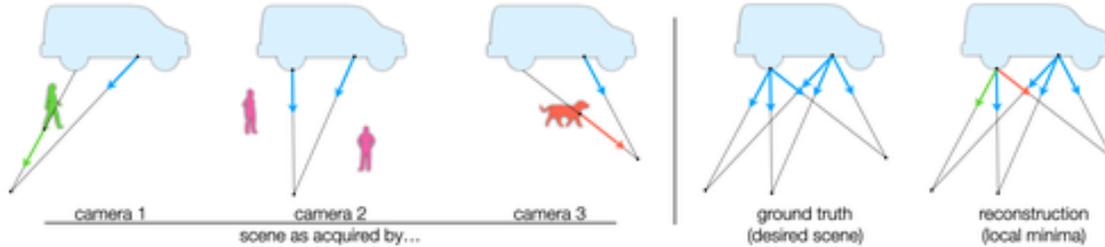}
\end{center}
\vspace{-1em}
\caption{
    \textbf{Ambiguity --}
    A simple 2D scene where a static object~(blue) is captured by three cameras.
    During the first and third capture the scene is not photo-consistent as a distractor was within the field of view.
    Not photo-consistent portions of the scene can end up being encoded as view-dependent effects -- even when we assume ground truth geometry.
}
\vspace*{-0.25cm}
\label{fig:\currfilebase}
\end{figure*}

For unstructured scenes like those considered here, the community has focused on reconstructing both static and non-static elements from video.
The most direct approach, treating time as an auxiliary input, leads to cloudy geometry and a lack of fine detail~\citep{xian2021space,gao2021dynamic}.
Directly optimizing per-frame latent codes as an auxiliary input has proved more effective~\citep{li2022neural,hypernerf,ddnerf}.
The most widely-adopted approach is to fit a time-conditioned deformation field mapping 3D points between pairs of frames~\citep{nsff,wang2021neural} or to a canonical coordinate frame~\citep{du2021neural,pumarola2021d,tretschk2021non,fang2022fast,liu2022devrf}.
Given how sparsely space-time is sampled, all methods require careful regularization, optimization, or additional training signals to achieve acceptable results.

Relatively little attention has been given to \emph{removing} non-static elements.
One common approach is to segment and ignore pixels which are likely to be distractors~\citep{tancik2022block, urf}.
While this eliminates larger objects, it fails to account for secondary effects like shadows.
Prior attempts to model distractors as outliers still leave residual cloudy geometry~\citep{nerfw}.

\section{Method}
\label{sec:method}

The classical NeRF training losses \eq{nerfloss} are effective for capturing scenes that are photometrically consistent, leading to the photo-realistic novel-view synthesis that we are now accustomed to seeing in recent research.
However, ``\textit{what happens when there are elements of the scene that are not persistent throughout the entire capture session?}''
Simple examples of such scenes include those in which an object is only present in some fraction of the observed images, or may not remain in the same position in all observed images.
For example, \autoref{fig:ambiguity}
depicts a 2D scene comprising a persistent object (the truck), along with several transient objects (e.g., people and a dog).
While rays in blue from the three cameras intersect the truck, the green and orange rays from cameras 1 and 3  intersect transient objects.
For video capture and spatio-temporal NeRF models, the persistent objects comprise the ``static'' portion of the scene, while the rest would be called the ``dynamic''.

\subsection{Sensitivity to outliers}
\label{sec:lambertian}

For Lambertian scenes, photo-consistent structure is view independent, as scene radiance only  depends on the incident light  \citep{SpaceCarving200}.
For such scenes, view-dependent NeRF models like~\eq{nerfmodel}, trained by minimizing \eq{nerfloss}, admit local optima in which transient objects are explained by view-dependent terms.
\autoref{fig:ambiguity} depicts this,
with the outgoing color corresponding to the memorized color of the outlier -- i.e. view-dependent radiance.
Such models exploit the view-dependent capacity of the model  to over-fit observations, effectively memorizing the transient objects.
One can alter the model to remove dependence on $\dir$, but the L2 loss remains problematic as least-squares (LS) estimators are sensitive to outliers, or heavy-tailed noise distributions.
 
Under more natural conditions, dropping the Lambertian assumption, the problem becomes more complex as {\em both} non-Lambertian reflectance phenomena and outliers can be explained as view-dependent radiance.
While we want the models to capture photo-consistent view-dependent radiance, 
outliers and other transient phenomena should ideally be ignored.
And in such cases, optimization with an L2 loss~\eq{nerfloss} yields significant errors in reconstruction; see~\autoref{fig:teaser}.
Problems like these are pervasive in NeRF model fitting, especially in  uncontrolled environments with complex reflectance, non-rigidity, or independently moving objects.

\subsection{Robustness to outliers}

\paragraph{Robustness via semantic segmentation}
One way to reduce outlier contamination during NeRF model optimization is to rely on an {\em oracle}~$\oracle$ that specifies whether a given pixel~$\ray$ from image~$i$ is an outlier,  and should therefore be excluded from the empirical loss, replacing \eq{nerfloss} with:
\begin{equation}
\loss{oracle}^{\ray, i}(\paramss) = 
\oracle_i(\ray)
\cdot
||\C(\ray; \paramss) - \image_i(\ray) ||_2^2 
\label{eq:oracleloss}
\end{equation}
In practice, a \textit{pre-trained} (semantic) segmentation network $\segmenter$ might serve as an oracle,  $\oracle_i {=} \segmenter(\image_i)$.
E.g., Nerf-in-the-wild~\citep{nerfw} employed a semantic segmenter to remove pixels occupied by people, as they are  outliers in the context of  photo-tourism.
Urban Radiance Fields~\citep{urf} segmented out sky pixels, while LOL-NeRF~\citep{lolnerf} ignored pixels not belonging to faces.
The obvious problem with this approach is the need for an oracle to detect arbitrary distractors.

\paragraph{Robust estimators}
Another way to reduce sensitivity to outliers is to replace the conventional L2 loss  \eq{nerfloss} with a \textit{robust loss} (e.g., \citep{regcourse,robustloss}), so that photometrically-inconsistent observations can be down-weighted during  optimization.
Given a robust kernel $\kernel(\cdot)$, we rewrite our training loss as:
\begin{equation}
\loss{robust}^{\ray, i}(\paramss) = 
\kernel(||\C(\ray; \paramss) - \image_i(\ray) ||_2) 
\label{eq:robustloss}
\end{equation}
where $\kernel(\cdot)$ is positive and monotonically increasing.
MipNeRF~\citep{mipnerf}, for example, employs an L1 loss $\kappa(\residuall){=} | \residuall |$, which provides some degree of robustness to outliers during NeRF training.
Given our analysis, a valid question is whether we can straightforwardly employ a robust kernel to approach our problem, and if so, given the large variety of robust kernels~\citep{robustloss}, which is the kernel of choice.

\begin{figure}[t!]
\vspace*{-0.1cm}
\centering
\includegraphics[width=.7\linewidth]{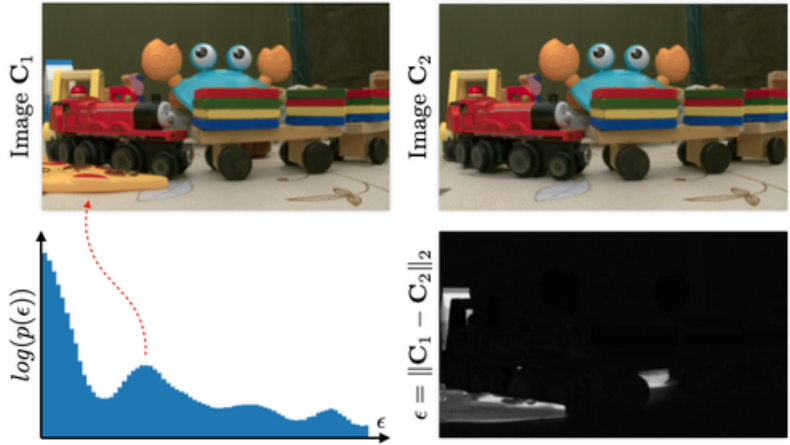}
\caption{\textbf{Histograms} -- 
Robust estimators perform well when the distribution of residuals agrees with the one implied by  the estimator (e.g., Gaussian for L2, Laplacian for L1).
Here we visualize the ground-truth distribution of residuals (bottom-left), which is hardly a good match with any simple parametric distribution.
}
\vspace*{-0.25cm}
\label{fig:histograms}
\end{figure}

Unfortunately, as discussed above, outliers and non-Lambertian effects can \textit{both} be modelled as view-dependent effects~(see \autoref{fig:histograms}).
As a consequence, with simple application of robust estimators it can be difficult to separate signal from noise.
\autoref{fig:kernels} shows examples in which outliers are removed, but fine-grained texture and  view-dependent details are also lost, or conversely, fine-grained details are preserved, but outliers cause artifacts in the reconstructed scene.
One can also observe mixtures of these cases in which details are not captured well, nor are outliers fully removed.
We find that this behaviour occurs consistently for many different robust estimators and parameter settings.

Training time can also be problematic. The robust estimator gradient w.r.t.\ model parameters can be expressed using the chain rule as
\begin{equation}
\left.
\frac{\partial \kernel(\residuall(\paramss))}{\partial \paramss}
\right|_{\paramss^{(t)}}
=
\left.
\frac{\partial \kernel(\residuall)}{\partial \residuall}
\right|_{\residuall(\paramss^{(t)})}
\cdot 
\left.
\frac{\partial \residuall(\paramss)}{\partial \paramss}
\right|_{\paramss^{(t)}}
\label{eq:chainrule}
\end{equation}
The second factor is the classical NeRF gradient.
The first factor is the kernel gradient evaluated at the \textit{current} error residual $ \residuall(\paramss^{(t)})$.
During training, large residuals can \textit{equivalently} come from high-frequency details that have not yet been learnt, or they may arise from outliers (see \autoref{fig:kernels}~(bottom)).
This explain why robust optimization, implemented as \eq{robustloss}, should not be expected to decouple high-frequency details from outliers.
Further, when \textit{strongly} robust kernels are employed, like redescending estimators, this also explains the loss of visual fidelity.  That is, because the gradient of (large) residuals get down-weighted by the (small) gradients of the kernel, \textit{slowing down} the learning of these fine-grained details (see \autoref{fig:kernels}~(top)).

\begin{figure}[t]
\begin{center}
\includegraphics[width=\linewidth]{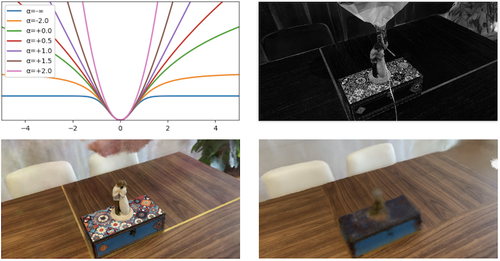}
\end{center}
\vspace*{-0.2cm}
\caption{
\textbf{Kernels --}
(top-left) Family of robust kernels~\citep{robustloss}, including L2~($\alpha{=}{2}$), Charbonnier~($\alpha{=}{1}$) and Geman-McClure~($\alpha{=}{-2}$). 
(top-right)
Mid-training, residual magnitudes are similar for distractors and fine-grained details, and pixels with large residuals are learned more slowly, as the gradient of re-descending kernels flattens out.
(bottom-right)
A too aggressive Geman-McClure in down-weighting large residuals 
removes both outliers and high-frequency detail.
(bottom-left)
A less aggressive Geman-McClure does not effectively remove outliers.}
\vspace*{-0.2cm}
\label{fig:\currfilebase}
\end{figure}

\subsection{Robustness via Trimmed Least Squares}
\vspace*{-0.1cm}

In what follows we advocate a form of iteratively reweighted least-squares (IRLS) with a Trimmed least squares (LS) loss for NeRF model fitting.

\vspace*{-0.05cm}
\paragraph{Iteratively Reweighted least Squares}
\begin{figure*}[t]
\centering
\includegraphics[width=0.925 \linewidth]{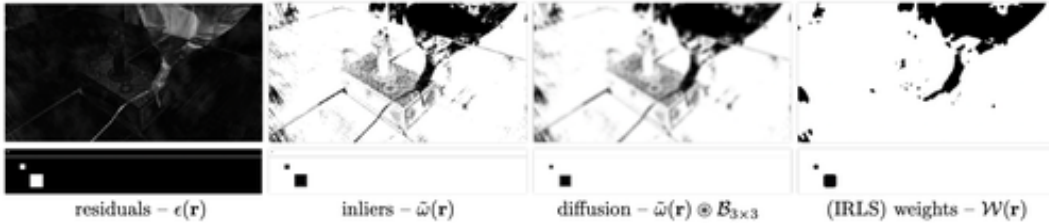}
\caption{\textbf{Algorithm} --
We visualize our weight function computed by residuals on two examples: 
(top) the residuals of a (mid-training) NeRF rendered from a \textit{training} viewpoint, 
(bottom) a toy residual image containing residual of small spatial extent (dot, line) and residuals of large spatial extent (squares).
Notice residuals with large magnitude but small spatial extent (texture of the box, dot, line) are included in the optimization, while weaker residuals with larger spatial extent are excluded.
Note that while we operate on patches, we visualize the weight function on the whole image to facilitate visualization.
}
\vspace*{-0.25cm}
\label{fig:\currfilebase}
\end{figure*}

IRLS is a widely used method for robust estimation that involves solving a sequence of weighted LS problems, the weights of which are adapted to reduce the influence of outliers.
To that end, at iteration $t$,  one can write the loss as
\begin{align}
\loss{robust}^{\ray, i}(\paramss^{(t)}) &= 
\weight(\residuals^{(t-1)}(\ray)) \cdot || \C(\ray; \paramss^{(t)}) - \image_i(\ray) ||_2^2
\label{eq:irls}
\nonumber
\\
\residuals^{(t-1)}(\ray) &= || \C(\ray; \paramss^{(t-1)}) - \image_i(\ray) ||_2
\end{align}
For weight functions given by 
$\weight(\residuall) {=} \residuall^{-1} \cdot \partial \kernel(\residuall) / \partial \residuall$
one can show that, under suitable conditions, the iteration converges to a local minima of \eq{robustloss} (see~\cite[Sec.~3]{regcourse}).

This framework admits a broad family of losses, including maximum likelihood estimators for heavy-tailed noise processes. Examples in \autoref{fig:kernels} include the Charbonnier loss 
(smoothed L1), and more aggressive redescending estimators such as the Lorentzian or Geman-McClure \citep{robustloss}.
The objective in~\eq{oracleloss} can also be viewed as a weighted LS objective, the binary weights of which are provided by an oracle.
And, as discussed at length below, one can also view several recent methods like NeRFW~\citep{nerfw} and \ddnerf~\citep{ddnerf} through the lens of IRLS and weighted LS.

Nevertheless, choosing a suitable weight function $\weight(\residuall)$ for NeRF optimization is non-trivial, due in large part to the intrinsic ambiguity between view-dependent radiance phenomena and outliers.
One might try to solve this problem by learning a neural weight function~\citep{acne}, although generating enough annotated training data might be prohibitive.
Instead, the approach taken below is to exploit inductive biases in the structure of outliers, combined with the simplicity of a robust, trimmed LS estimator.

\paragraph{Trimmed Robust Kernels}
Our goal is to develop a weight function for use in iteratively weighted LS optimization that is simple and captures useful inductive biases for NeRF optimization.
For simplicity we opt for a binary weight function with intuitive parameters that adapts naturally through model fitting so that fine-grained image details that are not outliers can be learned quickly.
It is also important to capture the structured nature of typical outliers, contrary to the typical i.i.d. assumption in most robust estimator formulations.
To this end, the weight function should capture spatial smoothness of the outlier process, recognizing that objects typically have continuous local support, and hence outliers are expected to occupy large, connected regions of an image~(e.g., the silhouette of a person to be segmented out from a photo-tourism dataset).

Surprisingly, a relatively simple weight function embodies these properties and performs extremely well in practice.
The weight function is based on so-called \textit{trimmed estimators} that are used in trimmed least-squares, like that used in trimmed ICP~\citep{trimmedicp}.
We first \textit{sort} residuals, and assume that residuals below a certain percentile are inliers.
Picking the 50\% percentile for convenience~(i.e., median), we define
\begin{equation}
\tilde{\weight}(\ray) = \residuall(\ray) \leq \threshold_\residuall ~,
~~~
\threshold_\residuall = \text{Median}_\ray \{\residuall(\ray)\} ~.
\label{eq:trimming}
\end{equation}

To capture spatial smoothness of outliers we spatially diffuse inlier/outlier labels $\weight$ with a $3{\times}3$ box kernel $\mathcal{B}_{3\!\times\!3}$.
Formally, we define
\begin{align}
\mathcal{W}(\ray) = \tilde\weight(\ray) | (\tilde\weight(\ray) \circledast \mathcal{B}_{3\times3}) \geq \threshold_\circledast ~,
~~~ \threshold_\circledast = 0.5 ~.
\label{eq:smoothing}
\end{align}
This helps to avoid classifying high-frequency details as outliers, allowing them to be captured by the NeRF model during optimization (see~\autoref{fig:algorithm}).

While the trimmed weight function (\ref{eq:smoothing}) improves the robustness of model fitting, it sometimes misclassifies fine-grained image details early in training where the NeRF model first captures coarse-grained structure. These localized texture elements may emerge but only after very long training times.
We find that stronger inductive bias to spatially coherence allows fine-grained details to be learned more quickly.
To that end, we  aggregate the detection of outliers on $16\!{\times}\! 16$ neighborhoods;
i.e., we label entire $8\!\times\! 8$ patches as outliers or inliers based on the behavior of $\mathcal{W}$ in the $16 {\times} 16$ neighborhood of the patch.
Denoting the $N {\times} N$ neighborhood of pixels around $\ray$ as $\mathcal{R}_{N}(\ray)$, we define
\newcommand{\sray}{\mathbf{s}}
\begin{align}
\weight(\mathcal{R}_8(\ray)) &= \mathcal{W}(\ray) | \expect{\sray \sim \mathcal{R}_{16}(\ray)}
{\left[ 
\mathcal{W}(\sray)
\right]} \geq \threshold_\mathcal{R} \, , ~~~ \threshold_\mathcal{R}=0.6 \, .
\label{eq:patching}
\end{align}
This robust weight function 
evolves during optimization, as one expects with IRLS where the weights are a function of the residuals at the previous iteration.
That is, the labeling of pixels as inliers/outliers \textit{changes} during training, and settles around masks similar to the one an oracle would provide as training converges (see~\autoref{fig:residuals}).

\begin{figure}[t!]
\centering
\includegraphics[width=.7\linewidth]{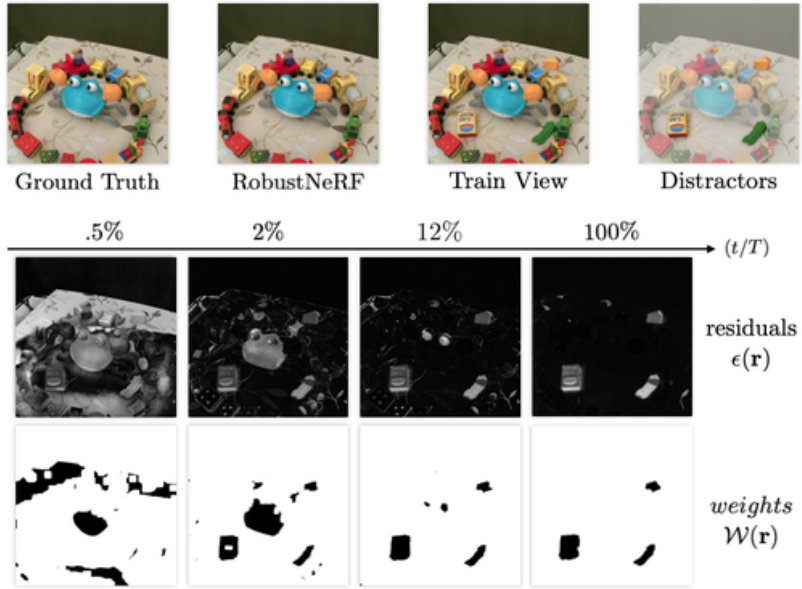}
\caption{
\textbf{Residuals -- }
For the dataset shown in the top row, we visualize the dynamics of the \RobustNeRF training residuals, which show how over time the estimated distractor weights go from being random~($(t/T){=}0.5\%$) to identify distractor pixels~($(t/T){=}100\%$) without any explicit supervision.
}
\vspace*{-0.25cm}
\label{fig:\currfilebase}
\end{figure}

\section{Experiments}
\label{sec:experiments}
\vspace*{-0.1cm}

We implement our robust loss function in the MultiNeRF codebase~\citep{multinerf2022} and apply it to \mipNeRFthreesixty~\citep{mipnerf360}.
We dub this method ``\RobustNeRF''.
To evaluate \RobustNeRF, we compare  against baselines on several scenes containing different types of distractors.
Where possible, we quantitatively compare reconstructions to held-out, distraction-free images;
we report three metrics, averaged across held-out frames, namely, 
PSNR, SSIM~\citep{ssim2004}, and LPIPS~\citep{zhang2018perceptual}.

We compare different methods on two collections of scenes, i.e., those provided by the authors of \DDNeRF, and \textit{novel} datasets described below.
We also present a series of illustrative experiments on synthetic scenes, shedding light on \RobustNeRF's efficacy and inner workings.

\subsection{Training Details}
\label{sec:trainingdetails}

While camera parameters are estimated on the full-resolution imagery, we downsample images by 8x for each natural scene dataset.
While \mipNeRFthreesixty and \RobustNeRF are capable of training on high resolution photos, we limit the resolution to accommodate \DDNeRF.
Unless otherwise stated, we train on all available cluttered images, and evaluate on a holdout set; i.e., 10 images for Android; 19 for the Statue dataset; 194 for Crab; and 202 for the BabyYoda dataset
(see \autoref{tab:dataset_details}).

\paragraph{\RobustNeRF}
\label{app:sec:robustnerf}
We implement \RobustNeRF by incorporating our proposed loss function into the MultiNeRF  codebase~\citep{multinerf2022}, replacing  \mipNeRFthreesixty's~\citep{mipnerf360} reconstruction loss.
All other terms in the loss function, such as regularizers, are included as originally published in \mipNeRFthreesixty.

We train \RobustNeRF for 250,000 steps with the Adam optimizer, using a batch size of 64 image patches randomly sampled from training images.
Each pixel within a 16x16 patch contributes to the loss function, except those identified as outliers
(see \autoref{fig:residuals} for a visualization).
The learning rate is exponentially decayed from 0.002 to 0.00002 over the course of training with a warmup period of 512 steps.

Our model architecture comprises a proposal Multilayer Perceptron (MLP) with 4 hidden layers and 256 units per layer, and a \NeRF MLP with 8 hidden layers, each with 1024 units.
We assign each training image a 4-dimensional GLO vector to account for unintended appearance variation.
Unless otherwise stated, we use the robust loss hyperparameters given in the main body of the chapter.
All models are trained on 16 TPUv3 chips over the course of 9 hours.
\subsection{Baselines}
\vspace*{-0.1cm}

We compare \RobustNeRF to variants of \mipNeRFthreesixty optimized with different loss functions~($L_2$, $L_1$, and Charbonnier).
These variants serve as natural baselines for models 
with limited or no robustness to outliers.We also compare to \DDNeRF, a recent method for reconstructing dynamic scenes from monocular \textit{video}. Unlike our method, \DDNeRF is designed to \textit{reconstruct} distractors rather than discard them.
While \DDNeRF is presented as a method for monocular video, it does not presuppose  temporal continuity,  and can be directly applied to unordered images.
We omit additional comparisons to \NeRFW as its performance falls short of~\DDNeRF~\citep{ddnerf}.

\paragraph{\mipNeRFthreesixty~\citep{mipnerf360}}
\label{app:sec:mipnerf360}
We use the reference implementation of \mipNeRFthreesixty from the MultiNeRF codebase.
Similar to \RobustNeRF, we train each variant of  \mipNeRFthreesixty with the Adam optimizer, using the same number of steps, batch size, and learning rate schedule.
\mipNeRFthreesixty uses a random sample of 16384 rays per minibatch.
Proposal and \NeRF MLP depth and width are identical to those for \RobustNeRF.
Training hardware and duration are also the same as \RobustNeRF.

\paragraph{\ddnerf~\citep{ddnerf}}
\label{app:sec:ddnerf}
We use the reference implementation of \DDNeRF~\citep{ddnerf} provided by the authors.
Model architecture, hierarchical volume sampling density, and learning rate are the same as published in~\cite{ddnerf}.
As in the original work, we train the model for 100,000 iterations with a batch size of 1024 rays, though over the course of 3 hours. Due to hardware availability, we employ four NVIDIA V100 GPUs in place of the A100 GPUs used in the original work.

Images are kept in the order of provided by the file system (i.e., ordered by position information alphanumerically). However, this image order is not guaranteed to represent a continuous path in space since the images were not captured along a continuous path, but rather at random locations. Below we discuss the effects of random ordering versus ordering the views along a heuristically identified path.

\DDNeRF training is controlled by five key hyperparameters, namely, skewness ($k)$, which encourages a binarization loss to favor static explanations, and four regularization weights that scale the skewed binarization loss ($\lambda_s$), ray regularization loss ($\lambda_r$), static regularization loss ($\lambda_{\sigma^s}$), and the view-correlated shadow field loss ($\lambda_\rho$). A hyperparameter search is performed in \ddnerf for 16 real world scenes to identify combinations best suited for each scene, and four primary configurations of these parameters are identified as optimal. In particular, the first configuration (i.e., $k = 1.75$, $\lambda_s = 1e^{-4} \rightarrow 1e^{-2}$, $\lambda_r = 1e^{-3}$, $\lambda_{\sigma^s} = 0$, and $\lambda_\rho = 1e^{-1}$) was reported to be most effective across the largest number of scenes real world (10 of 16). We additionally conduct a tuning experiment (see \autoref{fig:ddnerf_tune}) and confirm the first configuration as best suited. We apply this configuration in all additional \ddnerf experiments.

\subsection{Datasets -- \autoref{fig:dataset}}
\vspace*{-0.1cm}

In addition to scenes from \DDNeRF, we introduce a set of natural and synthetic scenes. They facilitate the evaluation of \RobustNeRF's effectiveness on illustrative use cases, and they enable empirical analysis under controlled conditions. See \autoref{tab:dataset_details} for key details.

\begin{figure}
\centering
\includegraphics[width=.7\linewidth]{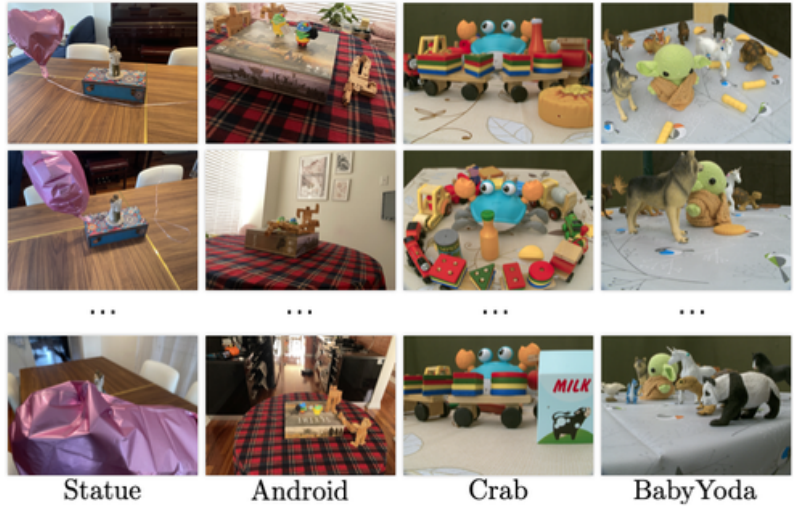}
\caption{\textbf{Dataset -- }
Sample training images showing the distractors in each scene. 
Statue and Android were acquired manually, and the others with a robotic arm. 
In the robotic setting we have pixel-perfect alignment of distractor vs.\ distractor-free images.
}
\vspace*{-0.25cm}
\label{fig:\currfilebase}
\end{figure}

\vspace{-0.05cm}
\subsubsection{Natural scenes}
We capture seven natural scenes exemplifying different types of distractors.
Scenes are captured in three settings, on the street, in an apartment and in a robotics lab.
Distractor objects are moved, or are allowed to move, between frames to simulate capture over extended periods of time.
We vary the number of unique distractors  from 1 (Statue) to 150 (BabyYoda),
and their  movements.
Unlike prior work on monocular video, frames are captured without a clear temporal ordering (see \autoref{fig:dataset}). 
The other three (i.e., Street1, Street2, and Gloss)  include view-dependence effects, the results of which are shown in the supplementary material.
We also capture additional frames \textit{without distractors} to enable quantitative evaluations.
Camera poses are estimated using COLMAP~\citep{schoenberger2016mvs}.

\begin{figure*}[t!]
\centering
\includegraphics[width=\textwidth]{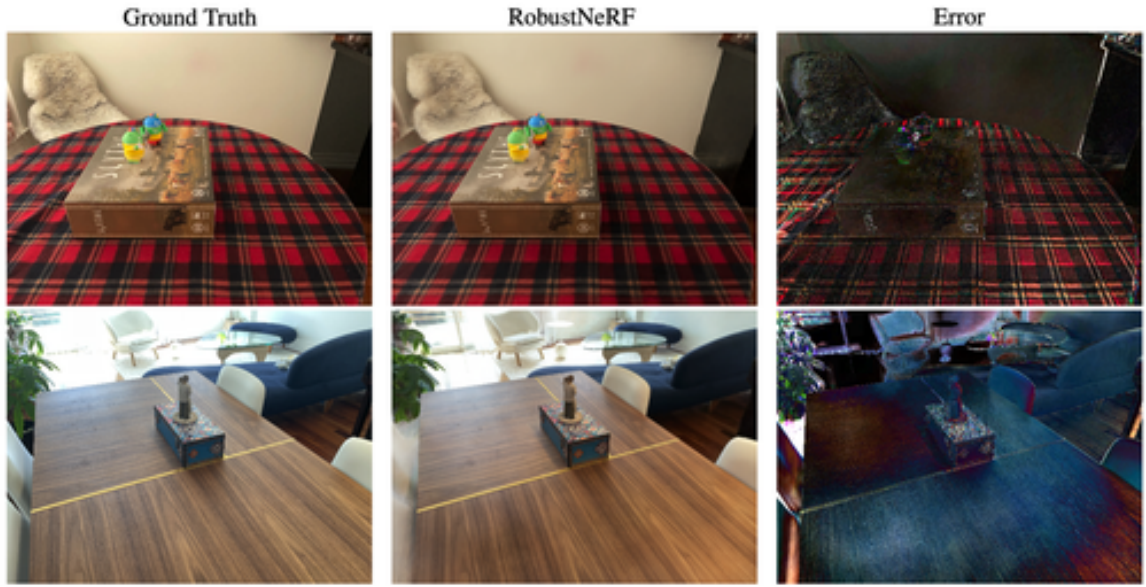}
\caption{
    \textbf{Challenges in Apartment Scenes -- }
  Each row, from left to right, shows a ground truth photo, a \RobustNeRF render, and the difference between the two. Best viewed in PDF.
   (Top) Note the fold in the table cloth in ground truth image and the lack of fine-grained 
    detail on the covered chair in the background.
    The table cloth moved during capture, and the background was not captured thoroughly enough for a high-fidelity reconstruction.
    (Bottom) The ground truth image for the Statue dataset exhibits overexposure and color calibration issues, and hence do not exactly match the \RobustNeRF render.
}
\vspace*{-0.1cm}
\label{fig:\currfilebase}
\end{figure*}

\paragraph{Apartment (Statue \& Android)}
\label{app:sec:apartmentscenes}
To mimic a casual home scenario, we capture two tabletop scenes in an apartment using a commodity smartphone.
Both captures focus on one or more objects on a table top, with photos taken from different viewpoints from a hemisphere of directions around the objects of interest.
A subset of objects on the table move from photo to photo as described below.
The photos within each scene do not have a clear temporal order.

The capture setup is as follows.
We employ an iPhone 12 mini and use ProCamera v15 to control camera exposure settings.
We use a fixed shutter speed of 1/60, 0.0 exposure bias, and a fixed ISO of 80 or 200 for the Statue and Android scenes, respectively.
We use the iPhone's standard wide lens with an aperture of f/1.6 and resolution of 4032x3024.
A tripod is used to reduce the effects of the rolling shutter.

The Android dataset comprises 122 cluttered photos and 10 clean photos
(i.e., with no distractors).
This scene depicts two Android robot figures standing on a board game box, which in turn is sitting on a table with a patterned table cloth.
We pose three small wooden robots atop the table in various ways in each cluttered photo to serve as distractors.

For the Statue scene, we capture 255 cluttered photos and 19 clean photos.
The scene depicts a small statue on top of a highly-detailed decorative box on a wooden kitchen table.
To simulate a somewhat persistent distractor, we float a balloon over the table which, throughout the capture, naturally changes its position slightly with each photo.
Unlike the Android scene, where distractors move to entirely new poses in each frame, the balloon frequently inhabits the same volume of space for multiple photos.
The decorative box and kitchen table both exhibit fine grained texture details. 

\begin{figure}
\centering

\resizebox{.7\linewidth}{!}{ %
\begin{tabular}{@{}l|cccccc@{}}
           & \# Clut. & \# Clean & \# Extra & Paired? & Res.      & Setting      \\
\midrule
Android     & 122      & 122        & 10       & No      & 4032x3024 & Apartment    \\
Statue    & 255      & 132        & 19       & No      & 4032x3024 & Apartment    \\
Crab       & 109      & 109      & 194      & Yes     & 3456x3456 & Robotics Lab \\
BabyYoda   & 109      & 109      & 202      & Yes     & 3456x3456 & Robotics Lab \\
\bottomrule
\end{tabular}
}

\caption{
    \textbf{Natural Scenes} --
    Key facts about natural scenes introduced in this work.
    Includes number of paired photos with (\# Clut.) and without (\# Clean) distractors.
    Extra photos (\# Extra) do not contain distractors and are taken from unpaired camera poses.
}
\label{tab:\currfilebase}
\end{figure}

We run COLMAP's~\citep{schonberger2016structure} Structure-from-Motion pipeline using the \texttt{SIMPLE\_RADIAL} camera model.
While COLMAP's camera parameter estimates are only approximate, we find that they are sufficient for training \NeRF models with remarkable detail.

The apartment scenes are considerably more challenging to reconstruct than the robotics lab scenes (described below).
An accurate \NeRF reconstruction must model not only the static, foreground content but also the scene's background.
Unlike the foreground, each object in the background is partially over- or underexposed and appears in a limited number of photos.
We further found it challenging to maintain a controlled, static scene during capture.
As a result, some objects in the background move by a small, unintended amount between photos (e.g., see \autoref{fig:residual_android}).

\paragraph{Robotics Lab (Crab \& BabyYoda)}
\label{app:sec:roboticslabscenes}
In an effort to control confounding factors in data acquisition, we capture two scenes in a Robotics Lab setting.
In these scenes, we employ a robotic arm to randomly position a camera within 1/4 of the hemisphere over a table.
The table is placed in a closed booth with constant, indoor lighting.
A series of toys are placed on the table, a subset of which are glued to the table's surface to prevent them from moving.
Between photos, distractor toys on the table are removed and/or new distractor toys are introduced.

For capture, we use a Blackfly S GigE camera with a TECHSPEC 8.5mm C Series fixed length lens.
Photos are center-cropped from their original resolution of 5472x3648 to 3456x3456 to eliminate lens distortion.
We capture 12-bit raw photos with an aperture of f/8 and exposure time of 650 ms.
Raw photos are automatically color-calibrated afterwards according to a reference color palette.

In each scene, we capture 109 pairs of photos from identical camera poses, one with distractors present and another without.
This results in a large number of unique distractors which are challenging to model directly.
This further allows us to investigate the counterfactual: What if distractors were \emph{not} present?
We further capture an additional $\sim$200 photos from random viewpoints, not aligned with those for training and without distractors, for the purposes of evaluation.
In total, because the placement of objects is done manually, one capture session often takes several hours.

\vspace{-0.1cm}
\subsubsection{Synthetic scenes}
To further evaluate \RobustNeRF, we generate synthetic scenes using the Kubric dataset generator~\citep{greff2021kubric}.
Each scene is constructed by placing a set of simple geometries in an empty, texture-less room.
In each scene, a subset of objects remain fixed while the other objects~(i.e., distractors) change position from frame to frame.
By varying the number of objects, their size, and the way they move, we control the level of distraction in each scene.
We use these scenes to examine \RobustNeRF's sensitivity to its hyperparameters.

\begin{figure}[t]
\centering
\includegraphics[width=.7\linewidth]{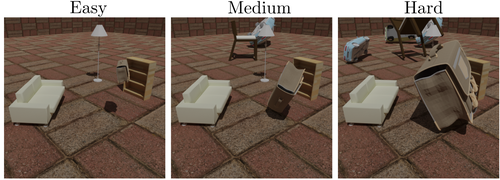}
\caption{
    \textbf{Synthetic Kubric Scenes} --
    Example Kubric synthetic images for three datasets with different ratio of outlier pixels. The sofa, lamp, and bookcase are static objects in all three setups. The easy setup has 1 small distractor, the medium setup has 3 medium distractors, and the hard setup has 6 large distractors.
}
\label{fig:\currfilebase}
\end{figure}

We generate three Kubric \citep{greff2021kubric} scenes similar to the \ddnerf synthetic scenes with different difficulty levels: easy, medium, and hard. These datasets are used to ablate our method with control on the proportion of outlier occupancy (see Sec.\  \ref{app:sec:sensitivity_to_hyperparameters}). 

 Each dataset contains 200 cluttered images for training  and 100 clean images for evaluation. In all three scenes the static objects include a sofa, a lamp and a bookshelf. \autoref{fig:synth} shows one example image from the training set for each dataset. The easy scene contains only one small distractor object (a bag). This dataset is similar to Kubric Bag dataset of \ddnerf. The medium scene has three distractors (a bag, a chair, and a car) which are larger in size and hence the outlier occupancy is  $4\times$ the outlier occupancy of the easy scene. The hard scene has six large distractors (a bag, a chair, and four cars). They occupy on average $10\times$ more pixels than the easy setup, covering  roughly half of each image. 

\newcommand{\mssim}{\scalebox{0.8}{MS-SSIM$\uparrow$}}

\begin{figure*}[t!]
\centering
\resizebox{\linewidth}{!}{ %
\begin{tabular}{@{}l|ccc|ccc|ccc|ccc@{}}
& \multicolumn{3}{c|}{\color{brown}\bf Statue} 
    & \multicolumn{3}{c|}{\color{dark_green} \bf Android} 
    & \multicolumn{3}{c|}{\color{red}\bf Crab} 
    & \multicolumn{3}{c}{\color{blue}\bf BabyYoda} 
\\
                             & \lpips & \ssim & \psnr & \lpips & \ssim & \psnr  & \lpips & \ssim & \psnr  & \lpips & \ssim & \psnr  \\
\midrule
\mipNeRFthreesixty($L_2$)    & 0.36  & 0.66 & 19.09 &  0.40  & 0.65 & 19.35 &  0.27 & 0.77  & 25.73  &  0.31  &  0.75 &  22.97      \\
\mipNeRFthreesixty($L_1$)    & 0.30  & 0.72 & 19.55 &  0.40  & 0.66 & 19.38 &  0.22 & 0.79  & 26.69  &  0.22   & 0.80  &  26.15      \\
\mipNeRFthreesixty(Ch.)      & 0.30  & 0.73 & 19.64 &  0.40 & 0.66 & 19.53 &  0.21  & 0.80  & 27.72  &  0.23 &  0.80  &  25.22     \\
\ddnerf                      & 0.48  & 0.49 &19.09 &   0.43 & 0.57 & 20.61  & 0.42  & 0.68 &  21.18  &  0.44 &  0.65  & 17.32    \\
\textbf{\RobustNeRF}         & \textbf{0.28 } & \textbf{0.75} & \textbf{20.89} &  \textbf{0.31} & \textbf{0.65} & \textbf{21.72} &  \textbf{0.21}  &   \textbf{0.81} & \textbf{30.75} & \textbf{0.20}  & \textbf{0.83}   & \textbf{30.87}  \\
\midrule
\mipNeRFthreesixty(clean)   & 0.19  &0.80 & 23.57 &  0.31 &  0.71 & 23.10 &   0.16      &  0.84    &   32.55    &  0.16     &   0.84   &   32.63    \\
\end{tabular}
}
\\[.5em]
\includegraphics[width=\linewidth]{robustnerf/fig/\currfilebase}
\vspace{-1em}
\captionof{figure}{
\textbf{Evaluation on Natural Scenes --}
\RobustNeRF 
outperforms baselines and  
\ddnerf~\citep{ddnerf} on novel view synthesis with  real-world captures.
The table provides a quantitative comparison of
\RobustNeRF, \ddnerf and \mipNeRFthreesixty using different reconstruction losses.
The last row reports \mipNeRFthreesixty trained on a distractor-free version of each dataset, giving an upperbound for \RobustNeRF performance. 
We also visualize samples from each scene rendered with each of the methods. See Supplementary Material for more samples.}
\label{fig:\currfilebase}
\vspace{-0.25cm}
\end{figure*}
\subsection{Evaluation}
\vspace*{-0.1cm}

\begin{figure*}[t]
\centering
\setlength\tabcolsep{1.8pt}
\resizebox{\linewidth}{!}{
\begin{tabular}{l|ccc|ccc|ccc|ccc|ccc}
\multicolumn{1}{c}{}  & \multicolumn{3}{c|}{Car}  & \multicolumn{3}{c|}{Cars} & \multicolumn{3}{c|}{\bf{Bag}} & \multicolumn{3}{c|}{Chairs} & \multicolumn{3}{c}{Pillow}
\\ & \lpips & \mssim & \psnr& \lpips & \mssim & \psnr& \lpips & \mssim & \psnr& \lpips & \mssim & \psnr& \lpips & \mssim & \psnr \\
\midrule
NeRF-W \citep{nerfw} &  .218 &  .814 & 24.23 &  .243 &  .873 & 24.51 &  .139 &  .791 & 20.65 &  .150 &  .681 & 23.77 &  .088 &  .935 & 28.24 \\
NSFF \citep{nsff} & .200 & .806 & 24.90 & .620 & .376 & 10.29 & .108 & .892 & 25.62 & .682 & .284 & 12.82 & .782 & .343 & 4.55 \\
NeuralDiff \citep{neuraldiff}  &  .065 &  .952 & 31.89 &  .098 &  .921 & 25.93 &  .117 &  .910 & 29.02 &  .112 &  .722 & 24.42 &  .565 &  .652 & 20.09 \\
\ddnerf~\citep{ddnerf} & {.062} & {.975} & {34.27} & {.090} & {.953} & {26.27} & {.076} & {.979} & {34.14} & {.095} &  .707 & {24.63} & {.076} & {.979} & {36.58}       \\
\textbf{\RobustNeRF} & \bf{.013} & \bf{.988} & \bf{37.73} & \bf{.063} & \bf{.957} & \bf{26.31} & \bf{.006} & \bf{.995} & \bf{41.82} & \bf{.007} & \bf{.992} & \bf{41.23} & \bf{.018} & \bf{.990} & \bf{38.95} \\ 
\end{tabular}
} %
\\[.25em]
\includegraphics[width=\linewidth]{robustnerf/fig/\currfilebase}
\vspace{-.94em}
\captionof{figure}{
\textbf{Evaluations on \ddnerf Synthetic Scenes --}
Quantitative and qualitative evaluations on the Kubric synthetic dataset introduced by \ddnerf, consisting of 200 training frames (with distractor) and 100 novel views for evaluation (without distractor).
}
\vspace*{-0.275cm}
\label{fig:\currfilebase}
\end{figure*}

We evaluate \RobustNeRF on its ability to \textit{ignore} distractors while accurately reconstructing the static elements of a scene.
We train \RobustNeRF, \DDNeRF, and variants of \mipNeRFthreesixty on scenes where distraction-free frames are available.
Models are \textit{trained} on frames with distractors and \textit{evaluated} on distractor-free frames.

\vspace{-0.1cm}
\paragraph{Comparison to \mipNeRFthreesixty{} -- \autoref{fig:sota}}
On natural scenes, \RobustNeRF generally outperforms variants of \mipNeRFthreesixty by 1.3 to 4.7 dB in PSNR.
As $L_2$, $L_1$, and Charbonnier losses weigh all pixels equally, the model is forced to represent, rather than ignore, distractors as ``clouds'' with view-dependent appearance.
We find clouds to be most apparent when distractors remain stationary for multiple frames.
In contrast, \RobustNeRF's loss isolates distractor pixels and assigns them a weight of zero (see \autoref{fig:residuals}).
To establish an upper bound on reconstruction accuracy, we train \mipNeRFthreesixty with Charbonnier loss on distraction-free versions of each scene,
the images for which are taken from (approximately) the same viewpoints.
Reassuringly, \RobustNeRF when trained on distraction-free frames, achieves nearly identical accuracy; see \autoref{fig:limitations}.

While \RobustNeRF consistently outperforms \mipNeRFthreesixty, the gap is smaller in the Apartment scenes (Statue, Android) than the Robotics Lab scenes (Crab, BabyYoda).
This can be explained by challenging background geometry, errors in camera parameter estimation, and imperceptible changes to scene appearance.
In experiments on natural scenes, as reported in \autoref{fig:sota}, the performance gap between \mipNeRFthreesixty (Ch.) and \RobustNeRF is markedly higher for the two scenes captured in the robotics lab (i.e., Crab, BabyYoda), compared to those in the apartment (i.e., Statue, Android).
We attribute this to the difficulty in reconstructing the apartment scenes, regardless of the presence of distractors.
This statement is supported by metrics for reconstruction quality of a \mipNeRFthreesixty model trained on clean, distractor-free photos.
In particular, while \mipNeRFthreesixty achieves over 32 dB PSNR on Crab and BabyYoda scenes, its PSNR is nearly 10 dB lower on Statue and Android.

Upon closer inspection of the photos and our reconstructions, we identified several reasons for this.
First, the apartment scenes contain non-trivial background content with 3D structure.
As the background was not the focus of these captures, background content is poorly reconstructed by all models considered.
Second, background content illuminated by sunlight is overexposed in some test images (see \autoref{fig:residual_android}).
While this challenge has already been addressed by RawNeRF~\citep{rawnerf}, we do not address it here as it is not a focus of this work.
Lastly, we find that some static objects were unintentionally moved during our capture.
The most challenging form of this is the movement of a table cloth prominently featured in the Android scene which lead to perturbed camera parameter estimates (e.g., see \autoref{fig:residual_android}).

\vspace{-0.05cm}
\paragraph{Comparison to \ddnerf~-- \autoref{fig:sotaddnerf}}
Quantitatively, \RobustNeRF matches or outperforms \DDNeRF by as much as 12 dB PSNR depending on the number of unique outlier objects in the capture. Results on
\ddnerf real scenes are provided in the supplementary material for qualitative comparison.
In Statue and Android, 1 and 3 non-rigid objects are moved around the scene, respectively.
\DDNeRF is able to model these objects and thus separate them from the scenes' static content.
In the remaining scenes, a much larger pool of 100 to 150 unique, non-static objects are used -- too many for \DDNeRF to model effectively.
As a result, ``cloud'' artifacts appear in its static representation, similar to those produced by \mipNeRFthreesixty.
In contrast, \RobustNeRF identifies non-static content as outliers and omits it during reconstruction.
Although both methods use a similar number of parameters, \DDNeRF's peak memory usage is 2.3x higher than \RobustNeRF and 37x higher when normalizing for batch size.
This is a direct consequence of model architecture: \DDNeRF is tailored to simultaneously modeling static and dynamic content and thus merits higher complexity.
To remain comparable, we limit image resolution to 0.2 megapixels for all experiments. 

\begin{figure}
\centering

\begin{tabular}{@{}l|ccc|ccc@{}}
    & \multicolumn{3}{c|}{\color{red}\bf Crab} 
    & \multicolumn{3}{c}{\color{blue}\bf BabyYoda} 
\\
                             & \lpips & \ssim & \psnr  & \lpips & \ssim & \psnr  \\
\midrule
Order 1                      &  0.43 & 0.66  & 20.19  &  0.44  &  0.66  &  18.17      \\
Order 2                    &  0.42 & 0.68  & 20.95  & 0.44    &  0.66  &  17.13      \\
\midrule
\end{tabular}

\includegraphics[width=.7\linewidth]{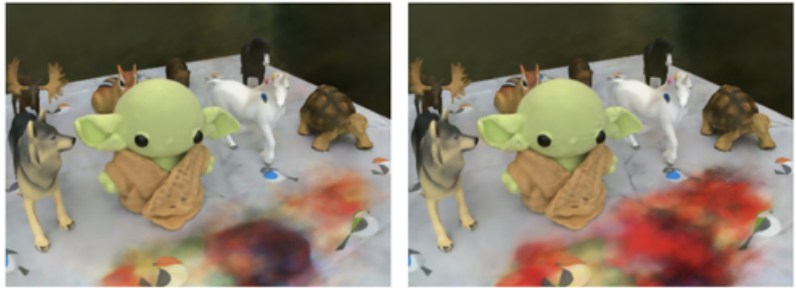}
\caption{
\textbf{Effect of Image Order on \DDNeRF} --
As this model is based on space-time NeRFs~\citep{hypernerf}, to make it compatible with our setting we create a 'temporal' indexing of the photos. 
Here, we visualize:
(left) with our heuristic ordering;
(right) with another random order.  We observe similar distractor-related artifacts in both cases.
}
\vspace{-.5cm}
\label{fig:\currfilebase}
\end{figure}

\begin{figure}
\centering

\begin{tabular}{@{}l|ccc|ccc@{}}
    & \multicolumn{3}{c|}{\color{brown}\bf Statue} 
    & \multicolumn{3}{c}{\color{red}\bf Crab} 
\\
                    & \lpips & \ssim & \psnr  & \lpips & \ssim & \psnr  \\
\midrule
Config 1            & 0.48  & 0.49  & 19.09  &  0.42  &  0.68  &  21.18      \\
Config 2            & 0.49  & 0.48  & 18.20  & 0.51  & 0.59  & 17.02 \\
Config 3            & 0.51  & 0.47  & 18.28  & 0.46  & 0.63  & 19.01 \\
Config 4            & 0.49  & 0.48  & 18.18  & 0.49  & 0.58  & 16.77 \\
\midrule
\end{tabular}
\\ [1ex]

\begin{tabular}{c | c c c c c} 
\hline
Config $\#$ & $k$ & $\lambda_s$ & $\lambda_r$ & $\lambda_{\sigma^s}$  & $\lambda_\rho$ \\ [0.5ex] 
\hline
Config 1 & 1.75 & 1e-4 $\rightarrow$ 1e-2 & 1e-3 & 0 & 1e-1 \\ 
Config 2 & 3 & 1e-4 $\Rightarrow$ 1 & 1e-3 & 0 & 1e-1 \\
Config 3 & 2.75 & 1e-5 $\Rightarrow$ 1 & 1e-3 & 0 & - \\
Config 4 & 2.875 & 5e-4 $\Rightarrow$ 1 & 0 & 0 & - \\ 
\hline
\end{tabular}

\caption{\textbf{\ddnerf HParam Tuning} --
The performance of \ddnerf{} is heavily influenced by the choice of hyperparameters. In particular, optimal choices of hyperparameters are noted to be strongly influenced by the amount of object and camera motion, as well as video length. We tune by applying four recommended configurations, and identify the first as optimal across the Statue and Crab datasets. Please note that $\rightarrow$ indicates linear increase in value and $\Rightarrow$ indicates exponential increase in value.
}
\label{fig:\currfilebase}
\end{figure}

Unlike \RobustNeRF, \ddnerf makes use of a time signal in the form of provided appearance and warp IDs to generate a code as additional input to the HyperNeRF model.
This explicitly models dynamic content alongside the static component of the scene. Two assumptions of \ddnerf are broken in our datasets: 1) the objects sporadically appear (by design); and 2) the views are not necessarily captured in a video-like order. Sporadic object appearance is central to our task, so we do not ablate this property. 
However, we do evaluate the effect of heuristically reordering camera views according to z position and radial angle of the robotic arm, thereby producing an image order for an imagined "continuous" path.
As a control, we pseudorandomly scramble the view order, and train \ddnerf in both settings. The results for BabyYoda and Crab can be seen in \autoref{fig:ddnerf}. We observe no consistent discernable improvement in performance as a result of view reordering and hypothesize that the major hurdle for \ddnerf is rather the modeling of sporadic artifacts. 

We also evaluate the effect of applying the four hyperparameter configurations provided by \ddnerf \citep{ddnerf}. 
We observe, as expected, that the first configuration performs best across our datasets.
Due to limited access to appropriate compute architecture for \ddnerf, we were not able to tune every scene, but selected configuration 1 for all experiments as it performed best in 10/16 real world scenes for \ddnerf as well as tuning experiments on two of our example datasets as see in in \autoref{fig:ddnerf_tune}.

\begin{figure}[t]
\centering
\begin{tabular}{@{}lcccc@{}}
& \lpips  & \ssim & \psnr & Updates to PSNR=30
\\ \midrule
\mipNeRFthreesixty ($L_2$) &  0.31 &  0.75    &  22.97   &   --\\
+ robust~\eq{trimming}  &  0.39 & 0.60   &  18.21    &     --\\
+ smoothing~\eq{smoothing} &   0.22 &  0.81   & 30.01    &   250K \\
+ patching~\eq{patching} & \textbf{0.21} & \textbf{0.81}   &  \textbf{30.75}    &   70K  \\
\midrule
oracle (clean) & 0.16 & 0.84 & 32.55 & 25K \\
\end{tabular}
\includegraphics[width=.7\linewidth]{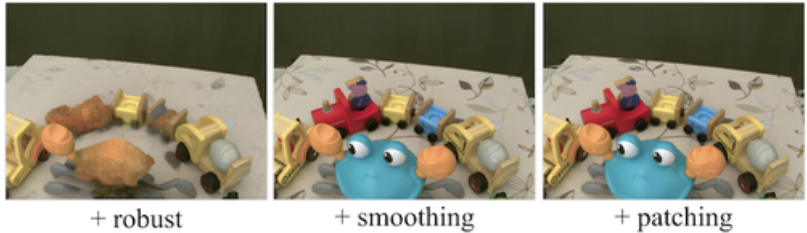}
\captionof{figure}{\textbf{Ablations --}
Blindly trimming the loss causes details to be lost.
Smoothing recovers fine-grained detail, while patch-based evaluation speeds up training and adds more detail.
Patching enables the model to reach PSNR of $30$, almost $4\times$ faster.
}
\vspace*{-0.25cm}
\label{fig:\currfilebase}
\end{figure}

\subsection{Ablations -- \autoref{fig:ablations}}
We ablate elements of the \RobustNeRF loss on the crab scene, comparing to an upper bound on the reconstruction accuracy of  \mipNeRFthreesixty trained on distractor-free (clean) images from identical viewpoints.
Our trimmed estimator
(\ref{eq:trimming}) 
successfully eliminates distractors at the expense of high frequency texture and a lower PSNR.
With smoothing
(\ref{eq:smoothing}), 
fine details are recovered, at the cost of longer training times.
With the spatial window
(\ref{eq:patching}),
\RobustNeRF training time is on-par with \mipNeRFthreesixty.
We also ablate patch size and the trimming threshold (see Supplementary Material); we find that \RobustNeRF is
insensitive to trimming threshold, and that
reducing the patch size offsets the gains from smoothing and patching.

\begin{figure}[ht]
\vspace*{-0.185cm}
\centering
\includegraphics[width=0.7\linewidth]{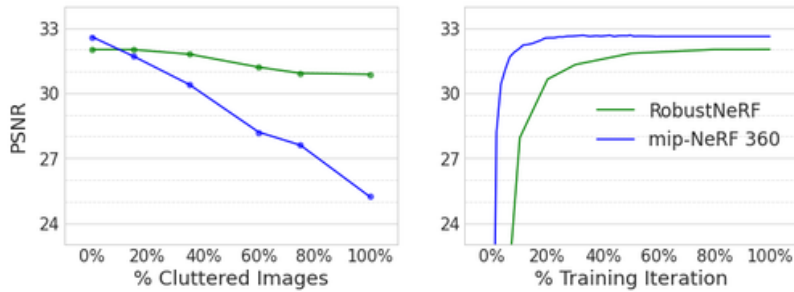}
\caption{
\textbf{Sensitivity and Limitations -- }
(left) Reconstruction accuracy for BabyYoda as we increase the fraction of train images with  distractors.
(right) Accuracy vs training time on \textit{clean} BabyYoda images~(distractor-free).
}
\vspace*{-0.255cm}
\label{fig:\currfilebase}
\end{figure}

\subsubsection{Distribution Sensitivity -- \autoref{fig:limitations}}
We find that \RobustNeRF is remarkably robust to the amount of clutter in a dataset.
We define an image as ``cluttered'' if it contains some number of distractor  pixels.
The figure shows how  the reconstruction accuracy of \RobustNeRF and \mipNeRFthreesixty depends on the fraction of training images with distractors, keeping the training set size constant.
As the fraction increases, \mipNeRFthreesixty's accuracy steadily drops from 33 to 25 dB, while \RobustNeRF's remains steadily above 31 dB throughout.
In the distraction-free regime, we find that \RobustNeRF mildly under-performs \mipNeRFthreesixty, both in reconstruction quality and the time needed for training.
This follows from the statistical inefficiency induced by the trimmed estimator~\eq{trimming}, for which a percentage of pixels will be discarded even if they do not correspond to distractors.

\subsubsection{Sensitivity to Hyperparameters}
\label{app:sec:sensitivity_to_hyperparameters}
\begin{figure}
\centering
\includegraphics[width=0.45\linewidth]{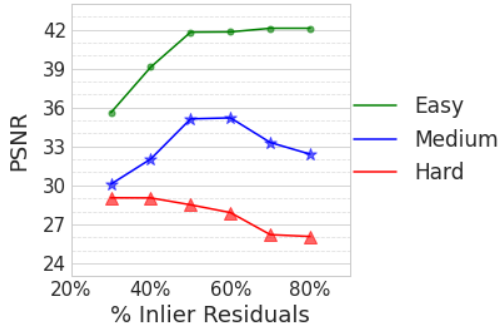}
\caption{
    \textbf{Sensitivity to $\threshold_\residual$ --}
    \RobustNeRF's reconstruction quality as a function of $\threshold_\residual$ on scenes with different inlier/outlier proportions.
    Overestimating $\threshold_{\residual}$ increases training time without affecting final reconstruction accuracy.
}
\label{fig:\currfilebase}
\end{figure}

We find that the choice of thresholds and filter sizes, described in Section~\ref{sec:method}, suffices for a wide range of datasets.
As long as the threshold $\threshold_\residual$ is greater than the proportion of outlier pixels in a dataset, \RobustNeRF will reliably identify and ignore outlier pixels; see \autoref{fig:sensitivity_to_outliers}. Easy has less than $10\%$ outlier pixels so any  $\threshold_\residual$ less than $80\%$ works. In the medium case at least a $\threshold_\residual$ of $60\%$ is required to remove  the outliers. In the hard case $44\%$ of pixels are on average occupied so any  $\threshold_\residual$ above $50\%$ has worse results. Training with less than $50\%$ of the loss slows down training significantly. Therefore, we observe that after the 250k iterations the model has not converged yet. On average training with $30\%$ of loss requires twice the number of training iterations to catch up.
In contrast, \DDNeRF requires careful, manual hyperparameter tuning for each scene (e.g., see \autoref{fig:ddnerf_tune}) for several hyperparameters.
In our experiments, we found that a single setting of neighborhood and patch sizes works well across all scenes.
We present model performance as a function of both hyperparameters on Crab in \autoref{tab:rebuttal_ablations}.
Larger neighborhood sizes are better regularizers, and we are bounded by the amount of device memory available.

\begin{figure}[t]
\small
\centering
\begin{tabular}{@{}lccccc@{}}
    Neigh./Patch & 4/2 & 8/4   & 16/2  & 16/4  & 16/8  \\
    \midrule
    $\threshold_\mathcal{R}=0.6$       & 18.3  & 23.08 & 30.22 & 30.35 & 30.75 \\
    $\threshold_\mathcal{R}=0.8$       & 28.28  & 30.7    & 30.72    & 30.69    & 30.72    \\
\end{tabular}
\\
\captionof{figure}{
    \textbf{Sensitivity to hyper-parameters.}
    PSNR on distractor-free frames on the Crab dataset as a function of \RobustNeRF's neighborhood size, patch size, and $\threshold_\mathcal{R}$.
}
\vspace*{-0.4cm}
\label{tab:\currfilebase}
\end{figure}

\subsubsection{View-dependent effects}
\label{app:sec:view_dependent}

We experimentally observed that \RobustNeRF performs similarly to \mipNeRFthreesixty in reconstructing scenes with non-Lambertian materials, semi-transparent objects, and soft shadows.
These phenomena are present in the Statue scene (tabletop is \textit{glossy}), and the toys in the Crab and BabyYoda scenes which cast soft \emph{shadows}.

To further emphasize these qualities, we  include results for additional scenes with glass, metallic, and reflective objects in  \autoref{fig:rebuttal_viewdep}. The first scene is captured with our Robotic rig, similar to Crab and BabyYoda scenes.  It contains a mirror, a shiny cylinder, a transparent vase and a glossy ceramic mug. The other two datasets are captured in the wild. One is with a mirror while pedestrians are moving (as distractions). The last scene  contains a transparent pitcher as the object of interest, while the photographer's body parts appear in the photos as the distractors.

\begin{figure}[t]
\centering

\includegraphics[width=.6\linewidth]{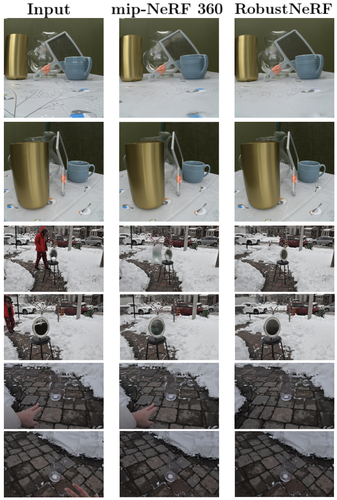}
\captionof{figure}{
    \textbf{Qualitative results on scenes with view-dependent effects.}
    \RobustNeRF naturally captures view-dependent effects in scenes with (3rd-6th rows) and without (1st and 2nd row) distractors.
}
\vspace*{-0.2cm}
\label{fig:\currfilebase}
\end{figure}

\subsection{More Qualitative Results}
\label{app:sec:sota_more}

We render images from different NerF models from more  viewpoints from each of our datasets to further expand the comparison with baselines, \ddnerf, and \RobustNeRF. Looking at Figures  \ref{fig:bigsota_statue} through
\ref{fig:bigsota_yoda} one can see that \ddnerf is only able to remove the outliers when there is a single distractor object (Statue dataset) and it fails on the other three datasets. The Android dataset has three wooden robots with articulated joints as distractors, and even in this setup where the texture of the distractor objects are similar to one another, \ddnerf fails to fully remove the outliers. In comparison, \RobustNeRF is able to remove the outliers irrespective of their number and diversity. 

For all four datasets \mipNeRFthreesixty, with either L1, L2, or Charbonnier loss, fails to detect the outliers;
one can see 'clouds' or even distinct floaters for these methods. The worst performing loss is L2, as expected. L1 and Charbonnier behave similarly in terms of outlier removal. Changing the loss to \RobustNeRF eliminates the  floaters and artifacts in all datasets. Video renderings for these scenes are also included in the zipfile with the supplementary material. The floaters in \mipNeRFthreesixty are  easier to resolve in the videos.

We have also experimented on the \ddnerf natural scenes in ~\cite{ddnerf}.
The qualitative samples are shown in \autoref{fig:rebuttal_d2nerf}.
We find that \RobustNeRF produces plausible, distraction-free models. 

\begin{figure}[t]
\centering

\includegraphics[width=.8\linewidth]{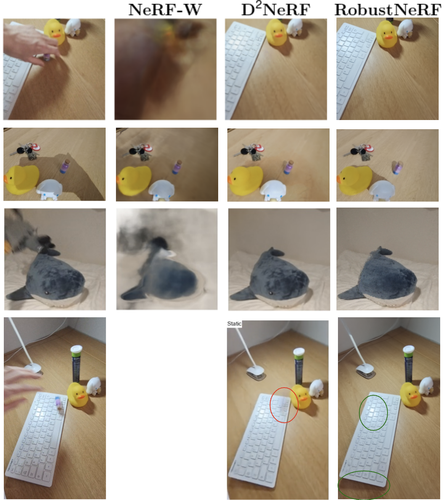}
\captionof{figure}{
\textbf{Qualitative results on \ddnerf Pick scene.}
Renders of static model components. 
Results for \NeRFW and \ddnerf are provided by \cite{ddnerf}. %
Note how \RobustNeRF naturally captures specular reflections and shadows (green, right).
}
\vspace*{-0.2cm}
\label{fig:\currfilebase}
\end{figure}

\clearpage
\begin{figure*}[t!]
\centering
\includegraphics[width=\textwidth]{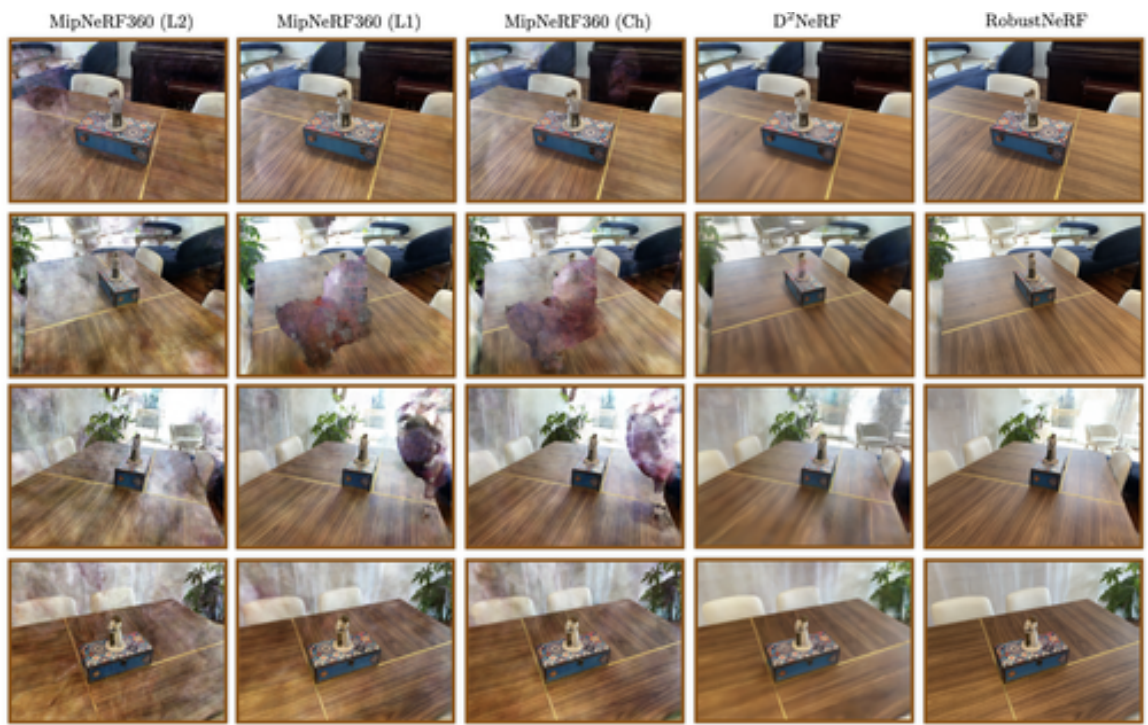}
\caption{
    \textbf{Statue --}
    Qualitative results on Statue. 
    It is helpful to zoom in to see details.
}
\label{fig:\currfilebase}
\end{figure*}

\begin{figure*}[b!]
\centering
\includegraphics[width=\textwidth]{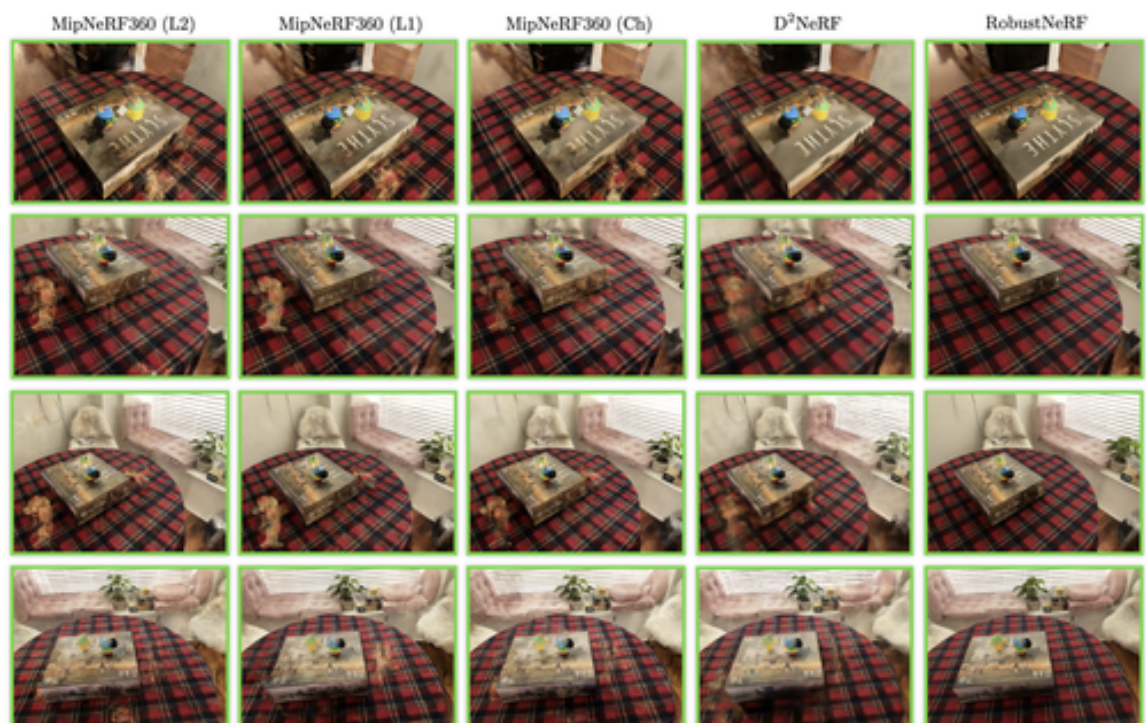}
\caption{
    \textbf{Android --}
    Qualitative results on Android.
    It is helpful to zoom in to see details.
}
\label{fig:\currfilebase}
\end{figure*}

\begin{figure*}[t]
\centering
\includegraphics[width=\textwidth]{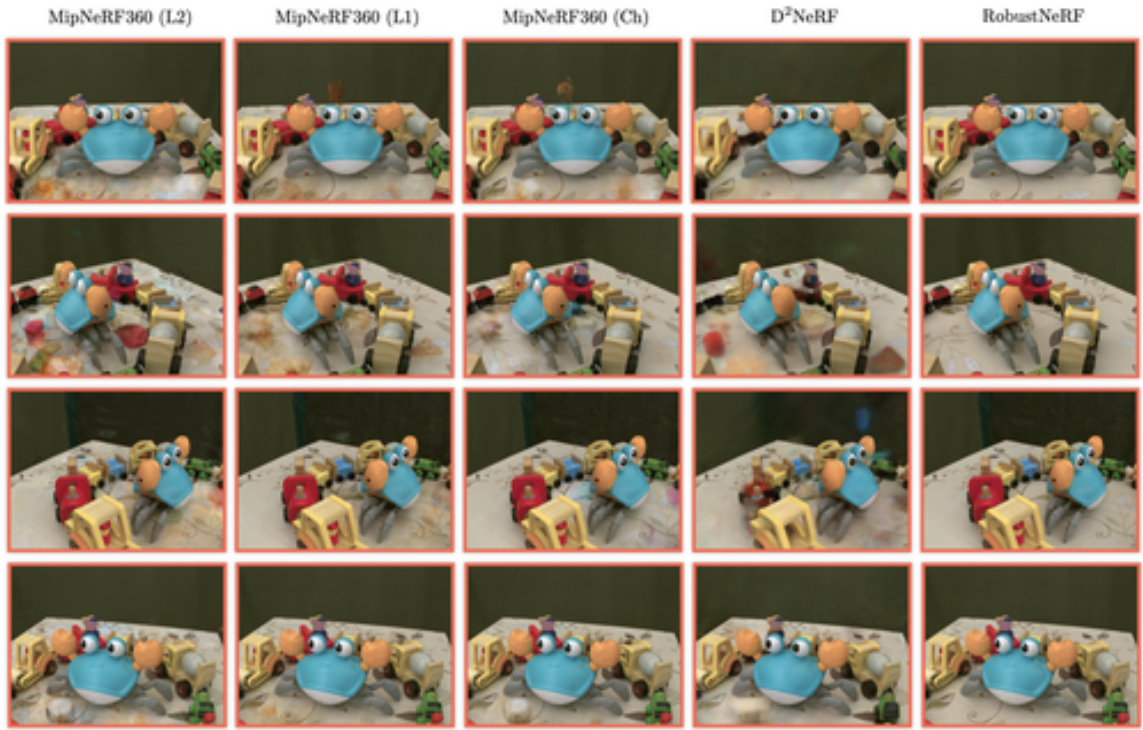}
\caption{
    \textbf{Crab --}
    Qualitative results on Crab. 
    It is helpful to zoom in to see details.
}
\label{fig:\currfilebase}
\end{figure*}

\begin{figure*}[b]
\centering
\includegraphics[width=\textwidth]{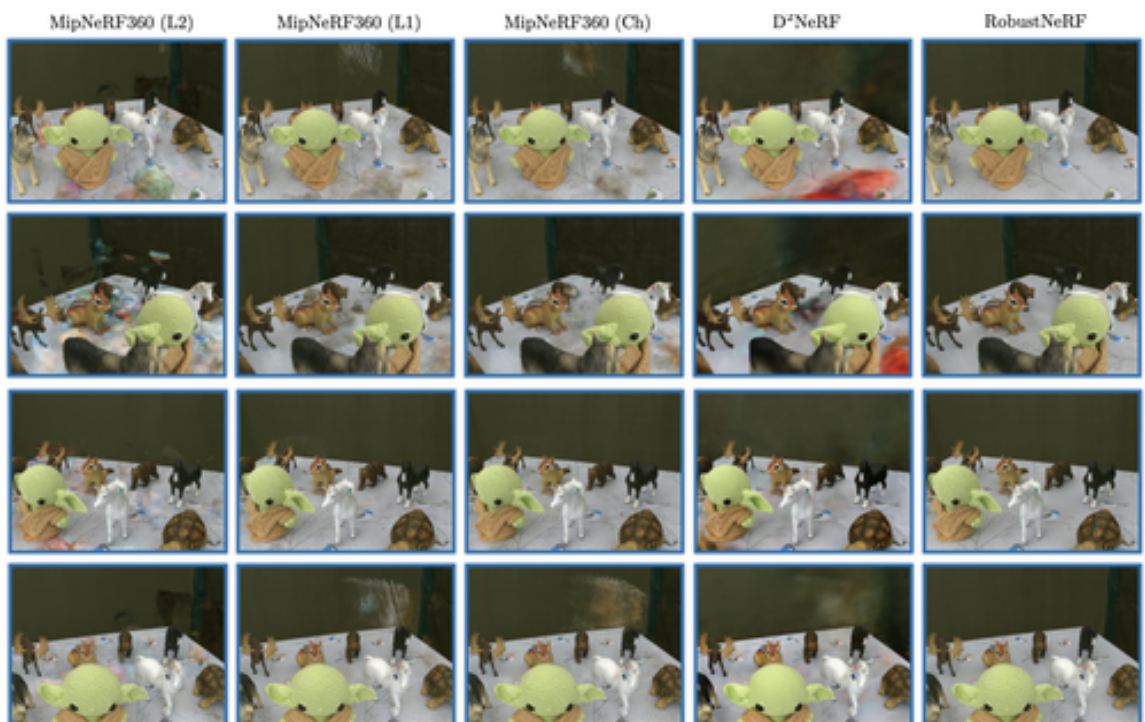}
\caption{
    \textbf{BabyYoda --}
    Qualitative results on BabyYoda. 
    It is helpful to zoom in to see details.
}
\label{fig:\currfilebase}
\end{figure*}

\color{black}

\vspace*{-0.1cm}
\section{Conclusions}
\label{sec:conclusions}
\vspace*{-0.1cm}

We address a central problem in training NeRF models, namely, optimization in the presence of distractors, such as  transient or moving objects and photometric phenomena that are not persistent throughout the capture session.

Viewed through the lens of robust estimation, 
we formulate training as a form of iteratively re-weighted least squares, with  
a variant of trimmed LS, and an inductive bias on the smoothness of the outlier process.
\RobustNeRF is surprisingly simple, yet effective on a wide range of datasets.
\RobustNeRF is shown to outperform recent state-of-the-art methods~\citep{mipnerf360, ddnerf}, qualitatively and quantitatively, on a suite of synthetic datasets, common benchmark datasets, and new datasets captured by a robot, allowing fine-grained control over distractors for comparison with previous methods.
While our experiments explore robust estimation in the context of  \mipNeRFthreesixty, the \RobustNeRF loss can be incorporated within other NeRF models.

\vspace*{-0.11cm}
\paragraph{Limitations} 
While \RobustNeRF performs well on scenes with distractors, the loss entails some statistical inefficiency.
On clean data, this yields somewhat poorer reconstructions, often taking longer to train~(see \autoref{fig:limitations}).
Future work will consider very small distractors,
which may require adaptation of the spatial support used for outlier/inlier decisions.
It would also be interesting to learn a neural weight function, further improving \RobustNeRF;  active learning may be useful in this context.
Finally, it would be interesting to include our robust loss in other NeRF frameworks.

\chapter{\methodname: Ignoring Distractors in 3D Gaussian Splatting
}
\label{chap:spotlesssplats}

3D Gaussian Splatting (3DGS) is a promising technique for 3D reconstruction, offering efficient training and rendering speeds, making it suitable for real-time applications. However, current methods require highly controlled environments—no moving people or wind-blown elements, and consistent lighting—to meet the inter-view consistency assumption of 3DGS. This makes reconstruction of real-world captures problematic. We present SpotLessSplats, an approach that leverages pre-trained and general-purpose features coupled with robust optimization to effectively ignore transient distractors.
Our method achieves state-of-the-art reconstruction quality both visually and quantitatively, on casual captures.

\begin{figure}
\includegraphics[width=\textwidth]{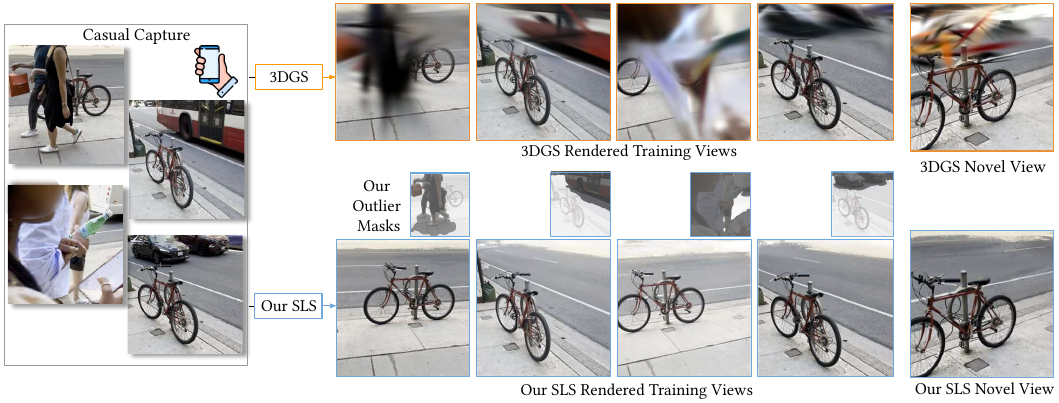}
\caption{{\methodname} cleanly reconstructs a scene with many transient occluders (\textbf{middle}), while avoiding artifacts~(\textbf{bottom}).
It correctly identifies and masks out all transients~(\textbf{top}), even in captures with a large number of them~(\textbf{left}).}
\label{fig:teaser}
\end{figure}

\section{Introduction}
\label{sec:introduction}

The reconstruction of 3D scenes from 2D images with neural radiance fields (NeRF)~\citep{Mildenhall20eccv_nerf} and, more recently, with 3D Gaussian Splatting (3DGS)~\citep{Kerbl2023tdgs}, has been the subject of intense focus in vision research.
Most current methods assume that images are simultaneously captured, perfectly posed, and noise-free.
While these assumptions simplify 3D reconstruction, they rarely hold in real-world, where moving objects (e.g., people or pets), lighting variations, and other spurious photometric inconsistencies degrade performance, limiting widespread application.

In NeRF training, robustness to outliers has been incorporated by 
down-weighting or discarding inconsistent observations based on the magnitude of color residuals \citep{Wu2022d2nerf, Sabour2023robustnerf, MartinBrualla21cvpr_nerfw, chen2024nerfhugs}.
Similar methods adapted to 3DGS \citep{Dahmani2024swag, Kulhanek2024wildgaussians, Wang2024wegs} address global appearance changes and single-frame transients seen in datasets like Phototourism~\citep{phototourism2006}.
Such captures include appearance changes occurring over weeks and different times of day, which are not common in most casual captures.
For 3DGS in particular, the \textit{adaptive densification} process itself introduces variance in color residuals, compromising detection of transients when directly applying existing ideas from robust NeRF frameworks.

In this chapter we introduce \methodname~(\methodshortname), a framework for robust 3D scene reconstruction with 3DGS, {via unsupervised detection of outliers in training images}.
Rather than detecting outliers in RGB space, we instead utilize a richer, \textit{learned feature space} from text-to-image models.
The meaningful semantic structure of this feature embedding allows one to more easily detect the spatial support of structured outliers associated, for example, with a single object.
Rather than employing manually-specified robust kernels for outlier identification~\citep{Sabour2023robustnerf}, we instead exploit adaptive methods in this feature space to detect outliers.
To this end we consider two approaches within this framework.
The first uses  non-parametric clustering of local feature embeddings as a simple way to find image regions of structured outliers.
The second uses an MLP, trained in an unsupervised fashion to predict the portion of the feature space that is likely to be associated with distractors.
We further introduce a (complementary and general purpose) sparsification strategy, compatible with our robust optimization, that delivers similar reconstruction quality with two to four times fewer splats, even on distractor-free datasets, yielding significant savings in compute and memory.
Through experiments on challenging benchmarks of casually captured scenes~\citep{Sabour2023robustnerf, Ren2024nerfonthego}, \methodshortname is shown to consistently outperform competing methods in reconstruction accuracy.

\vspace{1em}

\noindent
Our key contributions include:
\begin{itemize}
    \item {An adaptive, robust loss, leveraging text-to-image diffusion features, that reliably identifies transient distractors in causal captures, eliminating issues of overfitting to photometric errors.}
    \item A novel sparsification method compatible with our robust loss that significantly reduces the number of Gaussians, saving compute and memory without loss of fidelity.
    \item Comprehensive evaluation of \methodshortname on standard benchmarks, demonstrating SOTA robust reconstruction, outperforming existing methods by a substantial margin.
\end{itemize}

\section{Related work}
\label{sec:related}

Neural Radiance Fields (NeRF) \citep{Mildenhall20eccv_nerf}, have gained widespread attention due to the high quality reconstruction and novel view synthesis of 3D scenes.
NeRF represents the scene as a view dependent emissive volume.
The volume is rendered using the absorption-emission part of the volume rendering equation~\citep{Kajiya1984rtv}.
Multiple enhancements have followed.
Fast training and inference~\citep{Sun2022dvgo, Muller2022ingp, Yu21iccv_PlenOctrees, mobilenerf}, training with limited or single view(s)~\citep{Yu21cvpr_pixelNeRF, Jain21iccv_DietNeRF, Rebain22lolnerf} and simultaneous pose inference~\citep{Lin21iccv_BARF, Wang21arxiv_NeRFminusminus, Levy2024melon} have brought radiance fields closer to practical applications.
More recently, 3D Gaussian Splatting (3DGS)~\citep{Kerbl2023tdgs} was proposed as a primitive-based alternative to NeRFs with significantly faster rendering speed, while maintaining high quality.
3D Gaussians can be efficiently rasterized using alpha blending~\citep{Zwicker2001splatting}. This simplified representation takes advantage of modern GPU hardware to facilitate real-time rendering.
The efficiency and simplicity of 3DGS have prompted a shift in focus within the field, with many NeRF enhancements being quickly ported to 3DGS~\citep{Yu2024MipSplatting, charatan2024PixelSplat}.

\begin{figure}[t]
\centering
\includegraphics[width=0.85\linewidth]{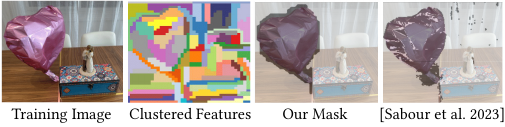}
\captionof{figure}{
Our outlier classification using clustered semantic features covers the distractor balloon fully, but an adapted robust mask from~\cite{Sabour2023robustnerf} misclassifies pixels with similar color to background, as inliers.
}
\label{fig:clustering}
\end{figure}

\paragraph{Robustness in NeRF}
The original NeRF paper made strong assumptions regarding the capture setup: the scene needs to be perfectly static, and the illumination should stay unchanged throughout the capture.
More recently, NeRF has been extended to train on unstructured~``in-the-wild'' captured images that violate these constraints.
Two influential works, NeRF-W~\citep{nerfw} and RobustNeRF~\citep{Sabour2023robustnerf} addressed the problem of transient distractors, both using photometric error as guidance.
NeRF-W~\citep{nerfw} models a 3D uncertainty field rendered to 2D outlier masks that down-weight the loss at pixels with high-error, and a regularizer that prevents degenerate solutions.
NeRF-W~\citep{nerfw} also models global appearance via learned embeddings, which are useful for images captured over widely varying lighting and atmospheric conditions. 
Urban Radiance Fields (URF)~\citep{Rematas2022urf} and Block-NeRF~\citep{Tancik2022blocknerf} similarly apply learned appearance embeddings to large-scale reconstruction.
HA-NeRF~\citep{Chen2022hanerf} and Cross-Ray~\citep{Yang2023crossray} model 2D outlier masks instead of 3D fields, leveraging CNNs or transformers for cross-ray correlations.

RobustNeRF~\citep{Sabour2023robustnerf}, approached the problem from a robust estimator perspective, with binary weights determined by thresholded rendering error, and a blur kernel to reflect the assumption that pixels belonging to distractors are spatially correlated.
However, both RobustNeRF and NeRF-W variants~\citep{Chen2022hanerf, Yang2023crossray} rely solely on RGB residual errors and because of this they often misclassify transients with colors similar to their background; see RobustMask in~\Cref{fig:clustering}.
To avoid this, previous methods require  careful tuning of hyper-parameters, i.e., the blur kernel size and thresholds in RobustNeRF and the regularizer weight in NeRF-W.
On the contrary, our method uses the rich representation of text-to-image models for semantic outlier modeling. This avoids direct RGB error supervision, as it relies on feature-space similarities for clustering.

\begin{figure}[t]
\centering
\setlength\tabcolsep{1.8pt}
\includegraphics[width=0.75\linewidth]{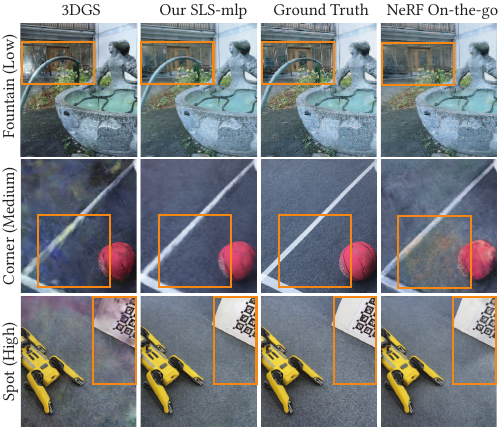}\\

\begin{tabular}{l|ccc|ccc|ccc}
 & \multicolumn{3}{c|}{Fountain} & \multicolumn{3}{c|}{Corner} & \multicolumn{3}{c}{Spot}\\
 & \psnr & \ssim & \lpips & \psnr & \ssim & \lpips & \psnr & \ssim & \lpips  \\ \hline
MipNerf360 & 13.91 & 0.29 & 0.55 & 20.41 & 0.66 & 0.34  & 17.82 & 0.30 & 0.46 \\
RobustNerf & 17.20 & 0.41 & 0.54 & 20.21 & 0.70 & 0.35 & 16.40 & 0.38 & 0.69 \\
NeRF On-the-go & 20.11 & 0.61 & 0.31 & 24.22 & 0.80 & 0.19 & 23.33 & 0.79 & 0.19 \\ \hline
\tdgsshortname & 21.70 & 0.79 & 0.16 & 24.05 & 0.86 & 0.13 & 20.72 & 0.76 & 0.31 \\
Our \methodsdmlp & \textbf{22.81} & \textbf{0.80} & \textbf{0.15} & \textbf{26.43} & \textbf{0.90} & \textbf{0.10} & \textbf{25.76} & \textbf{0.90} & \textbf{0.12}
\end{tabular}
\captionof{figure}{
Our method accurately reconstructs scenes with different levels of transient occlusion, avoiding leakage of transients or under-reconstruction evident by the quantitative and qualitative results on NeRF On-the-go~\citep{Ren2024nerfonthego} dataset.
}
\label{fig:nerfonthego1}
\end{figure}

NeRF On-the-go~\citep{Ren2024nerfonthego} released a dataset of casually captured videos with transient occluders. 
Similar to our method, it uses semantic semantic features from DINOv2~\citep{Oquab2023dinov2} to predict outlier masks via a small MLP. 
However, it also relies on direct supervision from the structural rendering error, leading to potential over- or under-masking of outliers. 
This is illustrated in~\Cref{fig:nerfonthego1}, where over-masking has removed the hose (`Fountain') and has smoothed the carpet (`Spot'), while under-masking caused distractor leaks and foggy artifacts (`Corner' and `Spot').
NeRF-HuGS~\citep{chen2024nerfhugs} combines heuristics from COLMAP's robust sparse point cloud~\citep{schoenberger2016sfm}, and off-the-shelf semantic segmentation to remove distractors.
Both heuristics are shown to fail under heavy transient occlusions in~\citep{Ren2024nerfonthego}.

\paragraph{Precomputed features}
The use of precomputed vision features, such as DINO~\citep{Caron2021dino, Oquab2023dinov2} have demonstrated the ability to generalize to multiple vision tasks.
Denoising Diffusion Probabalistic Models~\citep{song2019generative, ho2020denoising, Rombach2022stablediffusion}, known for their photorealistic image generation capabilities from text prompts~\citep{saharia2022photorealistic,ramesh2022hierarchical,rombach2021high}, have been shown to have internal features similarly powerful in generalizing over many tasks e.g. segmentation and keypoint correspondence~\citep{Amir2022deepvit, Tang2023emergent, Hedlin2024stablekeypoints, Zhang2023tale, Luo2023diffhyperfeatures}.

\paragraph{Robustness in 3DGS (concurrent works)}
Multiple concurrent works address 3DGS training on wild-captured data. 
SWAG~\citep{Dahmani2024swag} and GS-W~\citep{Zhang2024gsw} model appearance variation using learned global and local per-primitive appearance embeddings.
Similarly, WE-GS~\citep{Wang2024wegs} uses an image encoder to learn adaptations to the color parameters of each splat, per-image. Wild-GS~\citep{xu2024wildgs} learns a spatial triplane field for appearance embeddings.
All such methods~\citep{Zhang2024gsw, Wang2024wegs, xu2024wildgs} adopt an approach to outlier mask prediction like NeRF-W~\citep{nerfw}, with 2D outlier masks predicted to downweight high-error rendered pixels. SWAG \citep{Dahmani2024swag} learns a per-image opacity for each Gaussian, and denotes primitives with high opacity variance as transients.
{Notable are SWAG~\citep{Dahmani2024swag} and GS-W~\citep{Zhang2024gsw} that show no or little improvement over the local/global appearance modeling, when additional learned transient masks are applied to Phototourism scenes~\citep{phototourism2006}.}
SLS focuses on casual captures with longer duration transients and minimal appearance changes, common in video captures like those in the ``NeRF on-the-go'' dataset~\citep{Ren2024nerfonthego}.

\section{Background}
\label{sec:background}

We build our technique on top of 3D Gaussian Splatting~\citep{Kerbl2023tdgs}, or 3DGS for brevity, which represents a 3D scene as a collection of 3D anisotropic Gaussians~$\model{=} \{ \element_i \} $, henceforth referred to as splats.
Given a set of posed images $\{ \image_n \}_{n = 1}^{\nImages}$,  $\image_i \in \real^{H \times W}$ of a casually captured scene, we aim to learn a 3DGS reconstruction $\model$ of the scene.
Each splat $\element_i$, is defined by a mean $\mean_i$, a positive semi-definite covariance matrix $\covariance_i$, an opacity $\opacity_i$, and view dependent color parameterized by spherical harmonics coefficients~$\sphcoef_i$~\citep{Ramamoorthi2001sph}.

The 3D scene representation is rendered to screen space by rasterization.
The splat positions/means are rasterized to screen coordinates via classical projective geometry, while special care needs to be taken to rasterize the covariance matrix of each splat.
In particular, if we denote with $\viewxform$ the perspective transformation matrix, the projection of the 3D covariance to 2D screen space can be approximated following~\cite{Zwicker2001splatting} as $\tilde\covariance = \projjacobian \viewxform \covariance \viewxform^{T} \projjacobian^{T}$,
where $\projjacobian$ is the Jacobian of the projection matrix, which provides a linear approximation to the non-linear projection process.
To ensure $\covariance$ represents covariance throughout optimization~(i.e., positive semi-definite), the covariance matrix is parameterized as $\covariance = \rot\scale\scale^{T}\rot^{T}$,
where scale $\scale{=}\text{diag}(\mathbf{s})$ with $\mathbf{s}{\in}\real^3$, and rotation $\rot$ is computed from a unit Quaternion $q$.
Once splat positions and covariances in screen-spaces are computed, the image formation process executes volume rendering as alpha-blending, which in turn requires splat sorting along the view direction.
Unlike NeRF, which renders \textit{one pixel at a time}, 3DSG renders the \textit{entire image} in a single forward pass.

\subsection{Robust optimization of 3DGS}
\label{sec:robust}
Unlike typical capture data for 3DGS~\citep{Kerbl2023tdgs}, we do not assume the set of posed images~$\{ \image_n \}_{n = 1}^{\nImages}$ to be curated, but rather \textit{casually captured}.
That is, we \textit{do not} require images to be depictions of a perfectly 3D consistent and static world.
Following prior work, we (interchangeably) denote the portion of images that break these assumptions as \textit{distractors}~\citep{Sabour2023robustnerf} or \textit{transient effects}~\citep{MartinBrualla21cvpr_nerfw}.
And unlike prior works~\citep{kerbl2024hierarchical,MartinBrualla21cvpr_nerfw,Tancik2022blocknerf}, we do not make assumptions about the transient object class, appearance and/or shape.

We address this problem by taking inspiration from the pioneering work of~\cite{Sabour2023robustnerf} in RobustNeRF, which removes distractors by identifying the portion of input images that should be masked out in the optimization process.
The problem reduces to predicting (without supervision) inlier/outlier masks $\{ \mask_n \}_{n = 1}^{\nImages}$ for each training image, and optimizing the model via a \textit{masked} L1 loss:
\begin{equation}
    \argmin{\model} \sum_{n=1}^{\nImages}  \mask_{n}^{\step} \odot
    \| \image_{n} - \hat{\image}_{n}^{\step} \|_1.
    \label{eq:nvs}
\end{equation}
where $\hat{\image}_{n}^{\step}$ is a rendering of $\model$ at training iteration~$(t)$.
As in RobustNeRF~\citep{Sabour2023robustnerf}, transient effects can be detected by observing photometric inconsistencies during training; that is, image regions that are associated with a large loss value.
By denoting with $\residual^{\step}_n {=} \|\image_{n} - \hat{\image}_{n}^{\step}\|_1$ the image of residuals~(with a slight abuse of notation, as the 1-norm is executed \textit{pixel-wise}, along the color channel), the mask is computed as:
\begin{align}
    \mask_{n}^{\step} {=} \indicator\left\{\left (\indicator\{\residual_{n}^{\step} {>} \rho\} \circledast \filter \right) {>} 0.5 \right\}
    \!,\:
    P(\residual_{n}^{\step}{>}\rho) {=} \tau
    \label{eq:robustnerf}
\end{align}
where $\indicator$ is an indicator function returning $1$ if the predicate is true and $0$ otherwise, $\rho$ is a generalized median with $\tau$ being a hyper-parameter controlling the cut-off percentile\footnote{If $\tau{=}.5$ then $\rho{=}\text{median}(\residual_{n}^{\step})$}, and $\filter$ is a~(normalized) $3\times3$ box filter that performs a morphological dilation via convolution~($\circledast$).
Intuitively, RobustNeRF~\citep{Sabour2023robustnerf}, summarized by \cref{eq:robustnerf} above, extends a \textit{trimmed robust estimator}~\citep{trimmedicp} by assuming that inliers/outliers are \textit{spatially correlated}.
We found that directly applying ideas from~\citep{Sabour2023robustnerf} to 3DGS, even when not limited by cases of misleading color residual like those depicted in~\Cref{fig:clustering}, do not remove outliers effectively.
Rather, several adaptations are necessary in order to accommodate differences in the representation and training process of 3DGS; see~\Cref{sec:adaptations}.

\section{Method}
\label{sec:method}
The outlier mask in~\cref{eq:robustnerf} is built solely based on photometric errors in the novel view synthesis process.
Conversely, we propose to identify distractors based on their semantics, \textit{recognizing} their re-occurrence during the training process.
We consider semantics as \textit{feature maps} computed from a self-supervised 2D foundation model (e.g.~\cite{Tang2023emergent}).
The process of removing distractors from training images then becomes one of \ul{identifying the sub-space of features that are likely to cause large photometric errors}.
As an example, consider a dog walking around in an otherwise perfectly static scene.
We would like to design a system that {either spatially in each image (\cref{sec:agglomerative}) or more broadly, spatio-temporally in the dataset (\cref{sec:learnt})}, recognizes ``dog'' pixels as the likely cause of reconstruction problems, and automatically removes them from the optimization. 
Our method is designed to reduce  reliance on local color residuals for outlier detection and over-fitting to color errors, and instead emphasizing reliance on semantic feature similarities between pixels. We thus refer to our methods as ``clustering.''
In~\Cref{sec:semantic} we detail how to achieve this objective. 
In~\Cref{sec:adaptations} we then detail several key adjustments to adapt the ideas from RobustNeRF~\citep{Sabour2023robustnerf} to a 3DGS training regime; see~\Cref{sec:agglomerative,sec:learnt}.

\subsection{Recognizing distractors}
\label{sec:semantic}
Given the input images $\{ \image_n \}_{n = 1}^{\nImages}$, we pre-compute feature maps for each image using Stable Diffusion~\citep{Rombach2022stablediffusion} as proposed by~\cite{Tang2023emergent}, resulting in feature maps~$\{\features_n\}_{n=1}^{\nImages}$.
This pre-processing step is executed \textit{once} before our training process starts.
We then employ these feature maps to compute the inlier/outlier masks~$\mask^{\step}$; we drop the image index $n$ to simplify notation, as the training process involves one image per batch.
We now detail two different ways to detect outliers.

\subsubsection{Spatial clustering}
\label{sec:agglomerative}
In the pre-processing stage, we additionally perform  unsupervised clustering of image regions.
Similar to super-pixel techniques {\citep{li2015superpixel,ibrahim2020image}}, we over-segment the image into a fixed cardinality collection of $\numclusters$ spatially connected components; see `Clustered Features'~\cref{fig:clustering}.
In more detail, we execute agglomerative clustering~\citep{mullner2011agglomerative} on the feature map~$\features$, where each pixel is connected to its $8$ surrounding pixels.
We denote the clustering assignment of pixel $p$ into cluster $c$ as~$\cluster[c,p]{\in}\{0,1\}$, and clustering is initialized with every pixel in its own cluster.
Clusters are agglomerated greedily, collapsing those that cause the least amount of inter-cluster feature variance differential before/post collapse.
Clustering terminates when $\numclusters{=}100$ clusters remain.

We can then calculate the probability of cluster $c$ being an inlier from the percentage of its inlier pixels in \cref{eq:robustnerf}:
\begin{equation}
P(c \in \mask^{\step}) = \Bigl( \sum_p \cluster[c,p] \cdot \mask^{\step}[p] \Bigr) ~/~ {\sum_p \cluster[c,p]},
\label{eq:agglomerativeprob}
\end{equation}
and then propagate the cluster labels back to pixels as:
\begin{equation}
\mask^{\step}_\text{agg}(p) = \sum_c \indicator\{P(c \in \mask^{\step}) > 0.5\} \cdot \cluster[c,p]
\label{eq:agglomerative}
\end{equation}
We then use $\mask^{\step}_\text{agg}$, rather than $\mask^{\step}$, as inlier/outlier mask to train our 3DGS model in~\cref{eq:nvs}.
We designate this model configuration as `\methodsdcluster'.

\subsubsection{Spatio-temporal clustering}
\label{sec:learnt}

A second approach is to train a \textit{classifier} that determines 
whether or not pixels should be included in the optimization~\cref{eq:nvs}, based on their associated features.
To this end we use an MLP with  parameters~$\mlpparam$ that predicts pixel-wise inlier probabilities from pixel features:
\begin{equation}
\mask^{\step}_\text{mlp} = \mlp(\features; \mlpparam^{\step}).
\label{eq:learnt}
\end{equation}
As the $\mlpparam^{\step}$ notation implies, the classifier parameters are updated \textit{concurrently} with 3DGS optimization.
$\mlp$ is implemented with $1 {\times} 1$ convolutions, and hence acts in an i.i.d.\  fashion across pixels. 
We interleave the optimization of the MLP and the 3DGS model, such that the parameters of one are fixed while the other's are optimized, in a manner similar to alternating optimization.

\begin{figure}[t]
\centering
\setlength\tabcolsep{1.8pt}
\includegraphics[width=0.85\linewidth]{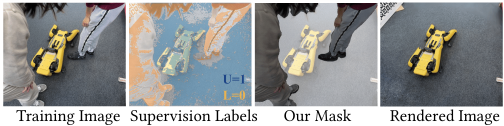}
\captionof{figure}{
Lower and upper error residual labels provide a weak supervision for training an MLP classifier for detecting outlier distractors.
}
\label{fig:mlp}
\end{figure}

The MLP classifier loss is given by
\begin{equation}
\loss(\mlpparam^{\step}) =  \loss{sup}(\mlpparam^{\step}) + \lambda \loss{reg}(\mlpparam^{\step}),
\label{eq:mlploss}
\end{equation}
with $\lambda{=}0.5$, and where $\loss{sup}$ supervises the classifier:
\begin{align}
   \loss{sup}(\mlpparam^{\step}) &= \maxx (\upperm^{\step} - \mlp(\features; \mlpparam^{\step}), 0)  \\ 
   &+ \maxx ( \mlp(\features; \mlpparam^{\step}) - \lowerm^{\step}, 0) \nonumber
\end{align}
and $\upperm$ and $\lowerm$ are self-supervision labels computed from the mask of the current residuals:
\begin{align}
\upperm^{\step} &= \mask^{\step} ~\text{from \cref{eq:robustnerf} with}~ \tau=.5
\\
\lowerm^{\step} &= \mask^{\step} ~\text{from \cref{eq:robustnerf} with}~ \tau=.9 
\end{align}
In other words, we directly supervise the classifier only on pixels for which we can confidently determine the inlier status based on reconstruction residuals, and otherwise we heavily rely on semantic similarity in the feature space; see \Cref{fig:mlp}.
To further regularize $\mlp$ to map similar features to similar probabilities, we minimize its Lipschitz constant via $\loss{reg}$ as detailed in~\cite[Eq.~(13)]{Liu2022}.
We then use $\mask^{\step}_\text{mlp}$, instead of $\mask^{\step}$, as inlier/outlier mask to train 3DGS in~\cref{eq:nvs}. 
We designate this configuration as `\methodsdmlp'. 
As we are \textit{co-training} our classifier together with the 3DGS model, additional care is needed in its implementation; see~\Cref{sec:sched}.

\subsection{Adapting 3DGS to robust optimization}
\label{sec:adaptations}
Directly applying any robust masking techniques to 3DGS can result in the robust mask overfitting to a premature 3DGS model (\cref{sec:sched}), with inlier estimator becoming skewed by image-based training~(\cref{sec:histogram}), or the densification tactics (\cref{sec:pruning}) of 3DGS. 
We propose solutions to these issues in what follows.

\subsubsection{Warm up with scheduled sampling}
\label{sec:sched}
We find it important to apply masks gradually, because the initial residuals are random.
This is doubly true if we use the learned clustering for masking since the MLP will not have converged early in the optimization, and predicts random masks.
Further, direct use of the outlier mask tends to  quickly overcommit to outliers, preventing valuable error back-propagation and learning from those regions.
We mitigate this by formulating our masking policy for each pixel as sampling from a Bernoulli distribution based on the masks:
\begin{equation}
    \mask^{\step} \sim \mathcal{B}\left( \sched \cdot 1 + (1-\sched) \cdot \mask^{\step}_{*}\right)\, ; 
\label{eq:schedule}
\end{equation}
where $\sched$ is a staircase exponential scheduler (detailed in~\ref{supp:scheduler}), going from one to zero, providing a warm-up.
This allows us to still sparsely sample gradients in areas we are not confident about, leading to better classification of outliers.

\paragraph{Warm-up scheduler}
\label{supp:scheduler}
As explained in~\cref{sec:adaptations}, we use a staircase exponential scheduler for our warm up phase:
\begin{equation}
    \sched = \exp \left(-\beta_1 \left\lfloor \frac{(t + 1)}{\beta_2} \right\rfloor \right),
\end{equation}
where $t$ is the time step in optimization, $\beta_1$ controls the speed of decay and $\beta_2$ determines the length of steps in the staircase function. We use $\beta_1 = 3\times10^{-4}$ and $\beta_2 = 1.5$ for all experiments, but the three highest occlusion rate scenes in NeRF On-the-go~\citep{Ren2024nerfonthego} dataset where we use $\beta_1 = 3\times10^{-3}$ for a fastest decay.

\subsubsection{Trimmed estimators in image-based training}
\label{sec:histogram}
As~\cite{Sabour2023robustnerf} implements a \textit{trimmed} estimator, the underlying assumption is that each minibatch (on average) contains the same proportion of outliers.
This assumption is broken in a 3DGS training run, where each minibatch is a \textit{whole} image, rather than a random set of  pixels drawn from the set of training images.
This creates a challenge in the implementation of the generalized median of~\cref{eq:robustnerf}, as the distribution of outliers is skewed between images.

We overcome this by tracking residual magnitudes over multiple training batches.
In particular, we discretize residual magnitudes into $B$ histogram buckets of width equal to the lower bound of rendering error ($10^{-3}$).
We update the likelihood of each bucket at each iteration via a discounted update to the bucket population, similar to fast median filtering approaches~\citep{medianfilter}.
This maintains a moving estimate of residual distribution, with constant memory consumption, from which we can extract our generalized median value~$\rho$ as the $\tau$ quantile in the histogram population; we refer the reader to our source code for implementation details.

\begin{figure*}[ht]
\includegraphics[width=\linewidth]{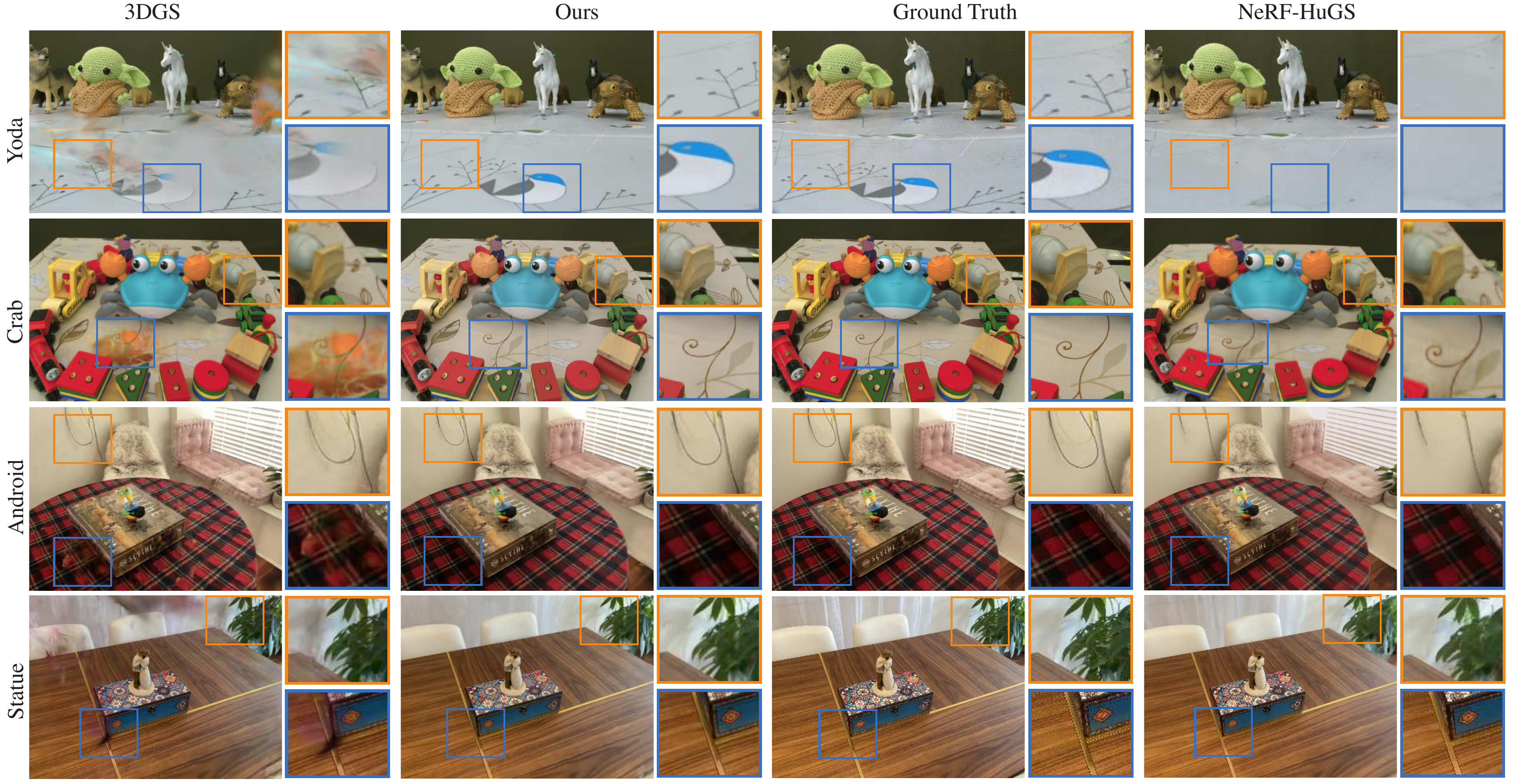}

\vspace*{0.2cm}
\setlength\tabcolsep{7.5pt}
\resizebox{\linewidth}{!}{
\begin{tabular}{l|ccc|ccc|ccc|ccc}
& \multicolumn{3}{c|}{Statue} & \multicolumn{3}{c|}{Android} & \multicolumn{3}{c|}{Yoda} & \multicolumn{3}{c}{Crab (1)} \\
 & \psnr & \ssim & \lpips & \psnr & \ssim & \lpips & \psnr & \ssim & \lpips & \psnr & \ssim & \lpips \\ \hline
MipNerf360 & 19.86 & .69 & .23 & 21.81 & .69 & .18 & 23.75 & .77 & .22 & 29.25 & .92 & .09  \\
RobustNerf & 20.60 & .76 & .15 & 23.28 & .75 & .13 & 29.78 & .82 & .15 & 32.22 & .94 & .06  \\
NeRF On-the-go & 21.58 & .77 & .24 & 23.50 & .75 & .21 & 29.96 & .83 & .24 & - & - & - \\
NeRF-HuGS & 21.00 & .77 & .18$^\dagger$ & 23.32 & .76 & .20$^\dagger$ & 30.70 & .83 & .22$^\dagger$ & 34.16 & .96 & \textbf{.07$^\dagger$} \\ \hline
\tdgsshortname & 21.68 & .83 & .14 & 23.33 & .80 & .15 & 27.15 & .92 & .13 & 31.80 & .96 & .08\\
{Our \methodsdmlp} & \textbf{22.69} & \textbf{.85} & \textbf{.12} & \textbf{25.15} & \textbf{.86} & \textbf{.09} & \textbf{33.60} & \textbf{.96} & \textbf{.10} & \textbf{35.85} & \textbf{.97} & .08 \\ \hline \hline
{\tdgsshortname} on \clean  & 28.02 & .95 & .05 & 25.42 & .87 & .07 & 33.69 & .94 & .12 & - & - & -\\
{\tdgsshortname}* on \clean & 28.63 & .95 & .04 & 25.38 & .87 & .07 & 36.34 & .97 & .07 & - & - & -
\end{tabular}
} %
\\[.75em]

\captionof{figure}{
Quantitative and qualitative evaluation on RobustNeRF~\citep{Sabour2023robustnerf} datasets show that SLS outperforms baseline methods on 3DGS and NeRF, by preventing over- or under-masking.
$\dagger$ denotes VGG LPIPS computed on NeRF-HuGS results rather than AlexNet LPIPS reported in NeRF-HuGS. 
{\tdgsshortname}* denotes \tdgsshortname with utility-based pruning.
}
\label{fig:robustnerf}
\end{figure*}

\subsubsection{A friendly alternative to ``opacity reset''}
\label{sec:pruning}
\cite{Kerbl2023tdgs} proposed to reset the opacity of all Gaussians every $M$ iterations. This opacity reset is a mechanism that deals with two main problems. 
First, in challenging datasets the optimization has the tendency to accumulate Gaussians close to the cameras. 
These are often referred to as \textit{floaters} in the literature.
Floaters are hard to deal with because they force camera rays to saturate their transmittance early and as a result gradients do not have a chance to flow through the occluded parts of the scene.
Opacity reset lowers the opacity of all Gaussians such that gradients can flow again along the whole ray. 
Second, opacity reset acts as a control mechanism for the number of Gaussians. 
Resetting opacity to a low value allows for Gaussians that never recover a higher opacity to be pruned by the adaptive density control mechanism~\citep{Kerbl2023tdgs}. 

However, opacity reset interferes with residual distribution tracking~(\cref{sec:histogram}), causing residuals to become artificially large in the iterations following opacity reset. 
Simply disabling does not work due to it's necessity to the optimization. 
Following~\cite{Goli2024bayesrays}, we instead propose utilization-based pruning (\methodprune).
We track the gradient of the rendered colors with respect to the projected splat positions\footnote{Please carefully note that this is the gradient of the rendered image with respect to Gaussian positions, and not the gradient of the loss.} $\x_g$ of each Gaussian $g$.
Computing the derivative with respect to projected positions, as opposed to 3D positions, allows for a less memory-intensive GPU implementation, while providing a similar metric as in~\cite{Goli2024bayesrays}. 
More concretely, we define the utilization as:
\begin{equation}
\util_{\element} =
\sum_{t \in \mathcal{N}_\steps(t)} 
\mathbb{E}_{w,h} \:
\left\|\mask^{(t)}_{h,w} \cdot \tfrac{ \partial \hat\image^{(t)}_{h,w}}{\partial\x^{(t)}_{\element}}\right\|_2^2
\label{eq:bayessplat}
\end{equation}
We average this metric across the image ($W{\times}H$), computing it every $\steps{=}100$ steps accumulated across the previous set of~$|\mathcal{N}_\steps(t)|{=}100$ images.
We prune Gaussians whenever $\util_{\element} {<} \pthresh$, with $\pthresh=10^{-8}$. 
Replacing opacity reset with utilization-based pruning achieves both original goals of opacity reset while alleviating interference to our residual distribution tracking. 
Utilization-based pruning significantly compresses scene representation by using fewer primitives while achieving comparable reconstruction quality even in outlier-free scenes; see \Cref{sec:exp_prune}. 
It also effectively deals with floaters; see \Cref{fig:ablation2}. 
Floaters, naturally, have low utilization as they participate in the rendering of very few views.
Furthermore, using masked derivatives as in~\cref{eq:bayessplat} allows for the removal of any splat that has leaked through the robust mask in the warm-up stage.

\subsubsection{Appearance modeling}
\label{sec:appearance}
While~\cite{Kerbl2023tdgs} assumed that the images of a scene~(up to distractors) are perfectly photometrically consistent, this is rarely the case for casual captures typically employing automatic exposure and white-balance.
We address this by incorporating the solution from~\cite{Rematas2022urf} adapted to the view-dependent colors represented as spherical harmonics from~\cite{Kerbl2023tdgs}.
We co-optimize a latent~$\aglolatent_{n} {\in} \mathbb{R}^{64}$ per input camera view, and map this latent vector via an MLP to a linear transformation acting on the harmonics coefficients~$\sphcoef$:
\begin{equation}
\hat{\sphcoef}_{i} = \mathbf{a} \odot \sphcoef_{i} + \mathbf{b},
\quad
\mathbf{a}, \mathbf{b} = \aglonet(\latent_n; \mlpparam_\aglonet)
\label{eq:glo}
\end{equation} 
where $\odot$ is the Hadamard product, $\mathbf{b}$ models changes in brightness, and $\mathbf{a}$ provides the expressive power for white-balance. 
During optimization, the trainable parameters also include $\mlpparam_\aglonet$ and $\{\latent_n\}$.
Such a reduced model can prevent $\latent_n$ from explaining distractors as per-image adjustments, as would happen in a simpler GLO~\citep{MartinBrualla21cvpr_nerfw}; see~\cite{Rematas2022urf} for an analysis.

\section{Results}
\label{sec:experiments}

In what follows, we compare our proposed method on established datasets of casual distractor-filled captures~(\cref{sec:exp_reconstruct}), comparing with other methods.
We then investigate the effect of our proposed opacity reset alternative pruning~(\cref{sec:exp_prune}). Finally, we report a complete analysis of different variants of our clustering, along with an ablation study of our design choices ~(\cref{sec:exp_ablation}).

\paragraph{Datasets}
We evaluate our method on the RobustNeRF~\citep{Sabour2023robustnerf} and NeRF on-the-go~\citep{Ren2024nerfonthego} datasets of \textit{casual captures}. 
The RobustNeRF dataset includes four scenes with distractor-filled and distractor-free training splits, allowing us to compare a robust model with a `clean' model trained on distractor-free images.
All models are evaluated on a clean test set.
The `Crab' and `Yoda' scenes feature variable distractors across images, not captured in a single casual video, but these exact robotic capture with twin distractor-free and distractor-filled images allow a fair comparison to the `clean' model.
Note the (originally released) Crab~(1) scene had a test set with same set of views as those in the train set, which is fixed in Crab~(2).
We compare previous methods on Crab~(1), and present full results on Crab~(2) in~\Cref{sec:exp_ablation}, and in the supplementary material~\ref{supp:crab}.
The NeRF on-the-go dataset has six scenes with three levels of transient distractor occlusion~(low, medium, high) and a separate clean test set for quantitative comparison.

\begin{figure}[t]
\centering
\setlength\tabcolsep{1.8pt}
\includegraphics[width=0.75\linewidth]{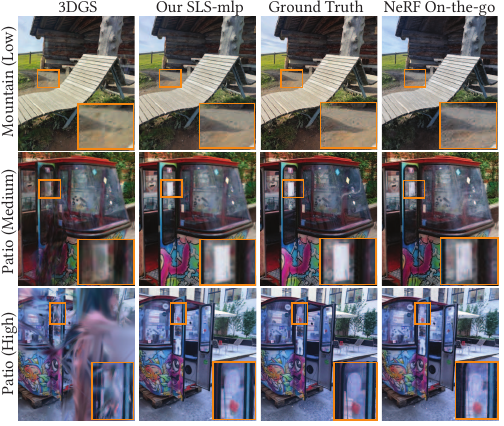}

\begin{tabular}{l|ccc|ccc|ccc}
 & \multicolumn{3}{c|}{Mountain} & \multicolumn{3}{c|}{Patio} & \multicolumn{3}{c}{Patio High}\\
 & \psnr & \ssim & \lpips & \psnr & \ssim & \lpips & \psnr & \ssim & \lpips  \\ \hline
MipNerf360 & 19.64 & 0.60 & 0.35 & 15.48 & 0.50 & 0.42 & 15.73 & 0.43 & 0.49 \\
RobustNerf & 18.07 & 0.49 & 0.49 & 17.55 & 0.53 & 0.45 & 12.99 & 0.35 & 0.61 \\
Nerf On-the-go & 20.15 & 0.64 & 0.26 & 20.78 & 0.75 & 0.22 & 21.41 & 0.72 & 0.24 \\ \hline
\tdgsshortname & 20.18 & 0.70 & 0.23 & 18.25 & 0.71 & 0.23 & 18.14 & 0.68 & 0.30 \\
Our \methodsdmlp & \textbf{22.53} & \textbf{0.77} & \textbf{0.18} & \textbf{22.24} & \textbf{0.86} & \textbf{0.10} & \textbf{22.84} & \textbf{0.83} & \textbf{0.16}
\end{tabular}
\captionof{figure}{
SLS reconstructs scenes from NeRF On-the-go~\citep{Ren2024nerfonthego} dataset in great detail. High-occlusion lingering distractors, lead to distractor leaks modeled as noisy floaters in baselines. Our method is free of such artifacts.
}
\label{fig:nerfonthego2}
\end{figure}

\paragraph{Baselines} 
Distractor-free reconstruction has yet to be widely addressed by 3D Gaussian Splatting methods.
Existing methods mostly focus on global appearance changes such as brightness variation~\citep{Dahmani2024swag, Wang2024wegs, Kulhanek2024wildgaussians}, and do not focus on the distractor-filled datasets of casual captures curated for this task.  
We therefore compare against vanilla 3DGS and robust NeRF methods. We further add GLO to the vanilla 3DGS baseline to be comparable with MipNeRF360 results that have GLO enabled.
We compare against state-of-the-art NeRF methods, NeRF on-the-go~\citep{Ren2024nerfonthego}, NeRF-HuGS~\citep{chen2024nerfhugs} and RobustNeRF~\citep{Sabour2023robustnerf}.
We also include MipNeRF-360~\citep{Barron2022mipnerf360} as a baseline for NeRF.

\paragraph{Metrics}
We compute the commonly used image reconstruction metrics of PSNR, SSIM and LPIPS.
We use normalized VGG features, as most do, when computing LPIPS metrics.
NeRF-HuGS~\citep{chen2024nerfhugs} reports LPIPS metrics from AlexNet features; for fair comparison, we compute and report VGG LPIPS metrics on their released renderings.
Finally, note NeRF on-the-go does not evaluate on `Crab', because of the aforementioned issue.

\begin{figure}[t]
 \centering
\includegraphics[width=0.85\linewidth]{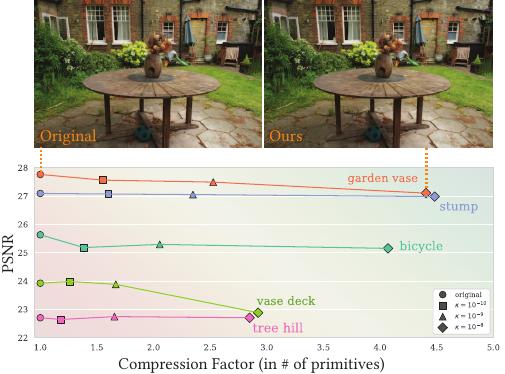}
\captionof{figure}{
Quantitative and qualitative results on MipNeRF360~\citep{Barron2022mipnerf360} dataset shows gradient-based pruning can reduce the number of Gaussians up to $4.5\times$ with only marginal degradation of image quality. 
}
\label{fig:pruning}
\end{figure}

\begin{figure*}[ht]
\centering
\setlength\tabcolsep{1.8pt}
\includegraphics[width=\linewidth]{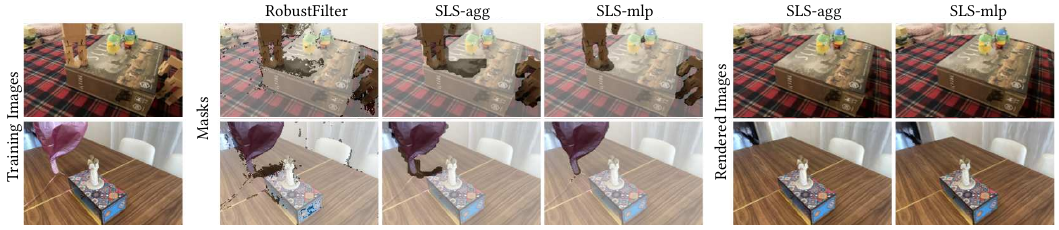}

\resizebox{\linewidth}{!}{
\begin{tabular}{l|cccc|cccccc|c}
 & Android & Statue & Crab (2) & Yoda & Mountain & Fountain & Corner & Patio & Spot & Patio-High & Average \\ \hline
\tdgsshortname  & $23.33 \pm 0.13$ & $21.68 \pm 0.16$ & $29.74 \pm 0.37$ & $27.15 \pm 0.61$ & $20.90 \pm 0.18$ & $21.85 \pm 0.27$ & $23.39 \pm 0.43$ & $18.33 \pm 0.27$ & $21.50 \pm 0.85$ & $18.06 \pm 0.71$ & $22.60$ \\
\methodfilter  & $24.50 \pm 0.05$ & $22.70 \pm 0.06$ & $31.34 \pm 0.13$ & $33.23 \pm 0.13$ & $22.29 \pm 0.07$ & $22.59 \pm 0.07$ & $25.20 \pm 0.10$ & $18.16 \pm 0.19$ & $25.54 \pm 0.08$ & $23.01 \pm 0.18$ & $24.85$ \\
\methodsdcluster & $24.94 \pm 0.08$ & $\mathbf{23.16 \pm 0.08}$ & $33.50 \pm 0.14$ & $35.01 \pm 0.21$ & $22.65 \pm 0.14$ & $23.03 \pm 0.17$ & $26.33 \pm 0.10$ & $22.31 \pm 0.13$ & $\mathbf{26.34 \pm 0.37}$ & $\mathbf{23.54 \pm 0.15}$ & $26.08$ \\
\methodsdmlp w/o UBP & $25.08 \pm 0.04$ & $22.75 \pm 0.14$ & $\mathbf{34.43 \pm 0.03}$ & $\mathbf{34.36 \pm 0.24}$ & $\mathbf{22.93 \pm 0.09}$ & $\mathbf{23.19 \pm 0.13}$ & $\mathbf{26.74 \pm 0.13}$ & $\mathbf{22.36 \pm 0.07}$ & $25.95 \pm 0.47$ & $23.27 \pm 0.13$ & $\mathbf{26.11}$ \\
\hline
\methodsdmlp w/ \methodprune& $\mathbf{25.15 \pm 0.05}$ & $22.69 \pm 0.16$ & $33.63 \pm 0.27$ & $33.60 \pm 0.30$ & $22.53 \pm 0.11$ & $22.81 \pm 0.10$ & $26.43 \pm 0.08$ & $22.24 \pm 0.19$ & $25.76 \pm 0.15$ & $22.84 \pm 0.32$ & $25.77$

\end{tabular}
}
\captionof{figure}{
We ablate our different robust masking methods on~\cite{Sabour2023robustnerf} and~\cite{Ren2024nerfonthego} datasets. The reconstruction metrics and qualitative masks illustrate the performance of \methodsdcluster~\cref{eq:agglomerative} and \methodsdmlp~\cref{eq:learnt} over a basic \methodfilter~\cref{eq:robustnerf} adapted from~\cite{Sabour2023robustnerf}, and baseline vanilla 3DGS~\citep{Kerbl2023tdgs}.
The final row enables Utility-Based Pruning (\methodprune) (\cref{sec:pruning}).
All methods use opacity reset disabled, the same scheduling in \cref{eq:schedule}, and GLO \cref{eq:glo} enabled on all runs including \tdgsshortname.
\methodsdcluster and \methodsdmlp are mostly within $2\sigma$ of each other on all tasks.
The $\sigma$ is calculated from $5$ independent runs each. 
}
\label{fig:robustprogress}
\end{figure*}

\paragraph{Implementation details} 
We train our 3DGS models with the same hyper-parameters as in the officially released codebase.
All models are trained for 30k iterations. 
We turn off the opacity-reset and only reset the non-diffuse spherical harmonic coefficients to 0.001 at the 8000th step.
This ensures that any distractors leaked in the earlier stages of MLP training do not get modeled as view dependent effects.
We run \methodprune every 100 steps, from the 500th to 15000th step.
For MLP training, we use the Adam optimizer with a 0.001 learning rate. 
We compute image features from the 2nd upsampling layer of Stable diffusion v2.1, denoising time step of 261, and an empty prompt.
\cite{Tang2023emergent} found this configuration most efficient for segmentation and keypoint correspondence tasks.
We concatenate positional encoding of degree 20 to the features as input to the MLP. 

\subsection{Distractor-free 3D reconstruction}
\label{sec:exp_reconstruct}

We evaluate our method by preforming 3D reconstruction on the RobustNeRF and NeRF on-the-go datasets. 
In \Cref{fig:robustnerf}, we quantitatively show that \methodsdmlp outperforms all the robust NeRF-based baselines on the RobustNeRF dataset.
The results further show that we improve significantly upon vanilla 3DGS, while having closer performance to the ideal clean models, specifically on `Yoda' and `Android'. We further qualitatively compare with vanilla 3DGS and NeRF-HuGS.
The qualitative results show that vanilla 3DGS tries to model distractors as noisy floater splats (`Yoda', `Statue') or view-dependent effects (`Android') or a mixture of both (`Crab'). 
NeRF-HuGS~\citep{chen2024nerfhugs} which uses segmentation-based masks shows signs of over-masking (removing static parts in all four scenes), or under-mask in challenging sparsely sampled views letting in transient objects (`Crab'). 

In \Cref{fig:nerfonthego1} and \Cref{fig:nerfonthego2}, we perform a similar analysis on the NeRF On-the-go~\citep{Ren2024nerfonthego} dataset. 
While we show superior quantitative results to both SOTA robust NeRF methods, we also achieve a significant performance boost compared to vanilla 3DGS. The results further show that for low occlusion scenes the robust initialization of vanilla 3DGS from COLMAP~\citep{Schonberger2016sfmrevisited} point clouds, specifically RANSAC's rejection of outliers, is enough to yield good reconstruction quality.
However, as the distractor density increases, 3DGS reconstruction quality drops with qualitative results showing leakage of distractor transients.
Additionally, qualitative results show that NeRF On-the-go fails to remove some of the distractors included in the early stages of training (`Patio', `Corner', `mountain' and `Spot'), showing further signs of overfitting to the rendering error.
This also is seen in the over-masking of fine details (`Patio High') or even bigger structures (`Fountain') removed completely.

\paragraph{Additional results on the Crab dataset}
\label{supp:crab}
The Crab dataset in RobustNeRF~\citep{Sabour2023robustnerf} has two released versions, one without any additional viewpoints for testing and one with an extra test set of camera viewpoints. In the main chapter we refer to the first version as Crab (1) and the second version as Crab (2). While previous work has only tested on Crab (1), Crab (2) has the conventional format of NeRF datasets with separate test views. We present additional results on Crab (2) scene in~\cref{tab:crab2}, showing that SLS-MLP has a very close performance to the ideal 3DGS model trained on clean data, when tested from different viewpoints than training datset.
\begin{table}[ht]
\centering
\begin{tabular}{c|c|c||c|c}
      & SLS-MLP & 3DGS  & 3DGS Clean & 3DGS* Clean \\ \hline
PSNR $\uparrow$ & \textbf{34.35}   & 26.33 & 33.43         & \textbf{35.58}          \\ \hline
SSIM $\uparrow$ & \textbf{0.96}    & 0.91  & 0.94          & \textbf{0.97}           \\ \hline
LPIPS $\downarrow$ & \textbf{0.03}    & 0.08  & 0.05          & \textbf{0.01}          
\end{tabular}
\captionof{table}{Quantitative result on Crab (2) dataset, where a test set with different viewpoints than training is provided, shows superior performance of SLS-MLP to vanilla 3DGS and close performance to the ideal model trained on clean data. 3DGS* denotes use of utility-based pruning.}
\label{tab:crab2}
\end{table}

\begin{figure*}
    \centering
    \includegraphics[width=\linewidth]{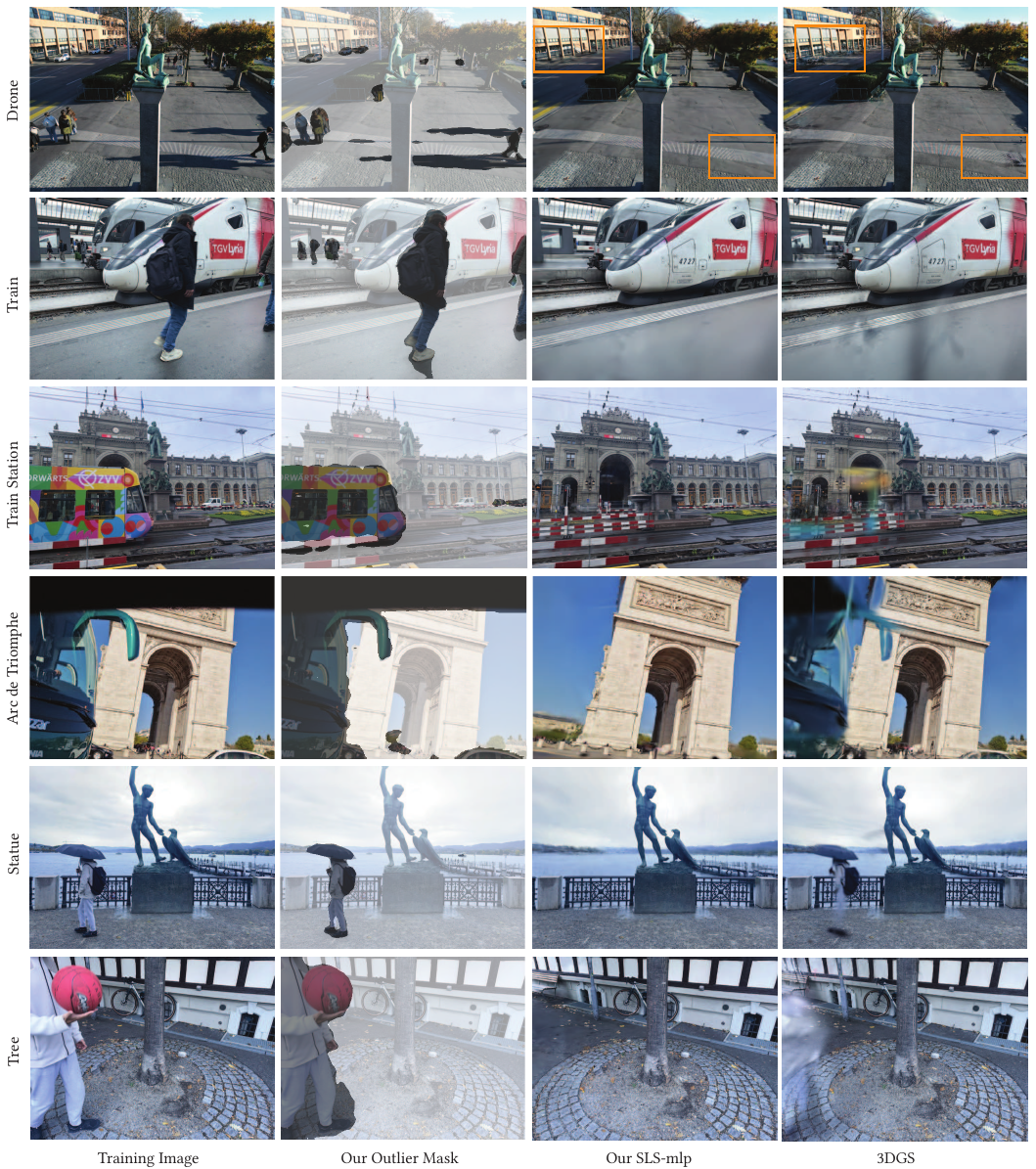}
    \captionof{figure}{Qualitative results on scenes from NeRF On-the-go~\citep{Ren2024nerfonthego} show robustness of our method to transparent surfaces such as glass windshields (Train Station, Arc de Triomphe), and similarly-colored distractors and backgrounds (Tree, Statue). Further, our method shows robustness to distractor shadows to a degree (Drone, Train).}
    \label{fig:supp}
\end{figure*}

\paragraph{Additional Results on NeRF On-the-go dataset}
NeRF On-the-go dataset~\citep{Ren2024nerfonthego} provides six additional scenes for qualitative evaluation only. We provide qualitative results on these scenes in~\cref{fig:supp}. In the `Drone' scene, our method is able to detect harder shadows of people and remove them seamlessly. However, complete robustness to softer shadows is a limitation of our work, as the semantic class of shadows is not reflected very well in the text-to-image features that we use. This can be seen in the `Train' scene, where shadows of people are only detected to a degree. Further, in the`Train Station' and `Arc de Triomphe' scenes we see that our model shows robustness to transparent surfaces on distractors such as glass windshields. Finally, in the `Statue' and `Tree' scenes, SLS-MLP works well in distinguishing between the distractor and the background, even though the distractors (mostly) have very similar color to their background.

\subsection{Effect of utilization-based pruning}
\label{sec:exp_prune}
\begin{figure}[t]
\centering
\setlength\tabcolsep{1.8pt}
\includegraphics[width=.75\linewidth]{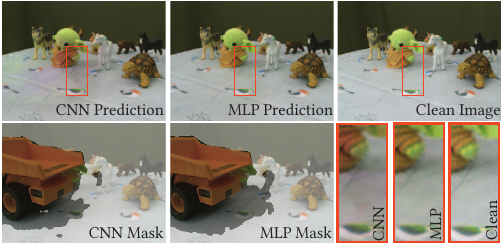}\\
\begin{tabular}{c|c|c|c|c||c|c|c}
                   & \begin{tabular}[c]{@{}c@{}}SLS- \\mlp\end{tabular} & \begin{tabular}[c]{@{}c@{}}CNN \\\end{tabular}    & \begin{tabular}[c]{@{}c@{}}SLS-mlp \\ ($\lambda=0.1)$\end{tabular} & \begin{tabular}[c]{@{}c@{}}SLS-mlp \\ ($\lambda=1)$\end{tabular} & \begin{tabular}[c]{@{}c@{}} SLS- \\ agg\end{tabular}   & \begin{tabular}[c]{@{}c@{}}SLS-agg \\ ($\numclusters=10$)\end{tabular} & \begin{tabular}[c]{@{}c@{}}SLS-agg \\ ($\numclusters=1000$)\end{tabular} \\ \hline
PSNR $\uparrow$    & 28.72 & 27.91 & 28.71                                                        & 28.61                                                      & 28.91 & 27.71                                                 & 29.00                                                   \\ \hline
SSIM $\uparrow$    & 0.90  & 0.89  & 0.90                                                         & 0.89                                                       & 0.90  & 0.89                                                  & 0.90                                                    \\ \hline
LPIPS $\downarrow$ & 0.10  & 0.13  & 0.11                                                         & 0.12                                                       & 0.10  & 0.11                                                  & 0.09                                                   
\end{tabular}
\captionof{figure}{
Ablations on variants from \Cref{sec:semantic} show
replacing the MLP \cref{eq:learnt} in \methodsdmlp with a CNN reduces quality. 
Varying its regularization coefficient $\lambda$ in \cref{eq:mlploss} shows minimal impact. 
More agglomerative clusters in \methodsdcluster \cref{eq:agglomerativeprob} improve performance, plateauing for $\numclusters{\geq}100$. Metrics averaged over all RobustNeRF dataset.
}
\label{fig:ablation1}
\end{figure}

\begin{figure}[t]
\centering
\setlength\tabcolsep{5.8pt}
\includegraphics[width=0.85\linewidth]{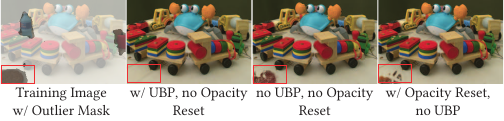}\\
\begin{tabular}{c|c|c|c|c|c|c}
                   & SLS-mlp   & No \methodprune & Opacity Reset & No GLO & No $\alpha$ & No $\mathcal{B}$ \\ \hline
PSNR $\uparrow$    & 28.72 & 29.27    & 27.64         & 28.43  & 28.41       & 28.41            \\ \hline
SSIM $\uparrow$    & 0.90  & 0.90     & 0.89          & 0.89   & 0.90        & 0.90             \\ \hline
LPIPS $\downarrow$ & 0.10  & 0.09     & 0.11          & 0.11   & 0.11        & 0.11            
\end{tabular}
\captionof{figure}{
Ablation on adaptations from \Cref{sec:adaptations} show
disabling \methodprune (\cref{sec:pruning}) may produce higher reconstruction metrics but leaks transients as seen in the lower-left corner of the image; replacing it with ``Opacity Reset'' as originally introduced in 3DGS is also ineffective.
GLO appearance modelling \cref{eq:glo} improves quality, as do scheduling~($\alpha$), and Bernoulli sampling ($\mathcal{B}$)~\cref{eq:schedule}. Experiments are executed on \methodsdmlp, with metrics averaged over all RobustNeRF dataset.
}
\label{fig:ablation2}
\end{figure}

In all our experiments, enabling our proposed utilization-based pruning (\methodprune) (\cref{sec:pruning}), decreases the number of Gaussians from $4\times$ to $6\times$.
This compression translates to at least a $2\times$ reduction in training time with \methodprune enabled and $3\times$ during inference.
\Cref{fig:ablation2} shows that enabling \methodprune may degrade quantitative measurements slightly, but in practice the final renderings are cleaner with less floaters (e.g. bottom left of the image).
Similar observations indicate that metrics such as PSNR and LPIPS may not completely reflect the presence of floaters as clearly as a rendered video.
Given the substantial reduction in number of Gaussians, we propose  \methodprune as a compression technique applicable to cluttered, \emph{as well as clean}, datasets. \Cref{fig:pruning} shows that on clean MipNeRF360 \citep{Barron21iccv_Mip_NeRF} datasets, using \methodprune instead of opacity reset reduces the number of Gaussians from $2\times$ to $4.5\times$ while preserving rendering quality.

\subsection{Ablation study}
\label{sec:exp_ablation}
In \Cref{fig:robustprogress}, we compare the performance of \methodshortname with a progression of other robust masking techniques.
The progression begins with a naive application of a robust filter~\eqref{eq:robustnerf}, followed by the application of \methodsdcluster, and finally the use of an MLP in \methodsdmlp.
We demonstrate that both \methodsdcluster and \methodsdmlp are capable of effectively removing distractors from the reconstructed scene, while maintaining maximal coverage of the scene.
Further, in~\Cref{fig:ablation1} and~\Cref{fig:ablation2} we ablate our choices in both architectural design, and the adaptations proposed in~\Cref{sec:adaptations}.
\cref{fig:ablation1} shows that using an MLP instead of a small CNN (both roughly having 30K parameters, and two non-linear activations) can adapt better to subtle transients like shadows.
The choice of regularizer weight $\lambda$ seems to have little effect. 
In agglomerative clustering, more clusters generally lead to better results, with gains diminishing after 100 clusters.
\Cref{fig:ablation2} further illustrates the effectiveness of \methodprune in removing leaked distractors.
Our other adaptations, GLO, warm-up stage and Bernoulli sampling all show improvements.

\subsection{Distinguishing between semantically similar instances}
\label{supp:similarity}
\begin{figure}[t]
    \centering
    \includegraphics[width=.7\linewidth]{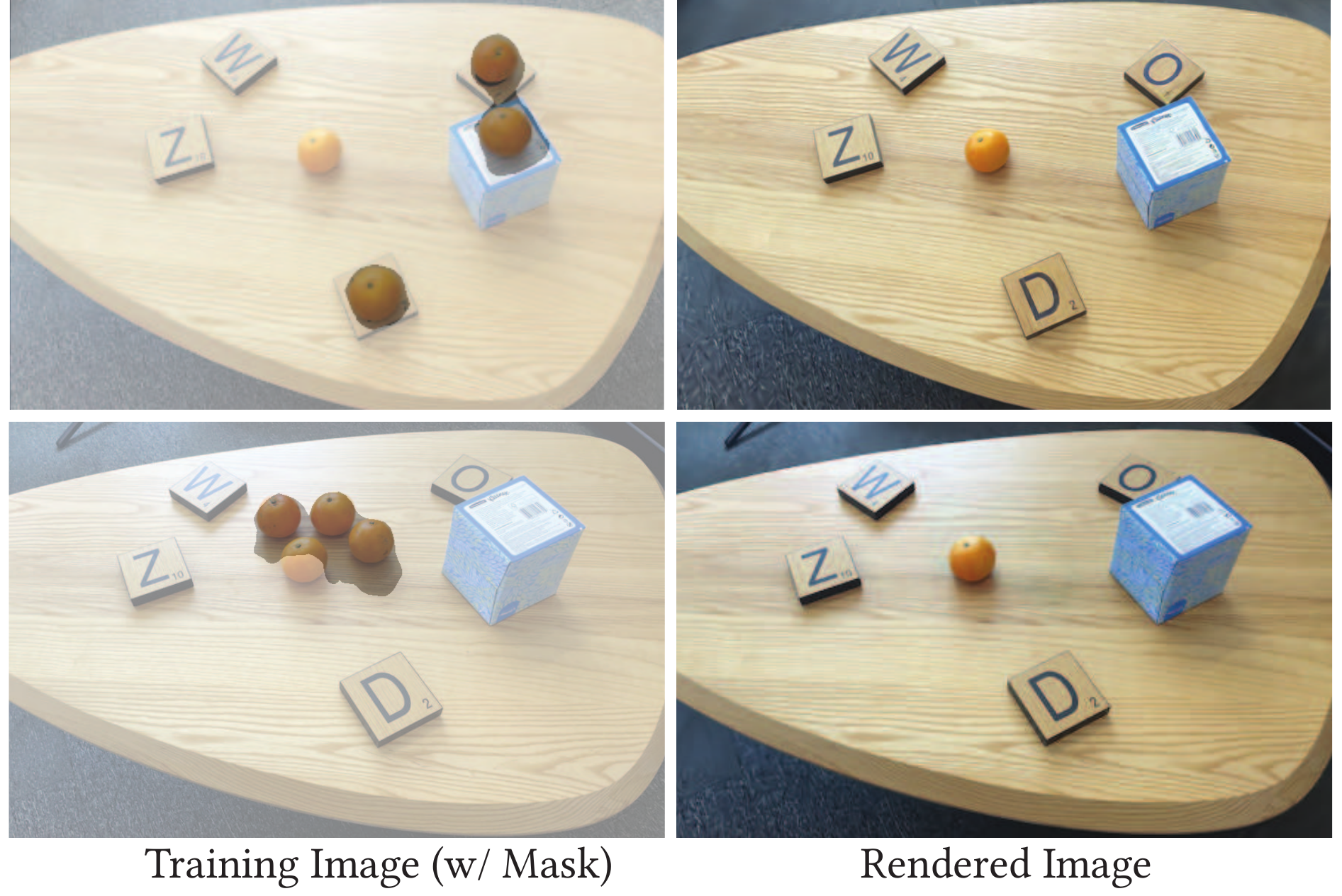}
    \caption{SLS-MLP can correctly distinguish between similar-looking oranges when the non-dsitractor instances are far from the distractor instance (centered on the table). However, the masking task becomes more difficult for nearby instances of distractor and non-distractors.}
    \label{fig:oranges}
\end{figure}
Our method relies on features extracted from text-to-image diffusion models, to reliably learn the subspace of features that represent distractors appearing in a casual capture. However, this implies that in cases where similar instances of a semantic class appear as both distractors and non-distractors, our model would not be able to distinguish the two. We show in~\cref{fig:oranges} that this is not generally true. When the instances of the same class are not very close in image space, the model can distinguish between distractor and non-distractor orange instances, however in scenarios where they are very close we can see over-masking of the non-distractor orange. We hypothesize that this is a result of foundation features encoding not only semantics, but also appearance and position in the image as shown in~\cite{El_Banani_2024_CVPR}.

\section{Conclusion}
We have presented \methodname, a method for transient distractor suppression for 3DGS.
We established a class of masking strategies that exploit semantic features to effectively identify transient distractors without any explicit supervision. 
Specifically, we proposed a spatial clustering method `\methodsdcluster' that is fast and does not require further training, simply assigning an inlier-outlier classification to each cluster.
We then proposed a spatio-temporal learned clustering based on training a light-weight MLP simultaneously with the 3DGS model, `\methodsdmlp', that allows for higher precision grouping of semantically associated pixels, while marginally slower than clustering.
Our methods leverage the semantic bias of Stable Diffusion features and robust techniques to achieve state of the art suppression of transient distractors.
We also introduced a gradient-based pruning method that offers same reconstruction quality as vanilla 3DGS, while using significantly lower number of splats, and is compatible with our distractor suppression methods.
We believe that our work is an important contribution necessary for widespread adoption of 3DGS to real-world in-the-wild applications.

\paragraph{Limitations} 
Our reliance on text-to-image features, although generally beneficial for robust detection of distractors, imposes some limitations.
One limitation is that when distractor and non-distractors of the same semantic class are present and in close proximity, they may not be distinguished by our model. Details are discussed further in the supplementary materials~\ref{supp:similarity}.
Further, the low-resolution features these models provide can miss thin structures such as the balloon string of \Cref{fig:robustprogress}.
Especially in the use of clustering, upsampling the features to image resolution results in imprecise edges. 
Our pruning strategy, is based on epistemic uncertainty computation per primitive which is effective in removing lesser utilized Gaussians. However if the uncertainty is thresholded too aggressively (e.g. `vase deck' in~\cref{fig:pruning}), it can remove parts of the scene that are rarely viewed in the training data.

\chapter{Discussion and Conclusions}   
In this thesis we discussed two tasks with regard to unsupervised object learning. When faced with the lack of supervision in computer vision, typically the only training signal that is left to the model is the reconstruction loss of the images. We can improve the quality of the results by augmenting the training signal with inputs from off-the-shelf models that are trained for another supervised task. We argue that relying on the supervised cross task signals is an efficient method for incorporating the available labelled datasets. An unsupervised method has advantages such as access to larger datasets, being less prone to human biases in labelling, and generalizing better to unforeseen data points.

Another challenge in unsupervised image understanding is defining the objects of interest and their semantic granularity. In essence, the two subtasks are defining foreground vs background and deciding on the granularity of objects and their parts that should be captured as separate entities. 

We can use motion and move-ability as the defining factor for the foreground objects. We test this hypothesis with our work on FlowCapsules. FlowCapsules shows that a network with explicit object representations grouped into capsules (slots) can learn to detect and segment the parts in a dataset that move rigidly. We use a layering hypothesis to explain occluding parts. The layering method combined with per object representation with explicit pose parameters leads to the model learning an inherent (latent) representation of each moving object in the dataset. Therefore, the model is able to reconstruct the occluded parts of the objects as well as detecting negative spaces that form the same object shape that has been seen in the dataset. 

The main caveat in FlowCapsules is restricting the movement into rigid transformations. This limitation makes the method infeasible for deploying in the wild. Instances such as highly deformable object movements, i.e. a waterfall or a balloon exploding, invalidates the part-rigidity hypothesis. Thus the model will struggle to explain the optical flow for fully non-rigid transformations. 

Another restriction is explicitly representing each object in the dataset with fixed dimensional embeddings. Disjointing the embeddings makes the model inefficient in terms of representation capacity. Therefore, not applicable to large and varied datasets. Further research in this line of methods with explicit object representations, mainly called slots, have also struggled with being applied to general large datasets and competing with supervised methods. Hence, one might consider other approaches to make the learnable embeddings more information efficient. One such approach could be splitting the objects into more granular parts that are shared between more objects such as the paradigm of words and sub word tokens in NLP research. The main current limiting factor in unsupervised object oriented learning with explicit shape and pose representation is how to scale them to real world applications.

There are several aspects that hinder the applicability of unsupervised object oriented learning to wide scale applications. In an unsupervised setup reconstruction error of the images are one of the main training signals. Therefore, the capacity of the decoder in generating realistic images is/was a bottleneck in a fully unsupervised approach. Recent advances in Diffusion Models has improved the generative ability of neural networks significantly. Incorporating diffusion techniques is currently being explored in the object oriented learning community with promising results. Another aspect is defining the object of interest. Currently, most approaches which are successful in processing real images are skirting this challenge. Approaches such as using off-the-shelf annotators or prompt based large-scale training can help with defining what an object is for the model. But having an inherent definition of object is still an open problem with no obvious answers. Some in the community argue that the definition of an object relies on the task. But having task based groupings of pixels into objects is not universally applicable. Also, most humans given an image of unknown cluttered objects can give a count of the number of objects. Therefore, there exists a common sensible grouping into an object. One interesting direction to explore is breaking the task into a continual learning paradigm. The model has to explain the scene with a collection of objects. We start with empty scenes and single objects from 3D datasets such as ShapeNet and graduate into more complicated object centric datasets such as Objectron and ImageNet. In each step the model has to explain only the high error region of the scene. Therefore, learns the concept of object as something new and unseen. The other challenging aspect is the efficiency of discretized embedding based on object shape and position. A direction worth exploring in this regard is having explicit memory. On the other hand, recent advances in large scale language models indicate a possibility of having billions of parameters on extremely large datasets alleviates the need for explicit separate memory.

In the remainder of the thesis, we discussed a real world task where removing the transient objects in an unsupervised fashion is essential. Extrapolating 3D renderings from a set of images of a scene has gained traction in the recent years. Neural Randiance Fields (NeRF) was among the first that was able to render photo realistic 3D models using ray tracing techniques. Since this is an unsupervised task, the only loss available to the model is an image reconstruction loss as expected. Therefore, NeRF relies on photo-metric consistency of images to extract the underlying 3D scene. The presence of transient objects, such as pedestrians in a tourist site, breaks the photo-metric consistency. As a result the model is not able to converge to a clean and sharp rendering. In RobustNeRF we design a segmentation mask based on the reconstruction loss trajectory of the model. The goal is to mask the high error regions of the scene in an object oriented fashion. Since other factors such as high frequency textures also increase the error during training we incorporate basic objectness into our mask. While the method is applicable in many scenarios and the results are significantly better than SOTA at the time, tuning the parameters for each dataset to differentiate the objectness of the masks is borderline infeasible. 

The next chapter, SpotLessSplats, mitigates the finickiness of RobustNerf object filtering by introducing semantic features for the pixels. SpotLessSplats uses off-the-shelf stable diffusion features in order to group the pixels into semantic objects. The grouping of pixels to generate a filtered masks is done via clustering or training a small MLP. Since the stable diffusion features are of lower resolution than images, the mlp generated masks are more pixel perfect than the clustering method. While the segmentation masks are significantly better than RobustNeRF masks, persistent transient objects are still challenging. If an object is static for more than a third of the images or the color scheme of the transient object is close to the background, the reconstruction error may not be high enough for our detector to trigger. Access to video frames can help with tracking the moving objects and removing them even if in some frames they are static or blend with the background. 

There are two main challenges still blocking casual capturing a 3D model of a scene. The first one is a robust camera calibration technique. Our methods, both RobustNerf and SpotlessSplats, rely on given camera coordinates. Therefore, a  camera calibration failure due to the presence of transients would disable the 3D modeling completely. Hence, the next step in casual capture rendering would be to improve the camera calibration in the presence of moving objects. An iterative refinement technique between structure-from-motion and SpotlessSplats segmenting the inconsistent parts of the scene is a valid research direction. It would be paramount to consider all the frames in a video as the task would be more akin to tracking the moving objects. Coincidentally, upgrading the segmentation task of transient objects into the tracking of transient objects can mitigate the second challenge of persistent transient objects. Currently, we only rely on outlier detection techniques in robust optimization for detecting the transient object. In the extreme case of an object being present in around half the frames the definition of an outlier becomes unclear. If the input format is a video of the scene tracking moving objects can resolve the standing transients. An interesting direction that would also align with object oriented learning is pixels or objects having a mobility score. So rather than using stable diffusion features, we use an off-the-shelf model that also understands the chance of that object or part moving in the world. An state of the art adaptation of FlowCapsules would be able to identify such objects. Then when faced with an incosistency in the scene (e.g. a parked car or pavement) the 3D rendering model can ignore the images with higher mobility score.

In conclusion, we show that unsupervised object based methods have various merits. FlowCapsules' unsupervised approach enables a comprehensive scene understanding by decomposing the scene in layers capable of explaining even the occluded regions. RobustNerf and SpotLessSplats show the necessity of an unsupervised approach when facing an application with an extremely limited number of images that requires flexibility in the definition of an object (such as a shadow object). The limitations presented here suggest many exciting future directions  to achieve a casual and in the wild unsupervised object based visual understanding.

  \backmatter
  
    \bibliographystyle{apalike}
    \bibliography{main}
\end{document}